\documentclass{article}

 \usepackage[preprint]{neurips_2026}

\usepackage[utf8]{inputenc} 
\usepackage[T1]{fontenc}    
\usepackage{hyperref}       
\usepackage{url}            
\usepackage{booktabs}       
\usepackage{array}
\usepackage{longtable}
\usepackage{pdflscape}
\usepackage{multirow}
\usepackage{amsfonts}       
\usepackage{nicefrac}       
\usepackage{microtype}      
\usepackage[table]{xcolor}  
\usepackage{graphicx}
\usepackage{caption}
\usepackage{wrapfig}
\usepackage[most]{tcolorbox}
\usepackage{fvextra}
\definecolor{groupgray}{gray}{0.94}
\definecolor{oursblue}{RGB}{237,246,255}
\newtcolorbox{algorithmfigurebox}{
  enhanced,
  sharp corners,
  boxrule=0pt,
  colback=white,
  colframe=white,
  left=0pt,
  right=0pt,
  top=3pt,
  bottom=3pt,
  borderline north={1.1pt}{0pt}{black},
  borderline south={1.1pt}{0pt}{black},
  before skip=0pt,
  after skip=0pt
}
\newcommand{\algcomment}[1]{\textcolor{gray}{\textbf{\# #1}}}
\newcommand{\algproc}[1]{\mbox{\textsc{#1}}}
\newcommand{\algbar}{\vrule width 0.35pt height 1.35ex depth 0.45ex}
\newcommand{\tabnum}[1]{\textsf{#1}}
\title{FunctionEvolve: Structure-Guided Symbolic Regression with LLMs}

\author{%
  Zeyu Xia\textsuperscript{1,2} \quad
  Jun Zhu\textsuperscript{2,3} \quad
  Dong Yan\textsuperscript{1}\thanks{Corresponding author.} \\
  \textsuperscript{1}Bosch Center for Artificial Intelligence \\
  \textsuperscript{2}Department of Computer Science and Technology, Tsinghua University \\
  \textsuperscript{3}Tsinghua-Bosch Joint Center for ML, Tsinghua University \\
  \texttt{xia-zy25@mails.tsinghua.edu.cn} \\
  \texttt{dcszj@mail.tsinghua.edu.cn} \\
  \texttt{sproblvem@gmail.com} \\
}

\begin{document}

\maketitle

\begin{abstract}
  Symbolic regression aims to uncover explicit scientific laws from data. Recent methods use LLMs to guide mutation from background text, which is more directed than random genetic programming. However, exact symbolic recovery requires both semantic guidance and explicit structure, so that domain-informed search are carried out through valid symbolic representation. Current LLM-driven systems remain structure-blind: they select among opaque candidates, lack explicit mechanisms for local mutation, and rely on brittle coefficient fitting that can undervalue correct skeletons. We propose FunctionEvolve, an evolutionary framework using expression trees to organize the whole search: structural summaries promote diverse parent selection, local tree edits preserve useful subexpressions, and structure-aware fitting decomposes, constrains, and simplifies coefficients for more reliable scoring. It uses only elementary function families, without additional domain-specific rules limiting generalization. On the 129-task synthetic subset of LLM-SRBench, FunctionEvolve with \emph{Claude Opus 4.6} recovers 107 exact forms, reaching 82.9\% $\mathrm{SA@50}$, 4.5x above same-backbone baselines, and 55.8\% $\mathrm{SA@1}$, 3.6x above the strongest previously published top-1 result. Ablations show that structure-visible search is central to reliable recovery, with LLM-guided refinements and structure-aware coefficient optimization serving as essential proposal and scoring mechanisms. We also audit the benchmark and show that collinearity in its materials-science subset creates identifiability issues.

\end{abstract}

\vspace{-1.0em}
\section{Introduction}
\label{sec:intro}
\vspace{-0.5em}
Scientific discovery increasingly requires methods that infer interpretable laws from experimental data \citep{kramer2023automatedscientificdiscoveryequation}. Black-box models can fit observations well, but they often obscure the mechanisms governing natural phenomena. Symbolic Regression (SR) addresses this gap by recovering explicit mathematical expressions from data \citep{Langley1981}, enabling interpretation, extrapolation, and knowledge transfer \citep{schmidt2009distilling}.

SR searches a combinatorial space of formulas, and genetic programming (GP) remains a standard way to represent expressions as abstract syntax trees (ASTs) and evolve them through local operators \citep{koza_genetic_1994}. However, SR is NP-hard \citep{virgolin2022symbolicregressionnphard}, and GP driven mainly by random variation and data-fit selection often lacks the scientific priors needed to navigate large hypothesis spaces efficiently. Recent LLM-driven methods therefore use background information typically describes the data scenario, including variable meanings and the scientific process being modeled and model knowledge to guide equation discovery. However, new synthetic benchmarks show that semantic guidance alone still leaves a large gap in exact symbolic recovery \citep{shojaee2025llmsrbenchnewbenchmarkscientific}.

This gap prompts a necessary re-examination of current SR methodologies. We identify three shared bottlenecks in traditional GP and its LLM-driven variants. First, the search often starts from generic seeds rather than useful structural priors. Second, parent selection is usually driven by fixed heuristics over opaque candidates, making it difficult to allocate search budget across genuinely different skeletons. Third, unreliable coefficient optimization can make structurally correct skeletons appear unpromising, causing them to be discarded prematurely.

Explicit expression structure offers a direct way to address these bottlenecks, especially for mutation, where progress often comes from local edits rather than full rewrites. For example, if the current candidate is $a_1\sin(x)+a_2x$ and the target is $b_1\sin(b_2x+b_3)+b_4x^{b_5}$, the remaining progress mainly requires editing the sine argument and the power term while preserving the rest. This is precisely where structure matters. GP can modify subexpressions without rewriting the whole formula, but its edits are largely random and do not exploit task context. Prior LLM-driven systems are more directed, but their textual or code-level edits are not necessarily AST-local, so semantically plausible revisions can still overwrite useful parent structure. FunctionEvolve therefore makes AST structure explicit and operative: AST-rule edits provide systematic local add/delete operations, while LLM-guided proposals use the exposed tree structure to make context-dependent refinements that preserve useful parent substructures. Beyond mutation, the same explicit AST structure helps the Selector enforce structural diversity and gives the structure-aware coefficient optimizer the skeleton needed to separate linear from nonlinear parameters, snap structurally constrained exponents, and simplify algebraically equivalent forms, reducing coefficient-search difficulty.

\begin{tcolorbox}[enhanced, sharp corners, boxrule=0.6pt, colback=oursblue!60!white, colframe=black!55, left=5pt, right=5pt, top=4pt, bottom=4pt, before skip=0.4em, after skip=0.4em]
\textbf{Core insight:} Explicit AST structure makes mutation follow SR's local-editing nature and also benefits parent-diversity selection and coefficient fitting.
\end{tcolorbox}

Building on this insight, we develop FunctionEvolve as an equation-discovery framework organized around AST-visible search. Our contributions are as follows:
\begin{itemize}
  \item We present FunctionEvolve, a structure-guided LLM framework for equation discovery that uses explicit expression trees to coordinate selection, mutation, and coefficient optimization.
  \item On the 129-task synthetic subset of LLM-SRBench, FunctionEvolve achieves 82.9\% $\mathrm{SA@50}$ and 55.8\% $\mathrm{SA@1}$ with \emph{Claude Opus 4.6}, corresponding to a 4.5x improvement over same-backbone baselines and a 3.6x top-1 improvement over strongest published PiT-PO.
  \item Ablations show that structure-visible search, LLM-guided local refinements, and structure-aware coefficient optimization are all critical for reliable symbolic recovery.
  \item We identify critical limitations in LLM-SRBench, including input-variable linear dependence and weak nonlinear factor variation that undermine structural identifiability.
\end{itemize}

\vspace{-1.0em}
\section{Related Work}
\label{sec:related_work}
\vspace{-0.5em}
Symbolic regression has long been studied as a route to scientific discovery, from early empirical-law discovery systems \citep{Langley1981} to later work recovering interpretable laws from experimental data \citep{schmidt2009distilling}. Modern SR spans symbolic neural networks, reinforcement learning, physics-inspired methods, and evolutionary systems \citep{martius2016extrapolationlearningequations,sahoo2018learningequationsextrapolationcontrol,petersen2021deepsymbolicregressionrecovering,doi:10.1126/sciadv.aay2631,cranmer2023interpretablemachinelearningscience}. Among them, GP-style methods remain strong baselines in comparisons \citep{lacava2021contemporarysymbolicregressionmethods}: they represent expressions as trees \citep{koza_genetic_1994}, enabling structured variation operators such as context-preserving, grammar-based, and strongly typed GP \citep{dhaesleer1994contextpreserving,whigham:1995:GBGP,montana:stgpEC}. Related work on prior-guided seeding, diversity, and coefficient fitting shows that symbolic recovery benefits from both expression-structure search and reliable numerical parameter identification \citep{lu2015usinggeneticprogramming,mundhenk2021symbolicregressionneuralguidedgenetic,burlacu2023populationdiversity,kommenda2020parameteridentification,reis2024benchmarkingsymbolicregressionconstant}.

Recent work applies LLMs to SR by using model knowledge and task context to guide equation proposal, refinement, or selection. Representative systems include LLM-SR, LLM4Ed, DRSR, SR-Scientist, PiT-PO, and LaSR \citep{shojaee2024llm,du2024llm4edlargelanguagemodels,wang2025drsrllmbasedscientific,xia2025srscientist,wang2026llmbasedscientificequationdiscovery,grayeli2024symbolicregressionlearnedconcept}; evolutionary coding agents such as OpenEvolve, AlphaEvolve, and ShinkaEvolve also demonstrate LLM-guided program evolution \citep{openevolve,novikov2025alphaevolvecodingagentscientific,lange2025shinkaevolveopenendedsampleefficientprogram}. These methods add semantic guidance beyond random GP, but often treat expressions as strings, programs, or opaque candidates rather than explicit expression trees shared across selection, mutation, and fitting.

The rise of LLM-based SR also makes benchmark design more important. Literature-derived benchmarks such as AI-Feynman, SRBench, and EmpiricalBench are valuable but may overlap with LLM pretraining data \citep{doi:10.1126/sciadv.aay2631,lacava2021contemporarysymbolicregressionmethods,cranmer2023interpretablemachinelearningscience}. LLM-SRBench reduces this risk with synthetic scientific tasks and shows that existing LLM-driven SR methods still struggle with exact symbolic recovery \citep{shojaee2025llmsrbenchnewbenchmarkscientific}; FunctionEvolve targets this gap with AST-visible LLM search and structure-aware coefficient fitting.

\vspace{-0.8em}
\section{Method}
\label{sec:method}
\vspace{-0.5em}
\subsection{Problem Formulation}
\vspace{-0.5em}
We consider LLM-driven symbolic regression over a dataset $\mathcal{D}=\{(\mathbf{x}_i, y_i)\}_{i=1}^n$ together with a task-level context $T$, where $\mathbf{x}_i \in \mathbb{R}^d$ is the numerical input and $y_i \in \mathbb{R}$ is the target. The context $T$ is a natural-language description of the task that is shared across all $n$ samples; depending on the benchmark, it may include the problem background and scientific scenario, the data source, and the names, physical meanings, and units (dimensions) of the variables. For dimension-only tasks such as AI-Feynman entries, $T$ may provide only variable names and units. The goal is to recover a concise, generalizable expression $\tilde{f}$ with $\tilde{f}(\mathbf{x}_i)\approx y_i$, using $T$ to guide the search.

\vspace{-0.5em}
\subsection{Overview}
\vspace{-0.5em}
Our method formulates SR as one-shot initialization followed by iterative evolution. FunctionEvolve first uses the Generator to synthesize task-conditioned domain knowledge and seed expressions. It then uses the Selector to choose promising parents, the Mutator to expand them through AST-rule and LLM-guided mutations, and the structure-aware Optimizer to fit, simplify, and score valid offspring before reinsertion. Unlike prior LLM-driven methods, FunctionEvolve uses LLMs for seed generation, parent selection, and mutation, with each part anchored to explicit expression structure. The workflow is illustrated in Figure~\ref{fig:functionevolve_a}, and the complete algorithm is in Appendix~\ref{app:functionevolve_algorithm}.


\vspace{-0.3em}
\begin{center}
  \centering
  \includegraphics[width=\textwidth]{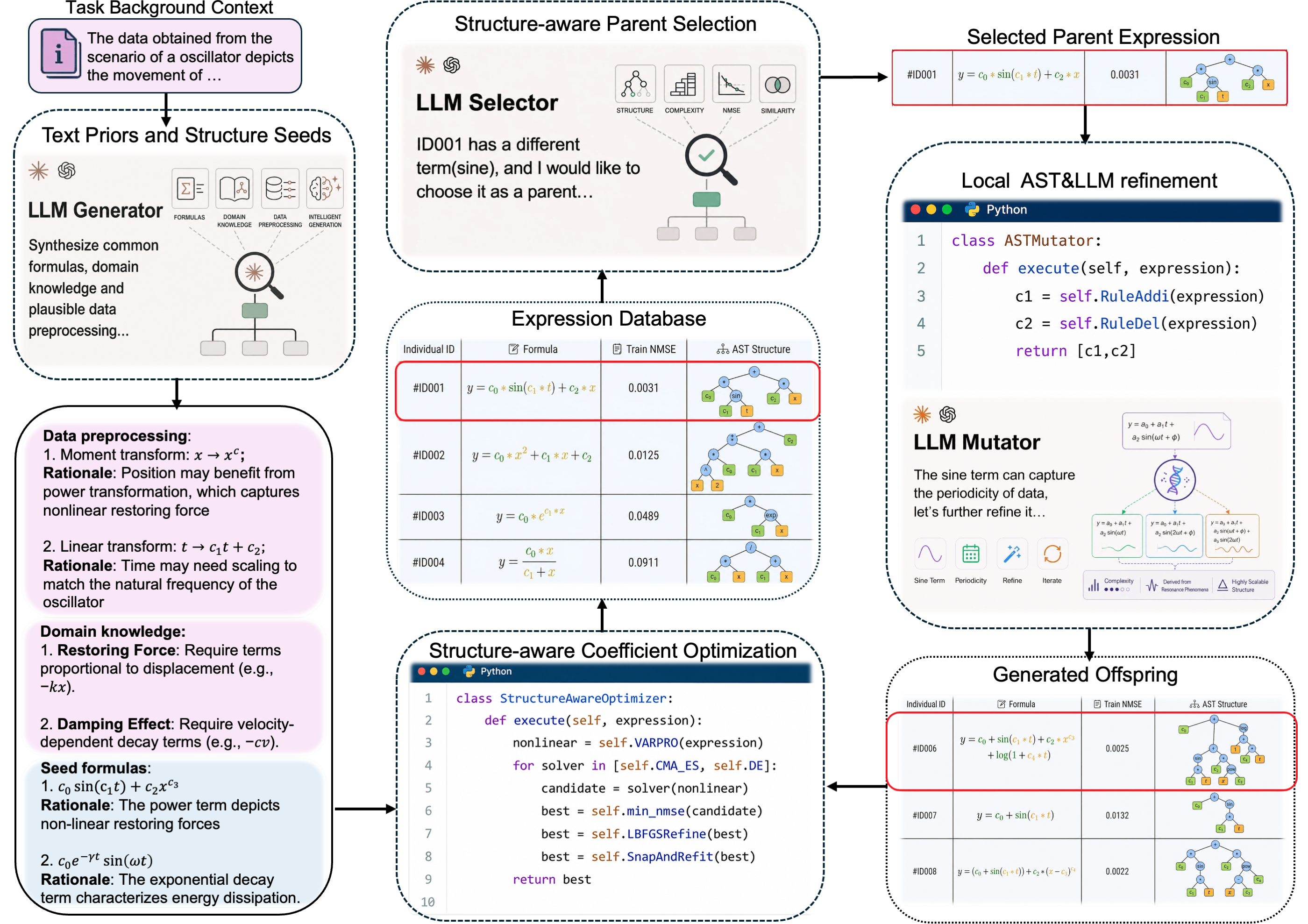}
  
  \captionof{figure}[Overview of FunctionEvolve]{FunctionEvolve initializes candidate expressions from task context and then iteratively selects, mutates, optimizes, and evaluates them over an evolving expression database.}
  \label{fig:functionevolve_a}
\vspace{-1.0em}
\end{center}

\vspace{-0.3em}
\subsection{Generator}

The Generator provides task-conditioned structural starting points through two roles: (1) it uses the available task context and LLM knowledge to synthesize domain formulas and structural heuristics; and (2) it initializes seed expressions from these patterns so the search does not begin from an unstructured pool. These outputs provide initial starting points and reusable contextual guidance for evolution, rather than encoded domain laws or fixed mutation rules. Details of the Generator procedure and prompt are provided in Appendix~\ref{app:generator_details}.

\vspace{-0.8em}
\subsection{Selector}

The Selector determines which expressions to expand next. Each iteration, the system extracts an evolution-tree summary and provides it to the Selector with the task context and previous selections. For each candidate node, this summary combines structural information (AST structure, parameter count, tree depth, and operator counts) with numerical information (training NMSE and fitted parameter values). Based on this information, the Selector returns a small set of parent nodes, allocating limited search budget to more promising directions. Details of the evolutionary tree and Selector are provided in Appendices~\ref{app:evolutionary_tree} and~\ref{app:selector_summary}, respectively.

The Selector is not purely greedy: it favors low-error, simple candidates while using selection history to avoid repeatedly expanding similar branches. To prevent data leakage, the tree summary exposes only training error, never test or out-of-distribution error.

\vspace{-0.5em}
\subsection{Mutator}

\begin{figure*}[t]
  \centering
  \includegraphics[width=0.98\textwidth]{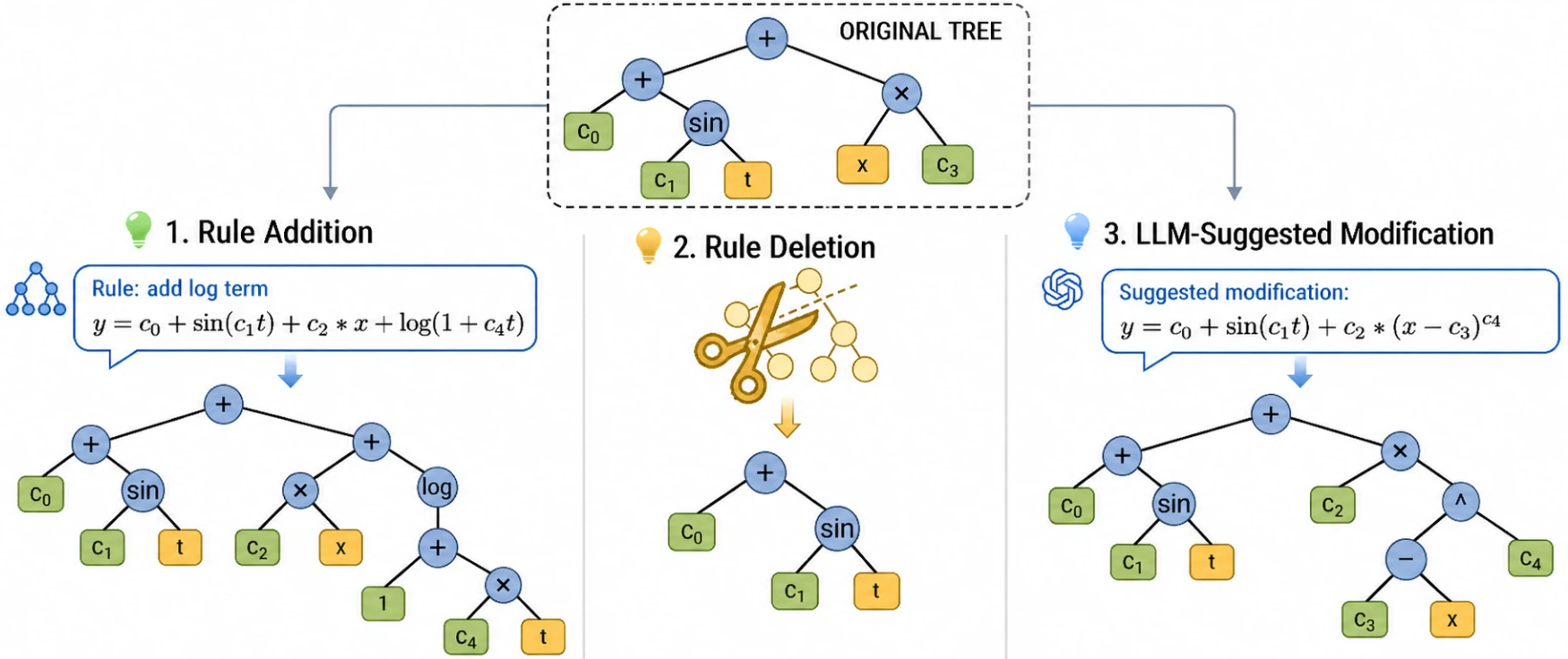}
  \caption{AST rule add/del and LLM-guided local mutation generate offspring from selected parents.}
  \label{fig:functionevolve_c}
\vspace{-1.2em}
\end{figure*}
The Mutator generates candidates by locally adjusting parent nodes selected by the Selector. For each selected parent expression $f(\mathbf{x};\mathbf{c})$, it applies three complementary operations, as illustrated in Figure~\ref{fig:functionevolve_c}: (1) \textit{Rule addition}, which preserves the parent and attaches fixed elementary functions (Table~\ref{tab:ast_mutation_templates}) through a small set of attachment operators detailed below; (2) \textit{Rule deletion}, which prunes or unwraps AST substructures to test whether current formula parts are redundant; and (3) \textit{LLM-guided proposal}, which uses parent structure, fitted information, rule-generated candidates, search history, and domain knowledge to suggest non-template, context-dependent refinements.

\begin{center}
\centering
\captionof{table}{Elementary content library used by rule-addition. For multivariate tasks, single-variable templates instantiate for each $x_j$, while pair families use ordered variable pairs $(x_i,x_j)$.}
\label{tab:ast_mutation_templates}
\setlength{\tabcolsep}{4pt}
\begin{tabular*}{\textwidth}{@{\extracolsep{\fill}}p{0.20\textwidth}p{0.34\textwidth}p{0.36\textwidth}@{}}
\toprule
Family & Library content $g$ & Multivariate handling \\
\midrule
Linear & $c_1 x_j + c_2$ & -- \\
Trigonometric & $c_1\sin(c_2 x_j + c_3)$ & -- \\
Power & $c_1 x_j^{c_2}$ & $c_1(x_i+c_2)^{c_3}x_j$ \\
Exponential & $c_1\exp(c_2 x_j)$ & -- \\
Logarithmic & $c_1\log(1+c_2 x_j)$ & -- \\
Rational-fraction & $(c_1x_j+c_2)/(c_3x_j+c_4)$ & $(c_1x_i+c_2)/(c_3x_j+c_4)$ \\
\bottomrule
\end{tabular*}
\end{center}

The deterministic library uses only generic elementary content and common algebraic attachment operators, without domain-specific rules. Rule addition combines this content with the parent through addition, multiplication, and division; unary wraps additionally apply $\exp(c f)$, $\log(1+c f)$, $\sin(c f)$, $|c f|$, and $f^c$. Its breadth comes from operator families rather than benchmark- or domain-level priors, preserving low inductive bias while making compact products, ratios, and compositions reachable. After mutation, candidates are normalized and deduplicated by structural fingerprints before entering the expression database. This design combines reliable rule-based local search with flexible LLM-guided refinements, allowing the search to expand beyond fixed templates while avoiding saturation by repeated or invalid candidates. Details of the AST representation, rule-based and LLM-guided mutation, and normalization procedures are provided in Appendix~\ref{app:mutator_details}.

\vspace{-0.5em}
\subsection{Optimizer}

\begin{figure*}[t]
  \centering
  \includegraphics[width=0.98\textwidth]{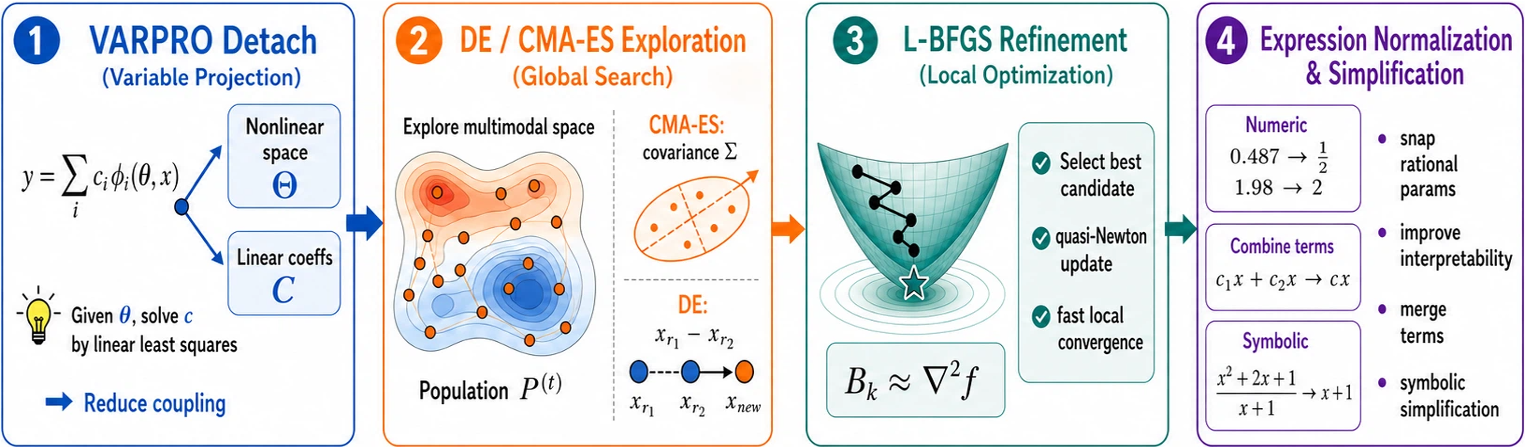}
  \caption{Structure-aware optimizer pipeline: variable projection, DE/CMA-ES global exploration, L-BFGS local refinement, parameter snapping, and symbolic simplification.}
  \label{fig:functionevolve_b}
\vspace{-1.2em}
\end{figure*}

The Optimizer fits coefficients for each fixed candidate skeleton and returns its score. Rather than treating coefficient fitting as black-box optimization, FunctionEvolve exploits expression structure throughout the pipeline (Figure~\ref{fig:functionevolve_b}). Coefficients at different structural positions carry algebraic priors: linear parameters can be solved by variable projection and OLS, remaining nonlinear parameters are searched with global and local optimizers, structurally constrained parameters can be snapped to stable integer or rational values, and redundant forms can be removed through coefficient merging and symbolic simplification before refitting. These priors reduce redundant dimensionality and make correct skeletons less likely to be discarded because of poor exploration or local optima. They are generic SR operations and therefore do not encode benchmark shortcuts. See Appendix~\ref{app:optimizer_details} for details.

\vspace{-1em}
\section{Experiments}
\label{experiments}
\vspace{-0.5em}
\subsection{Main Results}
\label{sec:main_results}
\vspace{-0.5em}
The main comparison in this subsection is conducted on the synthetic benchmark set of LLM-SRBench \citep{shojaee2025llmsrbenchnewbenchmarkscientific}. We choose this benchmark because its synthetic construction reduces the risk of contamination from memorized formulas in LLM pretraining, thereby providing a cleaner test bed for the primary symbolic-regression comparison.

Symbolic regression is commonly evaluated using three complementary metrics: (1) \textbf{Symbolic accuracy at $k$ ($\mathrm{SA@}k$)}, whether at least one of the top-$k$ candidates ranked by training NMSE matches the ground-truth symbolic structure, usually evaluated using an LLM-as-a-judge protocol; (2) \textbf{Accuracy to tolerance $\mathrm{Acc}_{\tau}$}, the number of tasks whose top-1 candidate reaches maximum relative error over all evaluation points at most $\tau$, where $\tau$ is typically set to 0.1 or 0.01; and (3) \textbf{Normalized mean squared error (NMSE)}, a metric that measures aggregate numerical error.

Since $\mathrm{SA@}k$ is calculated based on LLMs, we further audit the reliability of its LLM-as-a-judge protocol. On 500 sampled verification cases, \emph{GPT-5.2} and \emph{Claude Opus 4.6} agree on 476 decisions; manual auditing over these 500 cases gives judge accuracies of 96.4\% and 98.6\%, respectively. In the actual experiments, each candidate is verified by both models: agreeing verdicts are accepted, while disagreements are manually adjudicated (See Appendix~\ref{app:llm_judge_reliability}).

Table~\ref{tab:main_results} reports symbolic accuracy and numerical accuracy for each domain and for the full 129-task benchmark. We emphasize both $\mathrm{SA@50}$ and $\mathrm{SA@1}$: $\mathrm{SA@50}$ measures whether the search generates a correct symbolic form within its top-50 training-NMSE-ranked candidates, while $\mathrm{SA@1}$ measures whether the method's single selected expression already recovers the target structure. Numerical metrics are complementary: NMSE can be saturated by structurally incorrect surrogate expressions, and worst-point $\mathrm{Acc}_{\tau}$ can be brittle when small $y_{\mathrm{true}}$ values amplify relative-error thresholds. The ground-truth row is therefore included only as a dataset reference ceiling, not as a competing method.\footnote{An additional baseline, SR-Scientist~\citep{xia2025srscientist}, is compared separately in Appendix~\ref{app:strict_numerical_accuracy} rather than in this table: its headline $\mathrm{Acc}_{\tau}$ discards the worst 5\% of evaluation points before taking the maximum, making it a relaxed variant aligned with our $95\%\mathrm{Acc}_{\tau}$ rather than our strict all-point $\mathrm{Acc}_{\tau}$, and it reports symbolic accuracy only for a few representative methods, without an $\mathrm{SA@}k$ analysis.}


\begin{table*}[t]
\centering
\caption{Main results on the 129-task synthetic subset of LLM-SRBench. Entries report $\mathrm{SA@50}$ ($\mathrm{SA@1}$), $\mathrm{Acc}_{0.1}$ task counts, and median test NMSE; table headers abbreviate these as SA, Acc, and NMSE. Parenthesized-only SA values are published top-1 results without reported $\mathrm{SA@50}$. Bold marks the best non-reference values, while the ground-truth row is a reference upper bound. \textsuperscript{*} and \textsuperscript{**} denote results cited from LLM-SRBench~\citep{shojaee2025llmsrbenchnewbenchmarkscientific} and PiT-PO~\citep{wang2026llmbasedscientificequationdiscovery}.}
\label{tab:main_results}
\fontsize{6.5pt}{7.5pt}\selectfont
\setlength{\tabcolsep}{1pt}
\renewcommand{\arraystretch}{1.12}
\begin{tabular}{@{}>{\raggedright\arraybackslash}p{0.15\textwidth}*{5}{ccc}@{}}
\toprule
& \multicolumn{3}{c}{Chemistry (36)} & \multicolumn{3}{c}{Biology (24)} & \multicolumn{3}{c}{Physics (44)} & \multicolumn{3}{c}{MatSci (25)} & \multicolumn{3}{c}{Total (129)} \\
\cmidrule(lr){2-4}\cmidrule(lr){5-7}\cmidrule(lr){8-10}\cmidrule(lr){11-13}\cmidrule(lr){14-16}
\textbf{Model} & SA$\uparrow$ & Acc$\uparrow$ & NMSE$\downarrow$ & SA$\uparrow$ & Acc$\uparrow$ & NMSE$\downarrow$ & SA$\uparrow$ & Acc$\uparrow$ & NMSE$\downarrow$ & SA$\uparrow$ & Acc$\uparrow$ & NMSE$\downarrow$ & SA$\uparrow$ & Acc$\uparrow$ & NMSE$\downarrow$ \\
\midrule
\rowcolor{groupgray}\multicolumn{16}{c}{\textit{Direct Prompting}} \\
\midrule
\emph{GPT-4o-mini}\textsuperscript{*} & (\tabnum{0}) & \tabnum{5} & \tabnum{0.0221} & (\tabnum{0}) & \tabnum{1} & \tabnum{0.4648} & (\tabnum{2}) & \tabnum{4} & \tabnum{0.0647} & (\tabnum{0}) & \tabnum{0} & \tabnum{0.0484} & (\tabnum{2}) & \tabnum{10} & -- \\
\emph{Claude Opus 4.6} & \tabnum{0} (\tabnum{0}) & \tabnum{3} & \tabnum{1.50e-2} & \tabnum{1} (\tabnum{1}) & \tabnum{3} & \tabnum{2.43e-4} & \tabnum{0} (\tabnum{0}) & \tabnum{4} & \tabnum{2.18e-3} & \tabnum{1} (\tabnum{0}) & \tabnum{8} & \tabnum{2.86e-5} & \tabnum{2} (\tabnum{1}) & \tabnum{18} & \tabnum{1.36e-3} \\
\midrule
\rowcolor{groupgray}\multicolumn{16}{c}{\textit{PySR \citep{cranmer2023interpretablemachinelearningscience}}} \\
\midrule
PySR\textsuperscript{*} & (\tabnum{0}) & \tabnum{15} & -- & (\tabnum{0}) & \tabnum{6} & -- & (\tabnum{2}) & \tabnum{13} & -- & (\tabnum{0}) & \tabnum{17} & -- & (\tabnum{2}) & \tabnum{51} & -- \\
\midrule
\rowcolor{groupgray}\multicolumn{16}{c}{\textit{OpenEvolve \citep{openevolve}}} \\
\midrule
\emph{Claude Opus 4.6} & \tabnum{7} (\tabnum{2}) & \tabnum{11} & \tabnum{8.19e-7} & \tabnum{8} (\tabnum{1}) & \tabnum{2} & \tabnum{9.37e-6} & \tabnum{0} (\tabnum{0}) & \tabnum{5} & \tabnum{1.21e-4} & \tabnum{9} (\tabnum{2}) & \tabnum{9} & \tabnum{1.07e-7} & \tabnum{24} (\tabnum{5}) & \tabnum{27} & \tabnum{4.30e-6} \\
\midrule
\rowcolor{groupgray}\multicolumn{16}{c}{\textit{LaSR \citep{grayeli2024symbolicregressionlearnedconcept}}} \\
\midrule
Llama-3.1-8B\textsuperscript{*} & (\tabnum{0}) & \tabnum{10} & \tabnum{2.77e-4} & (\tabnum{1}) & \tabnum{4} & \tabnum{2.73e-4} & (\tabnum{2}) & \tabnum{11} & \tabnum{0.0018} & (\tabnum{2}) & \tabnum{16} & \tabnum{7.44e-5} & (\tabnum{5}) & \tabnum{41} & -- \\
\emph{GPT-4o-mini}\textsuperscript{*} & (\tabnum{1}) & \tabnum{14} & \tabnum{9.11e-5} & (\tabnum{2}) & \tabnum{5} & \tabnum{1.53e-4} & (\tabnum{4}) & \tabnum{14} & \tabnum{9.94e-4} & (\tabnum{7}) & \tabnum{18} & \tabnum{9.23e-6} & (\tabnum{14}) & \tabnum{51} & -- \\
\midrule
\rowcolor{groupgray}\multicolumn{16}{c}{\textit{LLM-SR \citep{shojaee2024llm}}} \\
\midrule
\emph{GPT-4o-mini}\textsuperscript{*} & (\tabnum{4}) & \tabnum{19} & \tabnum{4.12e-6} & (\tabnum{4}) & \tabnum{7} & \tabnum{3.06e-6} & (\tabnum{4}) & \tabnum{16} & \tabnum{7.62e-5} & (\tabnum{5}) & \tabnum{22} & \tabnum{3.21e-9} & (\tabnum{17}) & \tabnum{64} & -- \\
\emph{Claude Opus 4.6}\footnotemark & \tabnum{9} (\tabnum{4}) & \tabnum{14} & \tabnum{4.13e-7} & \tabnum{8} (\tabnum{5}) & \tabnum{6} & \tabnum{2.15e-7} & \tabnum{0} (\tabnum{0}) & \tabnum{3} & \tabnum{7.11e-5} & \tabnum{7} (\tabnum{2}) & \tabnum{17} & \tabnum{2.90e-9} & \tabnum{24} (\tabnum{11}) & \tabnum{40} & \tabnum{9.80e-7} \\
\midrule
\rowcolor{groupgray}\multicolumn{16}{c}{\textit{PiT-PO \citep{wang2026llmbasedscientificequationdiscovery}}} \\
\midrule
\emph{Llama-3.1-8B} (RL)\textsuperscript{**} & (\tabnum{5}) & \tabnum{28} & \tabnum{4.13e-7} & (\tabnum{7}) & \tabnum{17} & \tabnum{9.37e-8} & (\tabnum{5}) & \tabnum{18} & \tabnum{6.57e-5} & (\tabnum{3}) & \textbf{\tabnum{23}} & \tabnum{1.18e-8} & (\tabnum{20}) & \tabnum{86} & -- \\
\midrule
\rowcolor{groupgray}\multicolumn{16}{c}{\textit{FunctionEvolve (Ours)}} \\
\midrule
\rowcolor{oursblue!45}No LLM & \tabnum{7} (\tabnum{3}) & \tabnum{32} & \tabnum{1.76e-12} & \tabnum{12} (\tabnum{8}) & \tabnum{16} & \tabnum{6.66e-13} & \tabnum{7} (\tabnum{5}) & \tabnum{11} & \tabnum{1.89e-5} & \tabnum{3} (\tabnum{0}) & \tabnum{19} & \tabnum{2.83e-14} & \tabnum{29} (\tabnum{16}) & \tabnum{78} & \tabnum{1.47e-10} \\
\rowcolor{oursblue}\emph{Llama-3.1-8B} & \tabnum{20} (\tabnum{5}) & \tabnum{32} & \tabnum{7.52e-13} & \tabnum{20} (\tabnum{9}) & \tabnum{20} & \tabnum{3.85e-13} & \tabnum{17} (\tabnum{6}) & \tabnum{18} & \tabnum{4.59e-8} & \tabnum{5} (\tabnum{3}) & \tabnum{20} & \textbf{\tabnum{1.20e-14}} & \tabnum{62} (\tabnum{23}) & \tabnum{90} & \tabnum{1.17e-12} \\
\rowcolor{oursblue}\emph{Qwen3.6-27B} & \tabnum{27} (\tabnum{8}) & \textbf{\tabnum{35}} & \tabnum{1.99e-13} & \tabnum{21} (\tabnum{11}) & \textbf{\tabnum{23}} & \tabnum{2.08e-13} & \tabnum{29} (\tabnum{16}) & \tabnum{30} & \tabnum{5.19e-13} & \tabnum{9} (\tabnum{3}) & \tabnum{21} & \tabnum{1.44e-14} & \tabnum{86} (\tabnum{38}) & \tabnum{109} & \tabnum{2.06e-13} \\
\rowcolor{oursblue}\emph{DeepSeek-V4-Pro} & \textbf{\tabnum{32}} (\textbf{\tabnum{21}}) & \tabnum{34} & \tabnum{2.18e-13} & \tabnum{20} (\tabnum{14}) & \textbf{\tabnum{23}} & \tabnum{2.08e-13} & \tabnum{33} (\tabnum{30}) & \textbf{\tabnum{33}} & \tabnum{5.15e-13} & \tabnum{14} (\textbf{\tabnum{6}}) & \tabnum{21} & \textbf{\tabnum{1.20e-14}} & \tabnum{99} (\tabnum{71}) & \tabnum{111} & \tabnum{1.78e-13} \\
\rowcolor{oursblue}\emph{GPT-5.2 medium} & \tabnum{30} (\tabnum{13}) & \tabnum{34} & \tabnum{2.08e-13} & \tabnum{20} (\tabnum{15}) & \textbf{\tabnum{23}} & \tabnum{3.80e-13} & \tabnum{37} (\textbf{\tabnum{36}}) & \textbf{\tabnum{33}} & \tabnum{4.03e-13} & \textbf{\tabnum{16}} (\tabnum{5}) & \tabnum{21} & \tabnum{1.32e-14} & \tabnum{103} (\tabnum{69}) & \tabnum{111} & \tabnum{2.08e-13} \\
\rowcolor{oursblue}\emph{Claude Opus 4.6} & \tabnum{31} (\tabnum{16}) & \textbf{\tabnum{35}} & \textbf{\tabnum{1.85e-13}} & \textbf{\tabnum{22}} (\textbf{\tabnum{16}}) & \textbf{\tabnum{23}} & \textbf{\tabnum{2.06e-13}} & \textbf{\tabnum{39}} (\tabnum{34}) & \textbf{\tabnum{33}} & \textbf{\tabnum{2.44e-13}} & \tabnum{15} (\textbf{\tabnum{6}}) & \tabnum{22} & \tabnum{1.44e-14} & \textbf{\tabnum{107}} (\textbf{\tabnum{72}}) & \textbf{\tabnum{113}} & \textbf{\tabnum{1.38e-13}} \\
\midrule
\rowcolor{groupgray}\textit{Reference: GT} & \tabnum{36} (\tabnum{36}) & \tabnum{35} & \tabnum{1.49e-13} & \tabnum{24} (\tabnum{24}) & \tabnum{23} & \tabnum{1.43e-13} & \tabnum{44} (\tabnum{44}) & \tabnum{37} & \tabnum{1.14e-13} & \tabnum{25} (\tabnum{25}) & \tabnum{25} & \tabnum{1.87e-14} & \tabnum{129} (\tabnum{129}) & \tabnum{120} & \tabnum{4.98e-14} \\
\bottomrule
\end{tabular}
\vspace{-2.5em}
\end{table*}
\footnotetext{The public LLM-SRBench code reports only top-1 symbolic accuracy. We additionally audit the top-50 training-NMSE-ranked candidates for our \emph{Claude Opus 4.6} rerun and report entries as $\mathrm{SA@50}$ ($\mathrm{SA@1}$). With this released implementation, our rerun is weaker at $\mathrm{SA@1}$ than the originally reported \emph{gpt-4o-mini} result, possibly because the authors used a different evaluation method; related reproducibility concerns are noted by other researchers at \url{https://openreview.net/forum?id=0L4RWQV8Qa}.}

FunctionEvolve greatly improves symbolic recovery under both $\mathrm{SA@50}$ and $\mathrm{SA@1}$. The No LLM variant uses generic seed generation without semantic priors, training-NMSE-ranked selection, programmatic AST mutation, and the same structure-aware optimizer. It already recovers 29 $\mathrm{SA@50}$ forms and 16 $\mathrm{SA@1}$ forms, close to the strongest published PiT-PO top-1 result of 20. This shows that GP-style rule mutation combined with reliable structure-aware fitting is itself a strong symbolic-search baseline. Adding LLM-guided generation, selection, and mutation raises $\mathrm{SA@50}$ to 107 with \emph{Claude Opus 4.6}, while the strongest same-backbone rerun baselines, LLM-SR and OpenEvolve, each reach 24. At top-1, FunctionEvolve reaches 72 $\mathrm{SA@1}$, exceeding PiT-PO by 3.6x.

Additional backbones show the same trend: \emph{DeepSeek-V4-Pro} recovers 99 (71) tasks and is especially strong on chemistry, the mid-sized open-weight \emph{Qwen3.6-27B} reaches 86 (38), and \emph{Llama-3.1-8B} reaches 62 (23), improving substantially over the No LLM setting in $\mathrm{SA@50}$ but only slightly in $\mathrm{SA@1}$. This is notable because PiT-PO also builds on the Llama-3.1-8B backbone but further applies reinforcement learning, whereas our Llama-3.1-8B run uses no RL and still surpasses PiT-PO. In practice, Llama-3.1-8B also has weaker JSON-format following than newer models, so invalid structured outputs are replaced by the deterministic rule-based baseline in our implementation. When the LLM interface fails, the system falls back to the baseline. Therefore, the measured gains provide a conservative estimate of the benefit of LLM-guided search.

The gains are broad across domains: 31/36 chemistry tasks, 22/24 biology tasks, 39/44 physics tasks, and 15/25 materials-science tasks under $\mathrm{SA@50}$, with corresponding $\mathrm{SA@1}$ counts of 16/36, 16/24, 34/44, and 6/25. Worse performance on materials science is partly explained by benchmark identifiability issues discussed in Section~\ref{sec:discussion}. The gap between $\mathrm{SA@50}$ and $\mathrm{SA@1}$ shows that correct equations are often generated but not always ranked first by training NMSE. Our inspection suggests this ambiguity arises because some expressions are difficult to distinguish from high-order rational approximations or series expansions, especially over restricted data ranges.

\begin{tcolorbox}[enhanced, sharp corners, boxrule=0.6pt, colback=oursblue!60!white, colframe=black!55, left=5pt, right=5pt, top=4pt, bottom=4pt, before skip=0.4em, after skip=0.4em]
\textbf{Main result:} The main gap for LLM-driven SR lies in symbolic rather than numerical: FunctionEvolve closes much of this gap while preserving near-GT numerical precision.
\end{tcolorbox}

Numerically, FunctionEvolve with \emph{Claude Opus 4.6} reaches 113 tasks under $\mathrm{Acc}_{0.1}$, close to the GT reference value of 120, and maintains median test NMSE at $1.38\times 10^{-13}$. The updated \emph{Llama-3.1-8B} run recovers 62 (23) symbolic forms and has a low median test NMSE of $1.17\times 10^{-12}$, while reaching 90 tasks under $\mathrm{Acc}_{0.1}$ and 69/61 tasks under the stricter $\mathrm{Acc}_{0.01}/\mathrm{Acc}_{0.001}$ thresholds. This gap should not be interpreted as contradicting the symbolic result: exact symbolic recovery is the primary goal of scientific SR, while worst-point $\mathrm{Acc}_{\tau}$ is a numerically brittle auxiliary metric on this benchmark. Because $\mathrm{Acc}_{\tau}$ divides by $y_{\mathrm{true}}$ pointwise, very small $y_{\mathrm{true}}$ values can amplify tiny numerical perturbations into threshold failures, so even GT equations may not satisfy the relative-error threshold; Appendix~\ref{app:strict_numerical_accuracy} discusses the alternative 95\% $\mathrm{Acc}_{\tau}$ metric and stricter numerical thresholds. These gains are not due to larger LLM budgets: FunctionEvolve averages 66.86 LLM calls per task, versus 1000.27 for LLM-SR and 201.50 for OpenEvolve, with lower price-weighted token cost than LLM-SR and comparable cost to OpenEvolve (Appendix~\ref{app:resource_usage}).

\begin{center}
\centering
\begin{minipage}{0.48\textwidth}
  \centering
  \includegraphics[width=\linewidth]{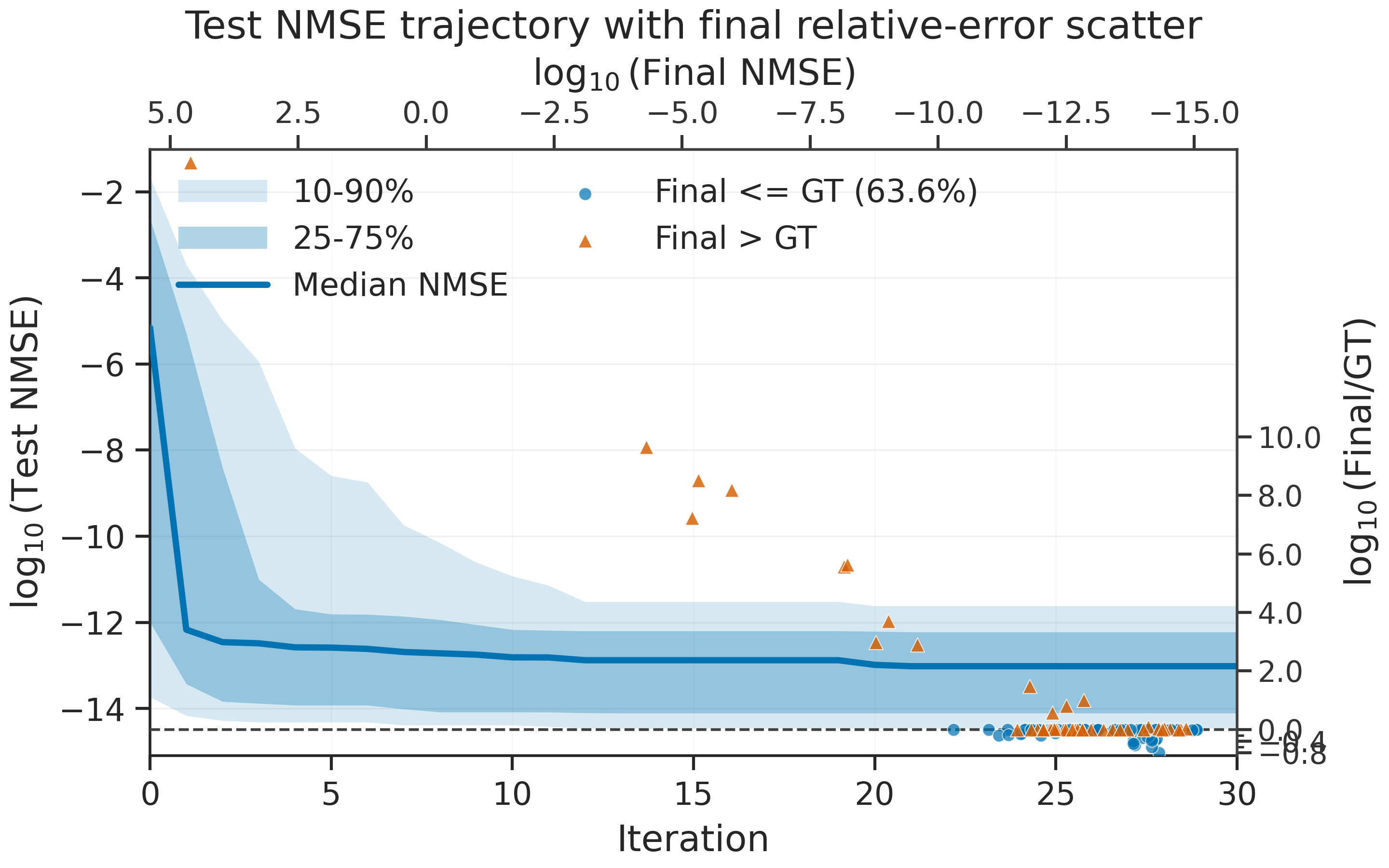}
  {\scriptsize (a) Test NMSE trajectory and final relative-error scatter}
\end{minipage}\hfill
\begin{minipage}{0.48\textwidth}
  \centering
  \includegraphics[width=\linewidth]{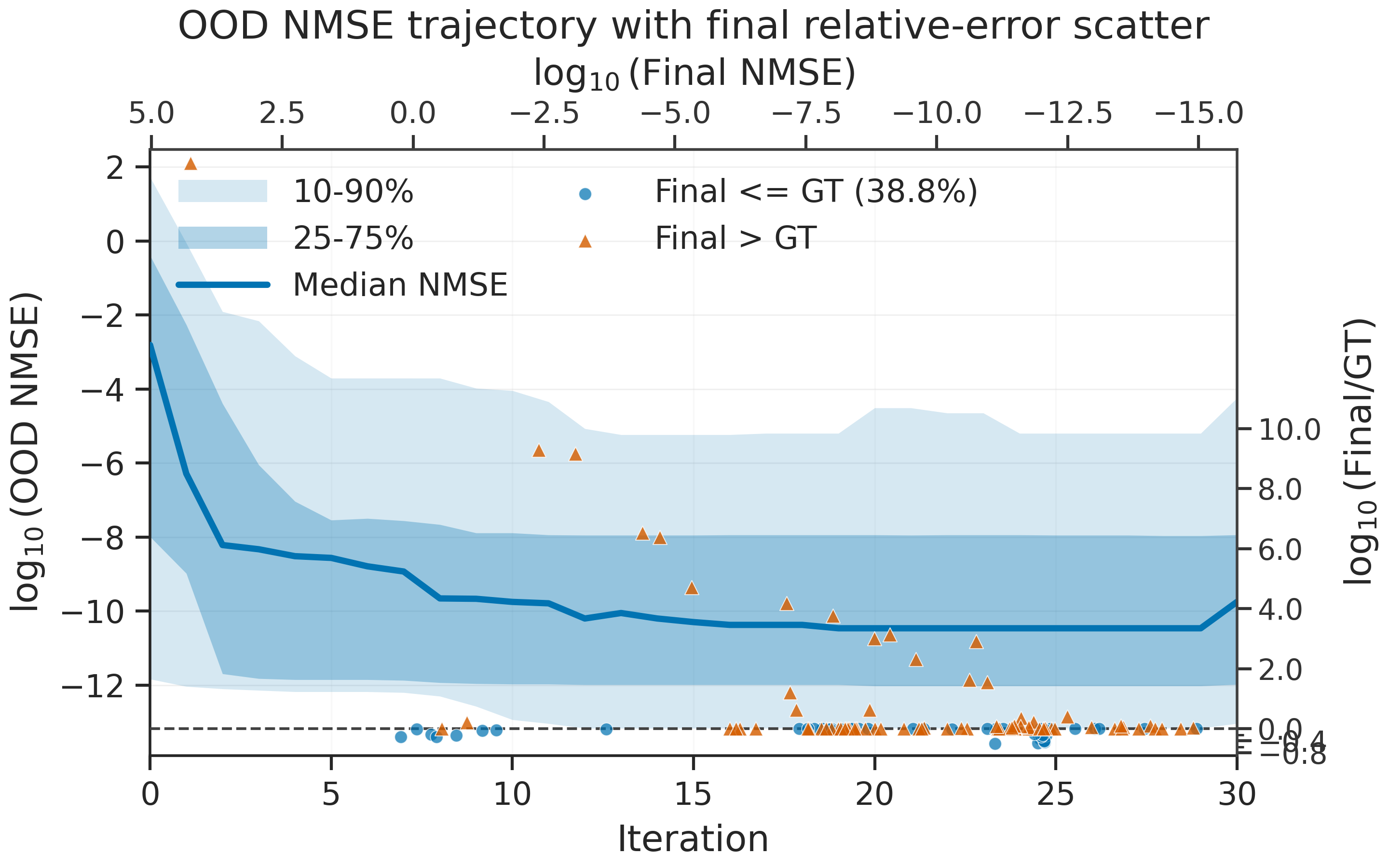}
  {\scriptsize (b) OOD NMSE trajectory and final relative-error scatter}
\end{minipage}
\captionof{figure}{Global NMSE trajectories and final relative errors. Curves show median $\log_{10}(\mathrm{NMSE})$, and shaded bands show 25--75\% and 10--90\% ranges using bottom and left axes. Overlaid points use top and right axes: horizontal coordinate is $\log_{10}(\mathrm{NMSE}_{\mathrm{final}})$ and vertical coordinate is $\Delta=\log_{10}(\mathrm{NMSE}_{\mathrm{final}}/\mathrm{NMSE}_{\mathrm{GT}})$. Blue circles indicate $\Delta\le 0$, and orange triangles indicate $\Delta>0$.}
\label{fig:opus46_nmse_distribution}
\vspace{-0.7em}
\end{center}

\begin{wrapfigure}[14]{r}{0.47\linewidth}
\vspace{-1.5em}
\centering
\includegraphics[width=\linewidth]{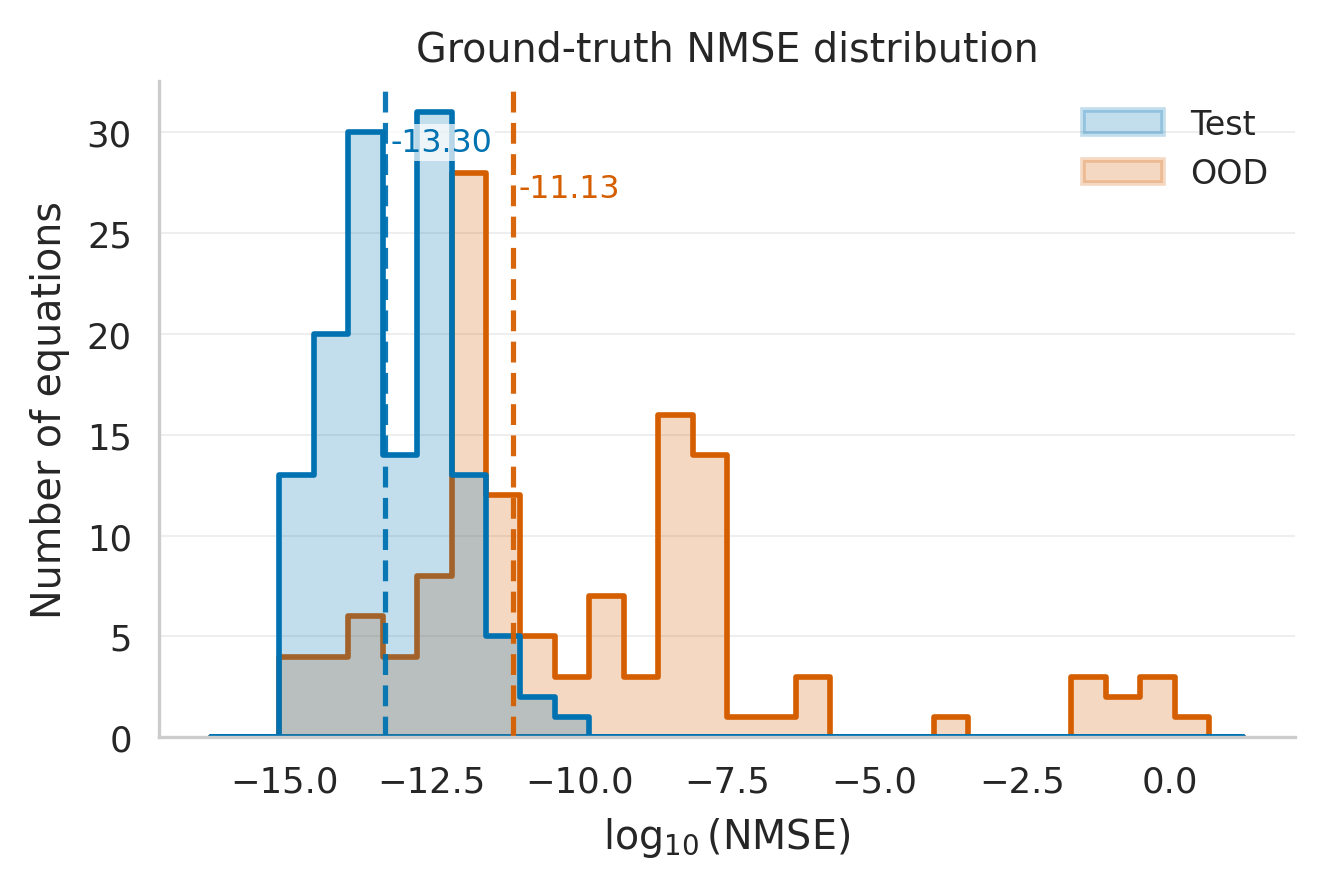}
\vspace{-1.7em}
\caption{Ground-truth NMSE distributions on the test and OOD splits. OOD has a larger and more dispersed error baseline.}
\label{fig:gt_nmse_distribution}
\end{wrapfigure}

To further inspect the numerical behavior behind the aggregate NMSE values, Figure~\ref{fig:opus46_nmse_distribution} shows the test and OOD NMSE search trajectories for the strongest \emph{Claude Opus 4.6} setting, with overlaid scatter points comparing final NMSE against the ground-truth NMSE for each task. On the test split, the global-best median NMSE drops to the numerical floor within the first few iterations and then remains close to the ground-truth NMSE distribution, indicating that the optimizer often fits recovered skeletons to near ground-truth numerical precision. This is also reflected in the paired comparison: the final expression is no worse than the ground-truth expression on 82/129 tasks (63.6\%).

Out-of-distribution evaluation shows the same qualitative error reduction, but the paired comparison is weaker, with 50/129 tasks (38.8\%) no worse than ground truth. This gap should be read against Figure~\ref{fig:gt_nmse_distribution}: even the ground-truth expressions have larger and more dispersed NMSE on held-out OOD samples, with a median about two orders of magnitude higher than on the test split. Thus, this setting has a higher numerical error floor even for the exact laws, and the weaker paired comparison largely reflects the intrinsic difficulty and lower numerical stability of these samples, although task-specific extrapolation effects can still amplify expression-level differences.

\vspace{-0.5em}
\subsection{Evaluation on AI-Feynman}
\label{sec:aifeynman}

\begin{wraptable}{r}{0.57\linewidth}
\vspace{-1.0em}
\centering
\caption{Reported exact-recovery results on AI-Feynman-style benchmarks. Protocols differ, so entries keep their original denominators.}
\label{tab:aifeynman_results}
\scriptsize
\setlength{\tabcolsep}{3pt}
\renewcommand{\arraystretch}{1.12}
\sffamily
\begin{tabular}{@{}p{0.36\linewidth}p{0.28\linewidth}p{0.27\linewidth}@{}}
\toprule
Method & Evaluation set & Recovery \\
\midrule
LaSR \citep{grayeli2024symbolicregressionlearnedconcept} & 100 original & 72/100 \\
SR-LLM \citep{guo2025srllmrag} & 90 held-out & 69/90 \\
QDSR \citep{bruneton2025qdsr} & 117-target subset & 107/117 \\
\rowcolor{oursblue}
FunctionEvolve (Ours) & 100+20 full set & SA@1 120/120 \\
\bottomrule
\end{tabular}
\vspace{-0.8em}
\end{wraptable}

We also evaluate FunctionEvolve on the AI-Feynman benchmark family \citep{doi:10.1126/sciadv.aay2631}. Because recent papers use slightly different protocols, Table~\ref{tab:aifeynman_results} reports each result in its native denominator rather than forcing a single leaderboard: LaSR reports the 100 original Feynman equations, SR-LLM reserves 10 of those equations as retrieval knowledge and tests on the remaining 90, and QDSR uses a 117-target Feynman-AI subset. Under the full 120-task setting, consisting of the 100 original equations plus 20 bonus equations, FunctionEvolve recovers all 120 tasks at top-1.

Figure~\ref{fig:first_sa_round_clean} provides a first-appearance diagnostic for symbolic matches. A match at round 0 means that an equivalent expression was already present in the LLM-proposed seed set, before evolutionary mutation. The AI-Feynman row is much more concentrated at round 0 than the synthetic LLM-SRBench rows, which is consistent with literature-derived equations being easier for the model to recall or directly propose from prior exposure.

\begin{center}
\centering
\includegraphics[width=1.00\textwidth]{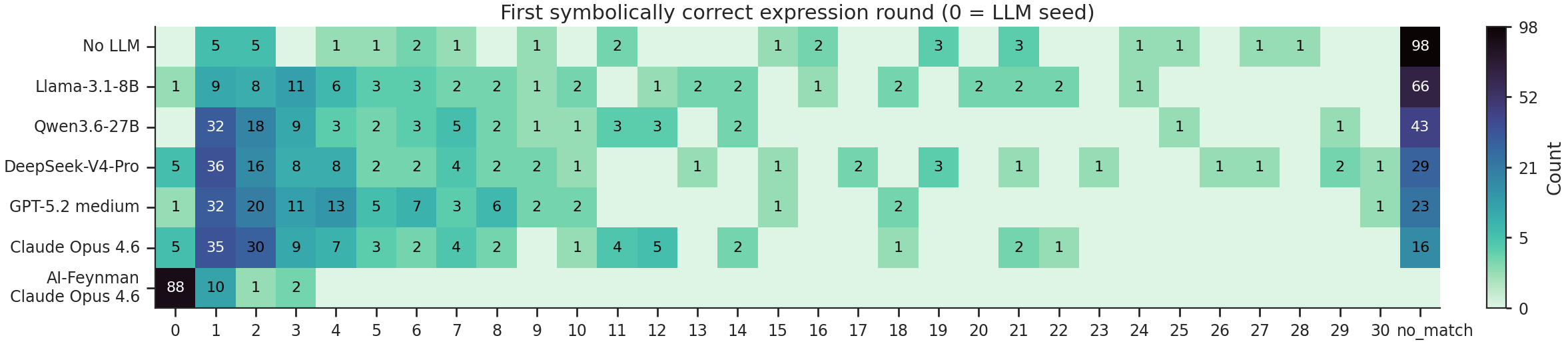}
\captionof{figure}{First-appearance round of the first symbolically correct expression for clean runs. Round 0 denotes a correct expression already present in the LLM-proposed seed set, while \texttt{no\_match} denotes runs without a symbolic match. The last row shows AI-Feynman results, while the others show LLM-SRBench results; concentration at round 0 evidences stronger AI-Feynman contamination.}
\label{fig:first_sa_round_clean}
\vspace{-0.7em}
\end{center}

These results are complementary to the LLM-SRBench evaluation rather than a replacement for it. AI-Feynman is a standard physics benchmark, but its literature-derived equations may be present in LLM pretraining data; LLM-SRBench therefore remains our primary lower-contamination test bed. The AI-Feynman result mainly shows that the same structure-guided search also performs strongly on the classical physics benchmark, including the harder bonus equations.

\vspace{-0.5em}
\subsection{Ablation Study}
\label{sec:ablation}
\vspace{-0.5em}

\begin{wraptable}{r}{0.47\linewidth}
\vspace{-1.3em}
\centering
\caption{Ablation study on 129 tasks. Entries report total $\mathrm{SA@50}$ ($\mathrm{SA@1}$). Domain-wise SA results are reported in Appendix~\ref{app:extended_ablation}.}
\vspace{-0.3em}
\label{tab:ablation}
\scriptsize
\setlength{\tabcolsep}{2pt}
\renewcommand{\arraystretch}{1.12}
\sffamily
\begin{tabular}{@{}lcc@{}}
\toprule
Method & \emph{GPT-5.2-medium} & \emph{Claude Opus 4.6} \\ 
\midrule
Full & \textbf{\tabnum{103}} (\textbf{\tabnum{69}}) & \textbf{\tabnum{107}} (\textbf{\tabnum{72}}) \\
w/o All & \tabnum{35} (\tabnum{6}) & \tabnum{34} (\tabnum{10}) \\
w/o Generator & \tabnum{79} (\tabnum{49}) & \tabnum{91} (\tabnum{58}) \\
w/o Selector & \tabnum{62} (\tabnum{44}) & \tabnum{74} (\tabnum{46}) \\
w/o LLM Mutator & \tabnum{45} (\tabnum{27}) & \tabnum{46} (\tabnum{31}) \\
w/o AST Mutator & \tabnum{70} (\tabnum{42}) & \tabnum{84} (\tabnum{50}) \\
w/o Struct. Optimizer & \tabnum{46} (\tabnum{16}) & \tabnum{53} (\tabnum{22}) \\
\bottomrule
\end{tabular}
\vspace{-1.6em}
\end{wraptable}

We further ablate the major components of FunctionEvolve using the degenerated variants described in Appendix~\ref{app:generator_details}--\ref{app:optimizer_details}. The \textit{w/o Generator} setting sets the extracted domain knowledge to empty and initializes the search from a fixed seed list. The \textit{w/o Selector} setting removes the LLM parent-selection module and instead uses Boltzmann sampling over fitness ranks. For mutation, we consider two complementary ablations: \textit{w/o LLM Mutator} disables LLM-generated local structural refinements and keeps only the AST Mutator, while \textit{w/o AST Mutator} disables rule-driven AST additions and deletions and keeps only LLM-generated ADD/SUBST edits. The \textit{w/o Structure-aware Optimizer} setting replaces the structure-aware coefficient optimizer with L-BFGS. Finally, \textit{w/o All} combines \textit{w/o Generator}, \textit{w/o Selector}, \textit{w/o AST Mutator}, and \textit{w/o Structure-aware Optimizer}. In both implementation and experimental behavior, this combined ablation is close to a parallel multi-offspring version of LLM-SR~\citep{shojaee2024llm}. Although this parallel setting evaluates more candidate expressions per round, the local evaluations are independent and can be distributed across CPU workers, so wall-clock cost does not increase in proportion to the candidate count; detailed resource comparisons are provided in Appendix~\ref{app:resource_usage}.

Table~\ref{tab:ablation} reports total $\mathrm{SA@50}$ ($\mathrm{SA@1}$) for \emph{GPT-5.2-medium} and \emph{Claude Opus 4.6}; domain-wise breakdowns are provided in Appendix~\ref{app:extended_ablation}. Under \emph{Claude Opus 4.6}, the combined ablation reduces SA from 107 (72) to 34 (10) tasks. Among components, removing the LLM Mutator and the structure-aware optimizer causes the largest $\mathrm{SA@50}$ drops, to 46 (31) and 53 (22), confirming that LLM-guided local refinement and reliable structure-aware coefficient fitting are both essential; removing the AST Mutator and Selector also degrades recovery, to 84 (50) and 74 (46), showing that the strongest performance comes from combining LLM reasoning with explicit symbolic structure and reliable coefficient scoring. We therefore first isolate the contribution of AST visibility, then examine whether candidate skeletons can be scored reliably by the structure-aware coefficient optimizer.

\begin{tcolorbox}[enhanced, sharp corners, boxrule=0.6pt, colback=oursblue!60!white, colframe=black!55, left=5pt, right=5pt, top=4pt, bottom=4pt, before skip=0.4em, after skip=0.4em]
\textbf{Ablation takeaway:} FunctionEvolve succeeds by combining explicit AST structure, reliable structure-aware optimization, and frontier-model symbolic proposals.
\end{tcolorbox}
\subsection{Analysis of AST Structure}
\label{sec:ast_structure_analysis}
To isolate the role of AST structure more directly, we use a stricter \textit{w/o AST Structure} variant under \emph{Claude Opus 4.6}. This variant keeps the LLM Selector and LLM Mutator, but removes all AST-related information and operations: the Selector no longer observes AST-derived structural fields, the AST Mutator is disabled, and the LLM Mutator no longer receives the annotated AST or AST-specific mutation guidance.

\begin{wraptable}[8]{r}{0.55\linewidth}
\vspace{-1.7em}
\centering
\caption{Effect of AST structure under \emph{Claude Opus 4.6}. Entries report $\mathrm{SA@50}$ ($\mathrm{SA@1}$).}
\label{tab:ast_structure_analysis}
\scriptsize
\setlength{\tabcolsep}{2pt}
\renewcommand{\arraystretch}{1.12}
\sffamily
\begin{tabular}{@{}lccccc@{}}
\toprule
Variant & Chemistry & Biology & Physics & Materials & Total \\
& (36) & (24) & (44) & (25) & (129) \\
\midrule
Full & \textbf{\tabnum{31}} (\textbf{\tabnum{16}}) & \textbf{\tabnum{22}} (\textbf{\tabnum{16}}) & \textbf{\tabnum{39}} (\textbf{\tabnum{34}}) & \textbf{\tabnum{15}} (\textbf{\tabnum{6}}) & \textbf{\tabnum{107}} (\textbf{\tabnum{72}}) \\
w/o AST Mutator & \tabnum{30} (\tabnum{16}) & \textbf{\tabnum{22}} (\tabnum{11}) & \tabnum{23} (\tabnum{20}) & \tabnum{9} (\tabnum{3}) & \tabnum{84} (\tabnum{50}) \\
\rowcolor{oursblue}
w/o AST Structure & \tabnum{24} (\tabnum{14}) & \tabnum{20} (\tabnum{15}) & \tabnum{8} (\tabnum{7}) & \tabnum{8} (\tabnum{4}) & \tabnum{60} (\tabnum{40}) \\
\bottomrule
\end{tabular}
\vspace{-0.5em}
\end{wraptable}

As shown in Table~\ref{tab:ast_structure_analysis}, removing all AST structure reduces total $\mathrm{SA@50}$ ($\mathrm{SA@1}$) from 107 (72) to 60 (40). This is substantially worse than only removing the AST Mutator, which still obtains 84 (50) when the LLM components can observe AST-derived structure. The gap indicates that AST structure is useful not only as a source of deterministic rule mutations, but also as an explicit representation that helps the LLM reason about expression complexity, reusable subtrees, and meaningful local edits. The degradation is especially severe on the Physics split, whose equations are more structurally complex in the benchmark.

\subsection{Analysis of Structure-aware Coefficient Optimization}
\label{sec:optimizer_pipeline_analysis}
\begin{center}
    \centering
    \includegraphics[width=1.0\linewidth]{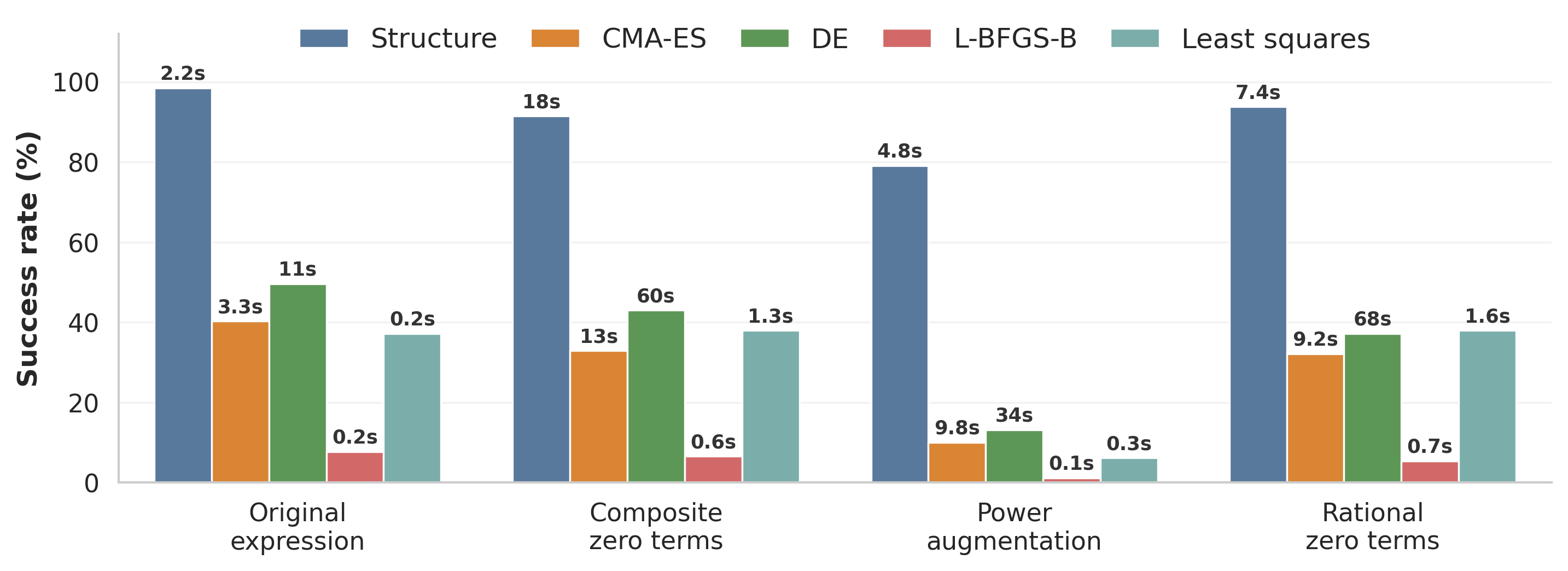}
    \captionof{figure}{Optimizer benchmark on ground-truth skeletons and transformed variants. Success is defined as $\mathrm{NMSE}<\text{1e-10}$; numbers above bars report median runtime.}
    \label{fig:optimizer_bench}
    \vspace{-0.5em}
\end{center}

The ablation in Section~\ref{sec:ablation} points to coefficient optimization as a major failure mode: under \emph{Claude Opus 4.6}, replacing the structure-aware coefficient optimizer with L-BFGS-B reduces total $\mathrm{SA@50}$ ($\mathrm{SA@1}$) from 107 (72) to 53 (22). To separate optimizer failure from symbolic-search failure, we benchmark optimizers on ground-truth skeletons and controlled variants (Appendix~\ref{app:optimizer_benchmark_variants}). A run is counted as successful only when the fitted expression reaches $\mathrm{NMSE}<\text{1e-10}$ on the training data. As shown in Figure~\ref{fig:optimizer_bench}, the structure-aware coefficient optimizer consistently outperforms standalone optimizers across all four transformation groups while keeping median runtime modest. It solves 98.4\% of original-expression cases, 91.5\% of composite zero-term cases, 79.1\% of power-augmentation cases, and 93.8\% of rational zero-term cases. This result explains why the optimizer ablation in Table~\ref{tab:ablation} is so large: lacking structural information, a generic optimizer must search a higher-dimensional and less constrained parameter space, making optimization less efficient and more susceptible to local optima. It can therefore underfit otherwise correct symbolic skeletons and create false negatives during fitness evaluation. Consistently, the No LLM row in Table~\ref{tab:main_results} shows pure GP mutation with the structure-aware optimizer already approaches PiT-PO SOTA.

\vspace{-0.6em}
\subsection{Analysis under Noisy Observations}
\label{sec:noise_analysis}
\vspace{-0.4em}

\begin{wraptable}{r}{0.45\linewidth}
\vspace{-1.5em}
\centering
\caption{Robustness under noisy observations with \emph{Claude Opus 4.6}. Entries report counts over 129 tasks.}
\label{tab:noise_analysis}
\scriptsize
\setlength{\tabcolsep}{3pt}
\renewcommand{\arraystretch}{1.12}
\sffamily
\begin{tabular}{@{}lccc@{}}
\toprule
Setting & $\mathrm{SA@50}$ ($\mathrm{SA@1}$) & $\mathrm{Acc}_{0.1}$ & Test NMSE \\
\midrule
Clean & \textbf{\tabnum{107}} (\textbf{\tabnum{72}}) & \textbf{\tabnum{113}} & \tabnum{1.38e-13} \\
1\% noise & \tabnum{54} (\tabnum{24}) & \tabnum{39} & \tabnum{1.32e-6} \\
5\% noise & \tabnum{40} (\tabnum{13}) & \tabnum{23} & \tabnum{3.05e-5} \\
\bottomrule
\end{tabular}
\vspace{-0.8em}
\end{wraptable}

We further evaluate FunctionEvolve on noisy versions of LLM-SRBench using the same \emph{Claude Opus 4.6} setting. As shown in Table~\ref{tab:noise_analysis}, exact symbolic recovery is sensitive to observation noise: total $\mathrm{SA@50}$ ($\mathrm{SA@1}$) drops from 107 (72) in the clean setting to 54 (24) under 1\% noise and 40 (13) under 5\% noise. The numerical criterion also degrades sharply, with test $\mathrm{Acc}_{0.1}$ decreasing from 113 tasks to 39 and 23 tasks, respectively. This behavior is expected because the search ranks candidates by noisy training fit while symbolic accuracy is judged against the clean ground-truth expression. Under noise, structurally different expressions can become numerically plausible explanations of the perturbed samples, so the evolutionary search may favor smooth or overfit alternatives that fit the observations but no longer match the exact law. These results suggest that noisy scientific equation discovery requires uncertainty-aware scoring and validation in addition to stronger symbolic search.

\vspace{-0.5em}
\subsection{Complexity-aware Final Candidate Selection}
\label{sec:complexity_aware_selection}
\vspace{-0.5em}

The gap between $\mathrm{SA@50}$ and $\mathrm{SA@1}$ indicates that the search often generates a correct expression but does not rank it first by training NMSE. We therefore examine complexity-aware final-candidate selectors for the \emph{Claude Opus 4.6} full run, with the three selection rules summarized in Figure~\ref{fig:heuristic_selection}. After search terminates, these selectors are applied to the complete expression trajectory to choose the final candidates to report. They are distinct from the evolutionary Selector in Section~\ref{sec:method}: they do not choose parents, guide mutation, affect coefficient fitting, or otherwise change the search trajectory.

\begin{wraptable}{r}{0.35\linewidth}
\vspace{-1.2em}
\centering
\caption{Final-candidate selection for the \emph{Claude Opus 4.6} full run on 129 clean LLM-SRBench tasks.}
\label{tab:complexity_aware_selection}
\scriptsize
\setlength{\tabcolsep}{3pt}
\renewcommand{\arraystretch}{1.12}
\sffamily
\begin{tabular}{@{}lcc@{}}
\toprule
Ranking rule & Shortlist & Exact matches \\
\midrule
Train NMSE & 1 & \tabnum{72} \\
Train NMSE & 5 & \tabnum{89} \\
Train NMSE & 10 & \tabnum{95} \\
Pareto & 5 & \tabnum{102} \\
Occam & 5 & \tabnum{101} \\
MDL & 5 & \tabnum{97} \\
Pareto & 10 & \tabnum{102} \\
Occam & 10 & \tabnum{104} \\
MDL & 10 & \tabnum{101} \\
Train NMSE & 50 & \tabnum{107} \\
Train NMSE & all\footnotemark & \tabnum{113} \\
\bottomrule
\end{tabular}
\vspace{-1.8em}
\end{wraptable}
\footnotetext{Here, ``all'' denotes $\mathrm{SA@all}$ over roughly the top 1000 full-trajectory candidates ranked by training NMSE.}

The final-candidate selectors use only training NMSE and expression complexity, never test or OOD NMSE. Pareto performs non-dominated sorting over training NMSE and complexity and fills the reporting budget from successive Pareto layers. The Occam variant first restricts candidates to a near-best training-NMSE neighborhood and then favors simpler expressions using tree size, operator count, parameter count, and special-function count. The MDL variant uses a linear penalty on training loss and complexity features. As shown in Table~\ref{tab:complexity_aware_selection}, the five-candidate Pareto and Occam selections recover 102 and 101 exact forms, already above the 95 exact forms obtained by the top-10 training-NMSE ranking and close to the top-50 result of 107. The approximate $\mathrm{SA@all}$ value is 113, showing that a few additional exact forms appear deeper in the search trajectory. This suggests that many apparent top-1 failures are final-ranking failures caused by numerically strong but structurally less concise surrogate expressions. However, a gap remains to $\mathrm{SA@all}$, indicating that complexity-aware ranking still cannot always identify the correct form within a very small reporting budget.

\begin{center}
  \centering
  \includegraphics[width=1.00\textwidth]{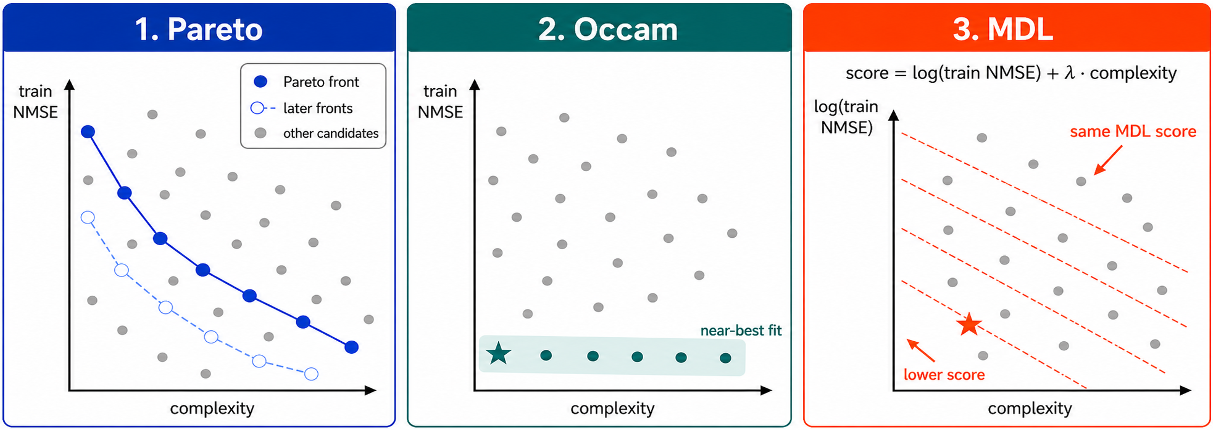}
  \vspace{-1.0em}
  \captionof{figure}{Complexity-aware final-candidate selection rules. Pareto treats each candidate as a point in training-NMSE--complexity space and selects from successive non-dominated fronts; Occam first keeps candidates with near-best training fit and then prefers simpler expressions; MDL combines training loss and complexity into a single scalar score. Details are provided in Appendix~\ref{app:complexity_aware_final_selection}.}
  \label{fig:heuristic_selection}
  \vspace{-0.6em}
\end{center}

\vspace{-0.8em}
\section{Discussion}
\label{sec:discussion}
\vspace{-0.5em}

\begin{wrapfigure}[18]{r}{0.50\linewidth}
\vspace{-1.5em}
\centering
\includegraphics[width=\linewidth]{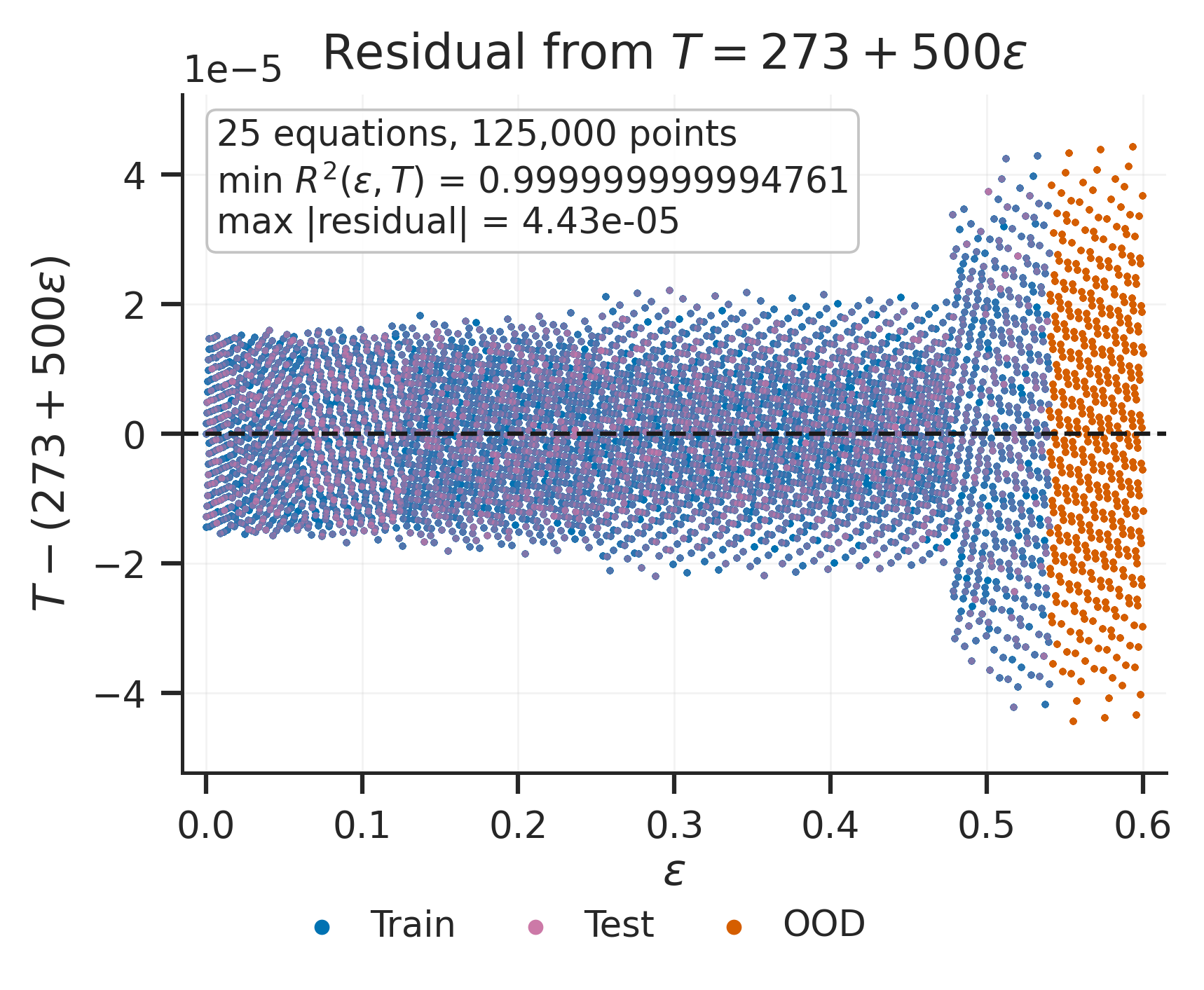}
\vspace{-2.2em}
\caption{Residuals from the MatSci. Across train, test, and OOD samples, the maximum absolute residual is $4.43\times 10^{-5}$.}
\label{fig:matsci_input_degeneracy}
\end{wrapfigure}

To diagnose weaker MatSci results in Section~\ref{experiments}, we identify two severe dataset identifiability issues. First, all 25 tasks use strain $\epsilon$ and temperature $T$, but the samples collapse to one dimension: across train, test, and OOD splits, $T \approx 273 + 500\epsilon$. Fitting $T=a+b\epsilon$ for each task and split gives correlations within $2.7\times10^{-12}$ of one and a maximum absolute residual of $4.43\times 10^{-5}$. This collinearity makes explicit temperature coupling unidentifiabl, so structurally different expressions can achieve identical NMSE by replacing temperature-coupled terms with functions of $\epsilon$ alone. Figure~\ref{fig:matsci_input_degeneracy} shows the residuals from this near-linear geometry. Second, Arrhenius-like factors vary too little to be reliably identified. Eight MatSci equations contain $\exp(-c/T)$ terms, but these factors change by $0.11\%$--$2.06\%$ on train/test and $0.006\%$--$0.108\%$ on OOD. In MatSci5, $\exp(-5.9485/T)$ changes from $0.978446$ to $0.989105$ on train and from $0.989106$ to $0.989672$ on OOD. Hence $\epsilon^p\exp(-c/T)$ behaves like a rescaled pure-strain term $C\epsilon^p$, making the Arrhenius structure difficult to distinguish from a constant multiplier.

In other dataset splits, the collinear issue is not dataset-wide as in MatSci, but individual tasks still exhibit strong pairwise dependence. Appendix~\ref{app:non_matsci_failure_correlation} reports the 12 FunctionEvolve failure cases outside MatSci and shows that most still contain a variable pair with large $R^2$. These dependencies can again weaken structural identifiability, because different symbolic mechanisms may become difficult to distinguish over the sampled trajectory even when the overall split is not globally degenerate.

Although our audit highlights limitations in LLM-SRBench and need for more reliable datasets, it remains the latest synthetic SR benchmark and reduces memorization risk relative to literature-derived benchmarks like AIFeynman\citep{doi:10.1126/sciadv.aay2631}; constructing a new benchmark is beyond the scope of this work.
\vspace{-0.5em}
\section{Conclusion}
\label{sec:conclusion}
\vspace{-0.5em}

We introduced FunctionEvolve, an LLM-driven SR framework combining semantic initialization, diversity-aware selection, local mutation by AST rule and LLM proposal, and structure-aware coefficient optimization. On the 129-task synthetic subset of LLM-SRBench, FunctionEvolve recovers 107 exact symbolic forms with \emph{Claude Opus 4.6}, substantially improving symbolic accuracy while preserving near-GT numerical precision; on the complementary 120-task AI-Feynman evaluation, it recovers the correct symbolic form for all 120 tasks as its top-1 prediction. Our ablations explain why: LLM-extracted domain priors provide semantic guidance that steers the search toward plausible expressions, explicit AST structure organizes symbolic exploration, and structure-aware optimization prevents promising skeletons from being discarded due to poorly fitted constants.

Beyond these results, our audit identifies critical limitations in LLM-SRBench itself: the MatSci split has severe collinearity between $T$ and $\epsilon$, making key mechanisms unidentifiable and allowing structurally incorrect expressions to attain very low NMSE. Taken together, these findings suggest that progress in scientific SR requires joint attention to symbolic representation, coefficient fitting, and benchmark design. Open directions include dynamical systems without closed forms, larger candidate-variable sets, and benchmarks that are both reliable and realistic.

\bibliographystyle{plainnat} 
\bibliography{references} 

\begin{thebibliography}{31}
\providecommand{\natexlab}[1]{#1}
\providecommand{\url}[1]{\texttt{#1}}
\expandafter\ifx\csname urlstyle\endcsname\relax
  \providecommand{\doi}[1]{doi: #1}\else
  \providecommand{\doi}{doi: \begingroup \urlstyle{rm}\Url}\fi

\bibitem[Bruneton(2025)]{bruneton2025qdsr}
Jean-Philippe Bruneton.
\newblock Enhancing symbolic regression with quality-diversity and physics-inspired constraints, 2025.
\newblock URL \url{https://arxiv.org/abs/2503.19043}.

\bibitem[Burlacu et~al.(2023)Burlacu, Yang, and Affenzeller]{burlacu2023populationdiversity}
Bogdan Burlacu, Kaifeng Yang, and Michael Affenzeller.
\newblock Population diversity and inheritance in genetic programming for symbolic regression.
\newblock \emph{Natural Computing}, 23, 01 2023.
\newblock \doi{10.1007/s11047-022-09934-x}.

\bibitem[Cava et~al.(2021)Cava, Orzechowski, Burlacu, de~França, Virgolin, Jin, Kommenda, and Moore]{lacava2021contemporarysymbolicregressionmethods}
William~La Cava, Patryk Orzechowski, Bogdan Burlacu, Fabrício~Olivetti de~França, Marco Virgolin, Ying Jin, Michael Kommenda, and Jason~H. Moore.
\newblock Contemporary symbolic regression methods and their relative performance, 2021.
\newblock URL \url{https://arxiv.org/abs/2107.14351}.

\bibitem[Cranmer(2023)]{cranmer2023interpretablemachinelearningscience}
Miles Cranmer.
\newblock Interpretable machine learning for science with {PySR} and {SymbolicRegression.jl}, 2023.
\newblock URL \url{https://arxiv.org/abs/2305.01582}.

\bibitem[D'haeseleer(1994)]{dhaesleer1994contextpreserving}
Patrik D'haeseleer.
\newblock Context preserving crossover in genetic programming.
\newblock 1:\penalty0 256 -- 261 vol.1, 07 1994.
\newblock \doi{10.1109/ICEC.1994.350006}.

\bibitem[dos Reis et~al.(2024)dos Reis, Caminha, and Penna]{reis2024benchmarkingsymbolicregressionconstant}
L.~G.~A dos Reis, V.~L. P.~S. Caminha, and T.~J.~P. Penna.
\newblock Benchmarking symbolic regression constant optimization schemes, 2024.
\newblock URL \url{https://arxiv.org/abs/2412.02126}.

\bibitem[Du et~al.(2024)Du, Chen, Wang, Nie, and Zhang]{du2024llm4edlargelanguagemodels}
Mengge Du, Yuntian Chen, Zhongzheng Wang, Longfeng Nie, and Dongxiao Zhang.
\newblock {LLM4Ed}: Large language models for automatic equation discovery, 2024.
\newblock URL \url{https://arxiv.org/abs/2405.07761}.

\bibitem[Grayeli et~al.(2024)Grayeli, Sehgal, Costilla-Reyes, Cranmer, and Chaudhuri]{grayeli2024symbolicregressionlearnedconcept}
Arya Grayeli, Atharva Sehgal, Omar Costilla-Reyes, Miles Cranmer, and Swarat Chaudhuri.
\newblock Symbolic regression with a learned concept library, 2024.
\newblock URL \url{https://arxiv.org/abs/2409.09359}.

\bibitem[Guo et~al.(2025)Guo, Wang, Tian, Yang, Yu, Na, Kovacs, Li, Ioannou, and Wang]{guo2025srllmrag}
Zelin Guo, Siqi Wang, Yonglin Tian, Jing Yang, Hui Yu, Xiaoxiang Na, Levente Kovacs, Li~Li, Petros~A. Ioannou, and Fei-Yue Wang.
\newblock {SR-LLM}: An incremental symbolic regression framework driven by {LLM}-based retrieval-augmented generation.
\newblock \emph{Proceedings of the National Academy of Sciences}, 122\penalty0 (52):\penalty0 e2516995122, 2025.
\newblock \doi{10.1073/pnas.2516995122}.

\bibitem[Kommenda et~al.(2020)Kommenda, Burlacu, Kronberger, and Affenzeller]{kommenda2020parameteridentification}
Michael Kommenda, Bogdan Burlacu, Gabriel Kronberger, and Michael Affenzeller.
\newblock Parameter identification for symbolic regression using nonlinear least squares.
\newblock \emph{Genetic Programming and Evolvable Machines}, 21, 09 2020.
\newblock \doi{10.1007/s10710-019-09371-3}.

\bibitem[Koza(1994)]{koza_genetic_1994}
John~R. Koza.
\newblock Genetic programming as a means for programming computers by natural selection.
\newblock \emph{Statistics and Computing}, 4\penalty0 (2):\penalty0 87--112, June 1994.
\newblock ISSN 1573-1375.
\newblock \doi{10.1007/BF00175355}.
\newblock URL \url{https://doi.org/10.1007/BF00175355}.

\bibitem[Kramer et~al.(2023)Kramer, Cerrato, Brugger, Džeroski, and King]{kramer2023automatedscientificdiscoveryequation}
Stefan Kramer, Mattia Cerrato, Jannis Brugger, Sašo Džeroski, and Ross King.
\newblock Automated scientific discovery: From equation discovery to autonomous discovery systems, 2023.
\newblock URL \url{https://arxiv.org/abs/2305.02251}.

\bibitem[Lange et~al.(2025)Lange, Imajuku, and Cetin]{lange2025shinkaevolveopenendedsampleefficientprogram}
Robert~Tjarko Lange, Yuki Imajuku, and Edoardo Cetin.
\newblock {ShinkaEvolve}: Towards open-ended and sample-efficient program evolution, 2025.
\newblock URL \url{https://arxiv.org/abs/2509.19349}.

\bibitem[Langley(1981)]{Langley1981}
Pat Langley.
\newblock Data-driven discovery of physical laws.
\newblock \emph{Cognitive Science}, 5\penalty0 (1):\penalty0 31--54, 1981.
\newblock \doi{https://doi.org/10.1111/j.1551-6708.1981.tb00869.x}.
\newblock URL \url{https://onlinelibrary.wiley.com/doi/abs/10.1111/j.1551-6708.1981.tb00869.x}.

\bibitem[Lu et~al.(2015)Lu, Ren, and Wang]{lu2015usinggeneticprogramming}
Qiang Lu, Jun Ren, and Zhiguang Wang.
\newblock Using genetic programming with prior formula knowledge to solve symbolic regression problem.
\newblock \emph{Computational Intelligence and Neuroscience}, 2016:\penalty0 1--17, 12 2015.
\newblock \doi{10.1155/2016/1021378}.

\bibitem[Martius and Lampert(2016)]{martius2016extrapolationlearningequations}
Georg Martius and Christoph~H. Lampert.
\newblock Extrapolation and learning equations, 2016.
\newblock URL \url{https://arxiv.org/abs/1610.02995}.

\bibitem[Montana(1995)]{montana:stgpEC}
David~J. Montana.
\newblock Strongly typed genetic programming.
\newblock \emph{Evolutionary Computation}, 3\penalty0 (2):\penalty0 199--230, Summer 1995.
\newblock ISSN 1063-6560.
\newblock \doi{10.1162/evco.1995.3.2.199}.
\newblock URL \url{http://vishnu.bbn.com/papers/stgp.pdf}.

\bibitem[Mundhenk et~al.(2021)Mundhenk, Landajuela, Glatt, Santiago, Faissol, and Petersen]{mundhenk2021symbolicregressionneuralguidedgenetic}
T.~Nathan Mundhenk, Mikel Landajuela, Ruben Glatt, Claudio~P. Santiago, Daniel~M. Faissol, and Brenden~K. Petersen.
\newblock Symbolic regression via neural-guided genetic programming population seeding, 2021.
\newblock URL \url{https://arxiv.org/abs/2111.00053}.

\bibitem[Novikov et~al.(2025)Novikov, Vũ, Eisenberger, Dupont, Huang, Wagner, Shirobokov, Kozlovskii, Ruiz, Mehrabian, Kumar, See, Chaudhuri, Holland, Davies, Nowozin, Kohli, and Balog]{novikov2025alphaevolvecodingagentscientific}
Alexander Novikov, Ngân Vũ, Marvin Eisenberger, Emilien Dupont, Po-Sen Huang, Adam~Zsolt Wagner, Sergey Shirobokov, Borislav Kozlovskii, Francisco J.~R. Ruiz, Abbas Mehrabian, M.~Pawan Kumar, Abigail See, Swarat Chaudhuri, George Holland, Alex Davies, Sebastian Nowozin, Pushmeet Kohli, and Matej Balog.
\newblock {AlphaEvolve}: A coding agent for scientific and algorithmic discovery, 2025.
\newblock URL \url{https://arxiv.org/abs/2506.13131}.

\bibitem[Petersen et~al.(2021)Petersen, Landajuela, Mundhenk, Santiago, Kim, and Kim]{petersen2021deepsymbolicregressionrecovering}
Brenden~K. Petersen, Mikel Landajuela, T.~Nathan Mundhenk, Claudio~P. Santiago, Soo~K. Kim, and Joanne~T. Kim.
\newblock Deep symbolic regression: Recovering mathematical expressions from data via risk-seeking policy gradients, 2021.
\newblock URL \url{https://arxiv.org/abs/1912.04871}.

\bibitem[Sahoo et~al.(2018)Sahoo, Lampert, and Martius]{sahoo2018learningequationsextrapolationcontrol}
Subham~S. Sahoo, Christoph~H. Lampert, and Georg Martius.
\newblock Learning equations for extrapolation and control, 2018.
\newblock URL \url{https://arxiv.org/abs/1806.07259}.

\bibitem[Schmidt and Lipson(2009)]{schmidt2009distilling}
Michael Schmidt and Hod Lipson.
\newblock Distilling free-form natural laws from experimental data.
\newblock \emph{Science}, 324\penalty0 (5923):\penalty0 81--85, 2009.

\bibitem[Sharma(2025)]{openevolve}
Asankhaya Sharma.
\newblock {OpenEvolve}: an open-source evolutionary coding agent, 2025.
\newblock URL \url{https://github.com/algorithmicsuperintelligence/openevolve}.

\bibitem[Shojaee et~al.(2024)Shojaee, Meidani, Gupta, Farimani, and Reddy]{shojaee2024llm}
Parshin Shojaee, Kazem Meidani, Shashank Gupta, Amir~Barati Farimani, and Chandan~K Reddy.
\newblock {LLM-SR}: Scientific equation discovery via programming with large language models.
\newblock \emph{arXiv preprint arXiv:2404.18400}, 2024.

\bibitem[Shojaee et~al.(2025)Shojaee, Nguyen, Meidani, Farimani, Doan, and Reddy]{shojaee2025llmsrbenchnewbenchmarkscientific}
Parshin Shojaee, Ngoc-Hieu Nguyen, Kazem Meidani, Amir~Barati Farimani, Khoa~D Doan, and Chandan~K Reddy.
\newblock {LLM-SRBench}: A new benchmark for scientific equation discovery with large language models, 2025.
\newblock URL \url{https://arxiv.org/abs/2504.10415}.

\bibitem[Udrescu and Tegmark(2020)]{doi:10.1126/sciadv.aay2631}
Silviu-Marian Udrescu and Max Tegmark.
\newblock {AI} {Feynman}: A physics-inspired method for symbolic regression.
\newblock \emph{Science Advances}, 6\penalty0 (16):\penalty0 eaay2631, 2020.
\newblock \doi{10.1126/sciadv.aay2631}.
\newblock URL \url{https://www.science.org/doi/abs/10.1126/sciadv.aay2631}.

\bibitem[Virgolin and Pissis(2022)]{virgolin2022symbolicregressionnphard}
Marco Virgolin and Solon~P. Pissis.
\newblock Symbolic regression is {NP}-hard, 2022.
\newblock URL \url{https://arxiv.org/abs/2207.01018}.

\bibitem[Wang et~al.(2026)Wang, Li, Liu, Li, Wang, Zhang, and Cheng]{wang2026llmbasedscientificequationdiscovery}
Boxiao Wang, Kai Li, Tianyi Liu, Chen Li, Junzhe Wang, Yifan Zhang, and Jian Cheng.
\newblock {LLM}-based scientific equation discovery via physics-informed token-regularized policy optimization, 2026.
\newblock URL \url{https://arxiv.org/abs/2602.10576}.

\bibitem[Wang et~al.(2025)Wang, Wang, Li, Zhang, and Cheng]{wang2025drsrllmbasedscientific}
Runxiang Wang, Boxiao Wang, Kai Li, Yifan Zhang, and Jian Cheng.
\newblock {DRSR}: {LLM}-based scientific equation discovery with dual reasoning from data and experience, 2025.
\newblock URL \url{https://arxiv.org/abs/2506.04282}.

\bibitem[Whigham(1995)]{whigham:1995:GBGP}
P.~A. Whigham.
\newblock Grammatically-based genetic programming.
\newblock In Justinian~P. Rosca, editor, \emph{Proceedings of the Workshop on Genetic Programming: From Theory to Real-World Applications}, pages 33--41, Tahoe City, California, USA, 9 July 1995.
\newblock URL \url{http://divcom.otago.ac.nz/sirc/Peterw/Publications/ml95.zip}.

\bibitem[Xia et~al.(2025)Xia, Sun, and Liu]{xia2025srscientist}
Shijie Xia, Yuhan Sun, and Pengfei Liu.
\newblock {SR-Scientist}: Scientific equation discovery with agentic {AI}, 2025.
\newblock URL \url{https://arxiv.org/abs/2510.11661}.

\end{thebibliography}


\appendix
\makeatletter
\@addtoreset{table}{section}
\makeatother
\renewcommand{\thetable}{\Alph{section}\arabic{table}}
\renewcommand{\thesubsection}{\Alph{section}\arabic{subsection}}

\section{Code and Data Availability}
\label{app:code_data_availability}

We provide an anonymized repository containing the FunctionEvolve implementation, experiment configuration files, prompt templates, scripts for running the main experiments and baselines, and raw result files used to produce the tables and figures in this paper. The repository also includes environment setup instructions, benchmark data access and preparation instructions, and command-line examples for reproducing the main LLM-SRBench experiments and ablations:
\url{https://github.com/Phoinikas03/FunctionEvolve}.

The existing benchmark and baseline assets used in our experiments are credited in the main text and are publicly available from their official repositories. LLM-SRBench and its associated LLM-SR benchmark code are available under the MIT license at
\url{https://github.com/deep-symbolic-mathematics/llm-srbench};
OpenEvolve is available under the Apache-2.0 license at
\url{https://github.com/algorithmicsuperintelligence/openevolve}.
The AI-Feynman benchmark is available from the official download page:
\url{https://space.mit.edu/home/tegmark/aifeynman.html}.

The main LLM-SRBench evaluation uses the 129-task synthetic subset, covering four scientific domains: biology/population growth, chemistry/reaction kinetics, physics/oscillation, and materials science. For the AI-Feynman evaluation in Section~\ref{sec:aifeynman}, we use the full benchmark family consisting of the 100 original Feynman equations and the 20 bonus equations, for a total of 120 tasks. Symbolic matches are judged against the benchmark target expressions over the benchmark-specified variable domains. Thus, domain-restricted identities are counted as matches when the candidate and target agree throughout the stated domain; equivalence outside that domain is not required.

\section{Resource Usage}
\label{app:resource_usage}


The main LLM-SRBench comparisons, ablations, and rerun baselines use the same 129-task synthetic subset. For the baseline reruns, we follow the public LLM-SR and OpenEvolve implementations and keep their released search and evaluator hyperparameters unchanged, rather than tuning them for FunctionEvolve. Concretely, FunctionEvolve runs for at most 30 evolution iterations per task, OpenEvolve runs for 200 iterations per task, and LLM-SR runs with a 1000-sample budget per task.

\subsection{LLM Usage}
\label{app:resource_llm_usage}

\paragraph{LLM call accounting.}
Both the \emph{Claude Opus 4.6} and \emph{GPT-5.2-medium} experiments are run through external LLM APIs. Unless otherwise stated, the accounting below uses the complete full-setting \emph{Claude Opus 4.6} logs over the 129 LLM-SRBench tasks.

In each FunctionEvolve evolution iteration, the Selector chooses 5 candidate expressions as parents, and the LLM Mutator proposes 20 new candidate expressions for each selected parent. Search terminates early once 50 mature expressions are found, where maturity is defined by training NMSE falling below the task-specific threshold. Retries are used only when parsing or validation fails. 

\begin{figure}[t]
\centering
\includegraphics[width=0.88\linewidth]{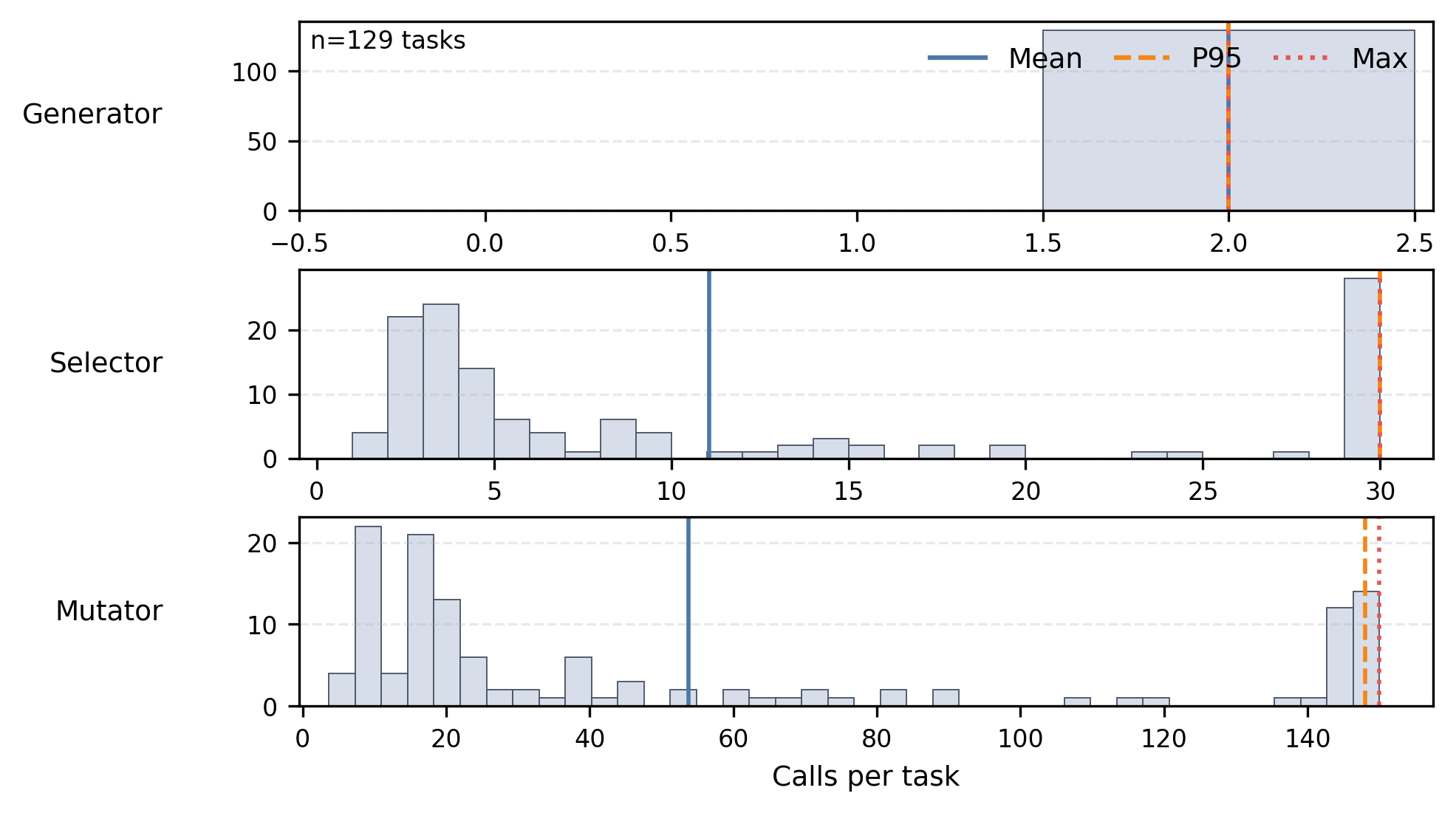}
\caption{Component-wise distributions of LLM call counts over 129 complete task logs from the full \emph{Claude Opus 4.6} FunctionEvolve experiment. Vertical lines mark the mean, 95th percentile, and maximum per-task call count for each main search component.}
\label{fig:llm_call_usage_by_component}
\end{figure}

For the baselines, LLM-SR generates one new candidate expression per sample, and OpenEvolve generates one new child program per iteration. Table~\ref{tab:llm_call_budget} reports the theoretical maximum number of LLM calls; actual usage can differ slightly because of early stopping, retry, or logging behavior. Figure~\ref{fig:llm_call_usage_by_component} and Table~\ref{tab:llm_usage_observed} summarize the observed call counts.

The observed call counts show that FunctionEvolve uses substantially fewer LLM requests than the baseline search loops. Although its theoretical per-task budget is 182 calls, early stopping reduces the average to 66.86 calls per task. Most of these calls are spent on the LLM Mutator, which averages 53.80 calls per task, while the Selector averages 11.06 calls and the Generator is fixed at 2 calls. In contrast, LLM-SR stays close to its 1000-iteration budget, and OpenEvolve stays close to its 200-iteration budget. Thus, FunctionEvolve reaches its main results with about 15x fewer LLM calls than LLM-SR and about 3x fewer calls than OpenEvolve on average.

\begin{center}
\centering
\captionof{table}{Theoretical maximum per-task LLM-call budget for FunctionEvolve and the LLM-driven baselines.}
\label{tab:llm_call_budget}
\small
\setlength{\tabcolsep}{4pt}
\begin{tabular}{@{}p{0.24\linewidth}p{0.34\linewidth}p{0.32\linewidth}@{}}
\toprule
Method/component & Trigger & Maximum call budget per task \\
\midrule
\multicolumn{3}{@{}l}{\textbf{FunctionEvolve}} \\
Generator & Domain-prior extraction and seed initialization & 2 calls \\
Selector & Once per evolution round & $\leq 30$ calls \\
LLM Mutator & Once per selected parent & $\leq 30 \times 5 = 150$ calls \\
\midrule
LLM-SR & One child hypothesis per iteration & 1000 calls \\
OpenEvolve & One child hypothesis per iteration & 200 calls \\
\bottomrule
\end{tabular}
\end{center}

\begin{center}
\centering
\captionof{table}{LLM call usage on the 129-task LLM-SRBench subset under \emph{Claude Opus 4.6}.}
\label{tab:llm_usage_observed}
\small
\setlength{\tabcolsep}{4pt}
\begin{tabular}{@{}lrrrrrr@{}}
\toprule
Method/component & Total calls & Mean/task & Std./task & Median & P95 & Max \\
\midrule
\textbf{FunctionEvolve} & 8625 & 66.86 & 64.88 & 30 & 180 & 182 \\
\quad Generator & 258 & 2.00 & 0.00 & 2 & 2 & 2 \\
\quad Selector & 1427 & 11.06 & 11.08 & 5 & 30 & 30 \\
\quad LLM Mutator & 6940 & 53.80 & 53.80 & 23 & 148 & 150 \\
\midrule
LLM-SR & 129035 & 1000.27 & 3.07 & 1001 & 1001 & 1001 \\
OpenEvolve & 25994 & 201.50 & 2.11 & 200 & 205 & 206 \\
\bottomrule
\end{tabular}
\end{center}

\begin{figure}[t]
\centering
\includegraphics[width=0.96\linewidth]{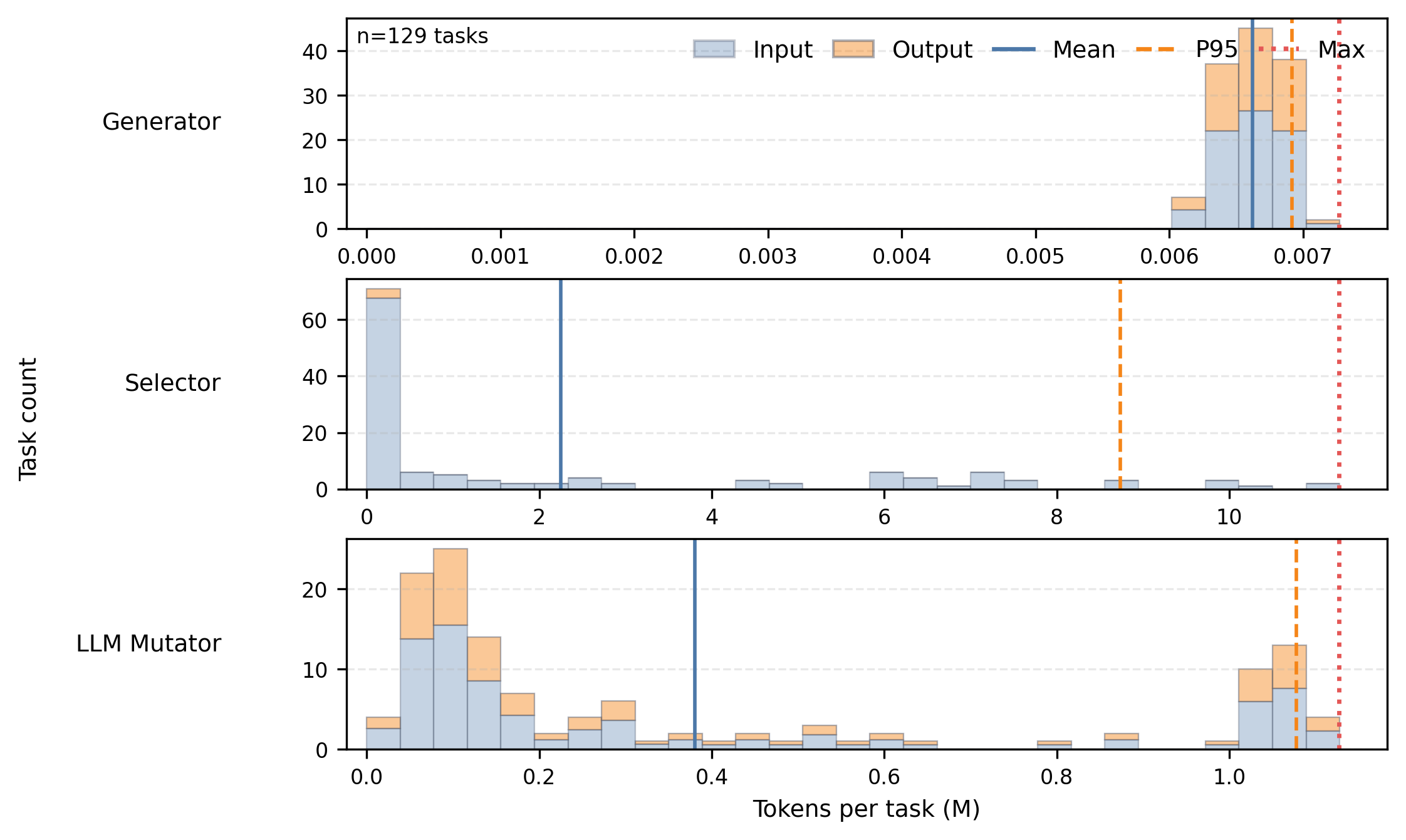}
\caption{Per-task token-usage distributions for the three main FunctionEvolve LLM components under \emph{Claude Opus 4.6}. Each row is a histogram over the 129 tasks; each bin is split into input-token and output-token contributions. Vertical lines mark the mean, 95th percentile, and maximum total tokens per task for the corresponding component.}
\label{fig:functionevolve_component_tokens_per_task}
\end{figure}

\paragraph{LLM token accounting.}
We follow the same accounting protocol as the LLM call accounting. Table~\ref{tab:llm_token_usage_observed} reports the aggregate prompt, completion, and total token counts, and Figure~\ref{fig:functionevolve_component_tokens_per_task} shows the per-task token distribution for the three FunctionEvolve LLM components, separating input and output tokens.

The token accounting shows a different tradeoff from the call accounting. FunctionEvolve uses fewer LLM calls than both LLM-SR and OpenEvolve, but each call is longer because the Selector observes the evolving tree summary and the LLM Mutator receives rich structural context. As a result, FunctionEvolve uses 340.36M total tokens, compared with 262.16M for LLM-SR and 244.79M for OpenEvolve. Within FunctionEvolve, token usage is highly concentrated in the Selector: it accounts for 290.44M tokens, or 85.33\% of the FunctionEvolve total, while the LLM Mutator accounts for 49.07M tokens and the Generator for only 0.85M. Figure~\ref{fig:functionevolve_component_tokens_per_task} shows that per-task token usage varies mainly in the Selector and LLM Mutator, while the Generator remains nearly constant across tasks because it is used only for initialization. Since prompt tokens are much cheaper than completion tokens in typical LLM pricing, we also report a price-weighted cost proxy in Table~\ref{tab:llm_token_usage_observed}, computed as prompt tokens$/5$ plus completion tokens. Under this weighting, FunctionEvolve costs 85.56M output-token equivalents, lower than LLM-SR's 143.45M and comparable to OpenEvolve's 68.13M despite using more total tokens.

\begin{center}
\centering
\captionof{table}{Observed token usage on the 129-task LLM-SRBench subset under \emph{Claude Opus 4.6}. Token counts and price-weighted costs are reported in millions. The weighted-cost column assumes one prompt token costs one fifth of one completion token.}
\label{tab:llm_token_usage_observed}
\small
\setlength{\tabcolsep}{5pt}
\begin{tabular}{@{}lrrrr@{}}
\toprule
Method/component & Prompt tokens & Completion tokens & Total tokens & Weighted cost \\
\midrule
\textbf{FunctionEvolve} & 318.50 & 21.86 & 340.36 & 85.56 \\
\quad Generator & 0.50 & 0.35 & 0.85 & 0.45 \\
\quad Selector & 288.86 & 1.58 & 290.44 & 59.35 \\
\quad LLM Mutator & 29.14 & 19.93 & 49.07 & 25.76 \\
\midrule
LLM-SR & 148.40 & 113.77 & 262.16 & 143.45 \\
OpenEvolve & 220.83 & 23.96 & 244.79 & 68.13 \\
\bottomrule
\end{tabular}
\end{center}

\subsection{CPU Usage}
\label{app:resource_cpu_usage}

All local computation was performed on modern multicore CPU servers with heterogeneous configurations, each providing at least 64 logical CPUs. The local CPU workload consists primarily of candidate validation, program or expression execution, coefficient optimization, and post-fit screening. The existing logs do not record process-level user/system CPU time or instantaneous utilization; therefore, we report observed wall-clock time, evaluator parallelism, candidate-evaluation counts, and timeout behavior rather than exact CPU-hours.

This hardware lower bound is important for interpreting wall-clock cost. The original LLM-SR loop generates one offspring from one parent at each iteration, and OpenEvolve similarly evaluates one child program per iteration. FunctionEvolve instead selects multiple parents and expands them into a batch of candidate expressions. Although the full setting evaluates 126.09 offspring per round on average, these evaluations are independent and can be distributed across modern multicore CPUs; consequently, wall-clock time does not increase in proportion to the number of candidates, while one-offspring baselines leave much of this parallelism unused.

\begin{figure}[t]
\centering
\includegraphics[width=0.82\linewidth]{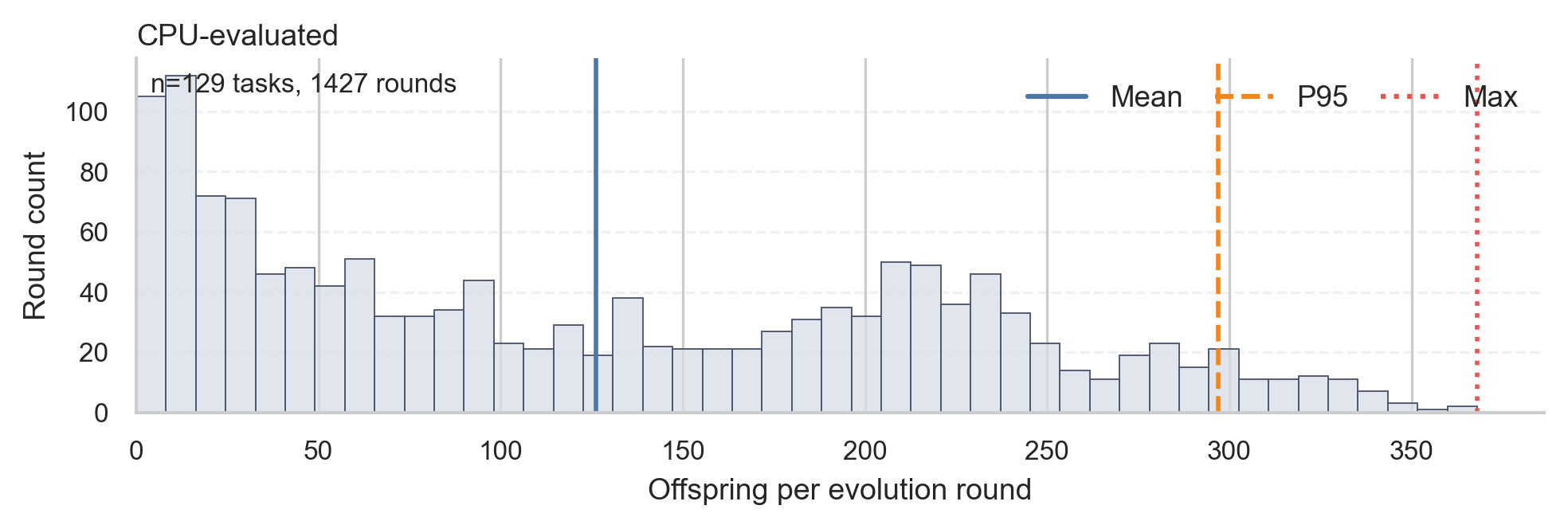}
\caption{Distribution of CPU-evaluated offspring counts per FunctionEvolve round over 129 complete full-run task logs. Counts include candidates that remain after structural deduplication and parameter-count filtering and therefore enter coefficient optimization. Vertical lines mark the mean, 95th percentile, and maximum.}
\label{fig:cpu_evaluated_offspring_per_round_distribution}
\end{figure}

Table~\ref{tab:cpu_evaluation_protocol} summarizes the local CPU-evaluation protocol. In FunctionEvolve, the 100 LLM-proposed offspring in each round come from 5 LLM Mutator calls over 5 selected parents, and deterministic rule mutations add further non-LLM offspring. The resulting batch-size distribution is shown in Figure~\ref{fig:cpu_evaluated_offspring_per_round_distribution}. Using the most conservative 64-logical-CPU reference, the mean batch size corresponds to roughly $126.09/64=1.97$ candidate-equivalent parallel waves, making the local wall-clock evaluation load comparable to the one-offspring baseline steps up to scheduling and optimizer-runtime variation.

\begin{table}[t]
\centering
\caption{CPU-evaluation protocol for FunctionEvolve and the LLM-driven baselines. Retry counts refer to local evaluator retries, not LLM API retries.}
\label{tab:cpu_evaluation_protocol}
\small
\setlength{\tabcolsep}{3pt}
\renewcommand{\arraystretch}{1.12}
\begin{tabular}{@{}p{0.17\linewidth}p{0.31\linewidth}p{0.26\linewidth}p{0.11\linewidth}p{0.10\linewidth}@{}}
\toprule
Method & Offspring per round & Evaluator & Timeout & Retries \\
\midrule
FunctionEvolve & 100 LLM proposals + rule mutations (mean: 126.09) & Structure-aware coefficient optimizer & 120 s & 0 \\
LLM-SR & 1 candidate per sample & BFGS optimizer & 30 s & 0 \\
OpenEvolve & 1 child program per iteration & BFGS optimizer & 90 s & 3\textsuperscript{*} \\
\bottomrule
\end{tabular}
\par\vspace{0.25em}
\footnotesize{\textsuperscript{*}OpenEvolve retries evaluation exceptions up to three times; evaluator timeouts are returned directly as timeout results.}
\end{table}


\subsection{GPU Usage}
\label{app:resource_gpu_usage}

GPU resources were used only for local LLM inference, not for symbolic candidate execution, coefficient optimization, or post-fit screening. The locally served \emph{Llama-3.1-8B}, \emph{Qwen3.6-27B}, and \emph{DeepSeek-V4-Pro} inference endpoints were deployed on 16 NVIDIA H20-3e data-center AI accelerators with CUDA 13.1 and 144 GiB HBM per card.

These GPU resources should therefore be interpreted as serving hardware for the local LLM backbones, which were served with vLLM 0.20.0 for inference. The CPU accounting in Appendix~\ref{app:resource_cpu_usage} remains the relevant local compute description for expression evaluation and fitting, while externally hosted LLMs such as \emph{Claude Opus 4.6} are represented only through the call and token accounting in Appendix~\ref{app:resource_llm_usage}.

\section{Additional Details of FunctionEvolve}
\label{app:functionevolve_details}

\subsection{Algorithmic Procedure}
\label{app:functionevolve_algorithm}

\begin{figure}[t]
  \centering
  \begin{minipage}{0.94\linewidth}
  \begin{algorithmfigurebox}
  {\large\textbf{FunctionEvolve}}\\[-0.15em]
  \rule{\linewidth}{0.8pt}
  \vspace{0.15em}

  \scriptsize
  \setlength{\tabcolsep}{2pt}
  \renewcommand{\arraystretch}{1.12}
  \begin{tabular}{@{}p{0.13\linewidth}p{0.82\linewidth}@{}}
  \textbf{Input} & LLM modules $\pi_G,\pi_S,\pi_M$, dataset $\mathcal{D}=\{(\mathbf{x}_i,y_i)\}_{i=1}^n$ with task-level context $T$\\
  \textbf{Output} & Ranked symbolic expressions with fitted constants and scores \\
  \end{tabular}

  \vspace{0.2em}
  \begin{tabular}{@{}p{0.035\linewidth}p{0.035\linewidth}p{0.035\linewidth}p{0.84\linewidth}@{}}
  \multicolumn{4}{@{}l}{\algcomment{Initialization by Generator}}\\
   & & & $\mathcal{K}\leftarrow \pi_G.\algproc{Knowledge}(T,\mathcal{D})$ \\
   & & & $\mathcal{S}_0\leftarrow \pi_G.\algproc{Seeds}(T,\mathcal{D},\mathcal{K})$ \\
   & & & $\mathcal{E}\leftarrow \emptyset,\ \mathcal{H}\leftarrow \emptyset$ \\
  \multicolumn{4}{@{}l}{\textbf{for} $f\in\mathcal{S}_0$ \textbf{do}}\\
  \algbar & & & $f,\Theta\leftarrow \algproc{Normalize}(f)$ \\
  \algbar & & & $s,\hat{\mathbf{c}}\leftarrow \algproc{Optimizer}(f,\Theta,\mathcal{D})$ \\
  \algbar & & & $\mathcal{E}\leftarrow \mathcal{E}\cup \{(f,\Theta,\hat{\mathbf{c}},s,\mathrm{root})\}$ \\
  \multicolumn{4}{@{}l}{\textbf{end for}}\\
  \multicolumn{4}{@{}l}{\algcomment{Evolution over the expression tree}}\\
  \multicolumn{4}{@{}l}{\textbf{while} stopping criterion is not met \textbf{do}}\\
  \algbar & & & $\mathcal{Z}\leftarrow \algproc{TreeSummary}(\mathcal{E},\mathcal{H})$ \\
  \algbar & & & $\mathcal{P}\leftarrow \pi_S.\algproc{SelectParents}(\mathcal{Z},T)$ \\
  \algbar & & & $\mathcal{C}\leftarrow \emptyset$ \\
  \algbar & & & \textbf{for} $p\in\mathcal{P}$ \textbf{do} \\
  \algbar & \algbar & & $\mathcal{A}\leftarrow \algproc{RuleAddition}(p)\cup\algproc{RuleDeletion}(p)$ \\
  \algbar & \algbar & & $\mathcal{L}\leftarrow \pi_M.\algproc{Mutate}(p,\algproc{AST}(p),\mathcal{A},\mathcal{K},\mathcal{Z})$ \\
  \algbar & \algbar & & $\mathcal{C}\leftarrow \mathcal{C}\cup\{(g,p):g\in\algproc{NormalizeDedup}(\mathcal{A}\cup\mathcal{L})\}$ \\
  \algbar & & & \textbf{end for} \\
  \algbar & & & \textbf{for} $(g,p)\in\mathcal{C}$ \textbf{do} \\
  \algbar & \algbar & & $g,\Theta_g\leftarrow \algproc{Normalize}(g)$ \\
  \algbar & \algbar & & \textbf{if} $\algproc{Valid}(g,\Theta_g,\mathcal{E})$ \textbf{then} \\
  \algbar & \algbar & \algbar & $s_g,\hat{\mathbf{c}}_g\leftarrow \algproc{Optimizer}(g,\Theta_g,\mathcal{D})$ \\
  \algbar & \algbar & \algbar & $\delta_g,\tilde g,\tilde\Theta_g\leftarrow \algproc{DegeneracyCheck}(g,\Theta_g,\hat{\mathbf{c}}_g)$ \\
  \algbar & \algbar & \algbar & $\mathcal{E}\leftarrow \mathcal{E}\cup\{(g,\Theta_g,\hat{\mathbf{c}}_g,s_g,\mathrm{parent}=p,\delta_g)\}$ \\
  \algbar & \algbar & \algbar & \textbf{if} $\delta_g=\textsc{simplified}$ \textbf{then} $\mathcal{C}\leftarrow\mathcal{C}\cup\{(\tilde g,g)\}$ \textbf{end if} \\
  \algbar & \algbar & & \textbf{end if} \\
  \algbar & & & \textbf{end for} \\
  \algbar & & & $\mathcal{H}\leftarrow \mathcal{H}\cup\mathcal{P};\quad \algproc{MarkMature}(\mathcal{E})$ \\
  \multicolumn{4}{@{}l}{\textbf{end while}}\\
  \multicolumn{4}{@{}l}{\textbf{return} $\algproc{TopExpressions}(\mathcal{E})$}\\
  \end{tabular}
  \end{algorithmfigurebox}
  \end{minipage}
\end{figure}

\paragraph{Operational subroutines}
\algproc{RuleAddition} is a deterministic parent-level expansion. For a selected parent $p$, it attaches the elementary content library in Table~\ref{tab:ast_mutation_templates} through $p+g$, $p\cdot g$, division templates, and unary wraps. There is no blanket preprocessing stage applied to all seeds or all offspring; each generated candidate corresponds to one explicit mutation step.

\textsc{DegeneracyCheck} is a deterministic state-management routine that decides whether an evaluated candidate remains active. It returns $\delta_g\in\{\textsc{ok},\textsc{simplified},\textsc{overfit},\textsc{timeout}\}$. It marks deep expressions with abnormal fitted coefficients ($|c_j|>10^3$) as \textsc{overfit}, applies constant-induced simplifications such as zeroing small multiplicative terms, snapping near-rational exponents, simplifying $\exp(0)$ and $\log(1)$, reducing simple trigonometric degeneracies, and merging nearly identical nonlinear terms. If the simplification removes all feature variables or creates infinity, the candidate is marked \textsc{overfit}; if it yields a different valid expression, the original evaluated candidate is marked \textsc{simplified} and the simplified expression is reinserted as its descendant for normal normalization, deduplication, and optimization. Timeouts are conservatively treated as \textsc{ok}.

These flags directly control the search tree. Nodes with $\delta_g\neq\textsc{ok}$ are kept only as evaluated audit records and are excluded from the Selector's tree summary, so they cannot become future parents. \textsc{MarkMature} marks an active evaluated node as mature once its training NMSE is below $\tau_{\mathrm{train}}$. Mature nodes may still be selected for a small annealing budget; after that budget is exhausted, they are removed from the Selector context. The run stops early after the configured number of mature nodes has been collected. \textsc{TopExpressions} returns mature nodes, supplemented if needed by non-mature evaluated nodes, ordered by training NMSE. Test and OOD NMSE are logged for post-hoc reporting only and are not exposed to the Selector or Mutator during search.

\paragraph{Full-run configuration}
Table~\ref{tab:run_full_config} lists the command-line configuration used by \texttt{run.sh full}; unspecified arguments use the defaults in \texttt{main.py}.

\begin{table}[h]
\centering
\caption{\texttt{run.sh full} configuration for the full FunctionEvolve setting.}
\label{tab:run_full_config}
\small
\begin{tabular}{@{}p{0.28\linewidth}p{0.36\linewidth}p{0.28\linewidth}@{}}
\toprule
Parameter & Meaning & Value \\
\midrule
\texttt{--optimizer} & Constant-fitting backend & \texttt{Structure} \\
\texttt{--max-steps} & Maximum evolutionary iterations per task & 30 \\
\texttt{--n-seeds} & Number of Generator seed expressions & 20 \\
\texttt{--candidate-num} & Number of parents selected per iteration & 5 \\
\texttt{--selector-context-size} & Maximum number of active nodes shown to the Selector & 200 \\
\texttt{--max-mature-nodes} & Early-stop target number of training-NMSE mature nodes & 50 \\
\texttt{--overfit-min-depth} & Minimum SymPy AST depth for large-coefficient overfit screening & 10 \\
\texttt{--equation-workers} & Active equation-level workers in full mode & 10 \\
\texttt{--n-eval-workers} & Local child-evaluation workers when no shared global worker pool is used & 16 (default) \\
\texttt{--timeout} & Per-task wall-clock timeout shared by normalization, evaluation, degeneracy check, and optimizer & 120 s (default) \\
\texttt{--refine-output} & Final parent-selection sweep for mature/supplement nodes & disabled in \texttt{full}; enabled only by \texttt{refine\_full} \\
\texttt{--data-dir} & LLM-SRBench data directory & optional; defaults to dataset loader path unless provided \\
\bottomrule
\end{tabular}
\end{table}

\subsection{Generator}
\label{app:generator_details}

\paragraph{Procedure}
The Generator is used during the one-shot initialization stage. It has two LLM-facing subroutines. First, \texttt{generate\_domain\_knowledge} reads the task context and available variables, infers either the scientific domain or, for dimension-only tasks such as AI-Feynman entries without natural-language background, a dimension-aware structural prior. It returns formula priors and heuristic structural cues. These LLM-proposed priors are combined with a fixed catalog of generic scientific laws or dimension/variable-name templates, so the downstream search receives both broad priors and task-specific or metadata-specific priors.

Second, \texttt{initialize\_seeds} uses the task context, variables, and extracted knowledge to generate the initial expression pool. The seed prompt asks for diverse structures spanning both compact monomial/product/ratio/composition forms and multi-term sums, with roughly half generic elementary-function structures and half knowledge-informed structures. Each response must be a JSON array of SymPy-parseable expressions together with the parameter placeholders used in the expression. The implementation validates the JSON schema, parses each expression with SymPy, checks that free symbols come only from dataset variables or listed \texttt{c0}, \texttt{c1}, \ldots constants, and requires the listed constants to match those actually used. If LLM generation yields too few valid seeds, a fixed fallback seed list supplements the pool.

\begin{table}[h]
\centering
\caption{Generator configuration used by the \emph{Claude Opus 4.6} full run. Component-specific overrides are read from \texttt{opus-4-6.yaml}.}
\label{tab:generator_config}
\small
\begin{tabular}{@{}p{0.38\linewidth}p{0.44\linewidth}p{0.10\linewidth}@{}}
\toprule
Parameter & Meaning & Value \\
\midrule
\texttt{n\_seeds} & Number of initial seed expressions & 20 \\
\texttt{generator.temperature} & Default Generator decoding temperature & 1.0 \\
\texttt{domain\_knowledge.temperature} & Domain-knowledge extraction temperature & 0.7 \\
\texttt{seed\_generation.temperature} & Seed-expression generation temperature & 0.9 \\
\texttt{describe\_batch.temperature} & Batch-description temperature & 0.5 \\
\texttt{max\_tokens} & Maximum response length for Generator subroutines & 128000 \\
\bottomrule
\end{tabular}
\end{table}

\paragraph{Degenerated Generator}
For ablations, FunctionEvolve can replace the LLM Generator with a deterministic \texttt{MockGenerator}. In this mode, no LLM calls are made: domain knowledge is empty and initial expressions are drawn from a fixed fallback list. This isolates the contribution of LLM-based initialization and domain-prior extraction from the rest of the search pipeline.

\paragraph{Prompting interface}
The Generator uses separate prompt interfaces for domain knowledge extraction and seed expression initialization. The prompt boxes below show the interfaces used by these subroutines in abbreviated form. The domain knowledge prompt returns a JSON object, while the seed prompt returns a JSON array. In both cases, the implementation requests JSON-only output, optionally attaches a guided JSON schema when the backend supports it, extracts a JSON block if one is present, and validates the parsed result before it is inserted into the search state. The same seed-generation interface is used across LLM-SRBench and AI-Feynman; for AI-Feynman tasks without natural-language background, only the domain-knowledge stage switches to a dimension/variable/range-aware variant, and the seed prompt labels the knowledge-informed half as dimension/variable-informed formulas.

\nolinenumbers
\begin{tcolorbox}[title=Generator domain-knowledge system prompt, breakable, colback=gray!5, colframe=gray!60]
\begin{Verbatim}[fontsize=\small, breaklines=true, breakanywhere=true]
You are a cross-disciplinary mathematical modeling expert, proficient in classical formulas and mathematical models from physics, chemistry, biology, engineering, economics, and other fields.

Given a background description of a symbolic regression task (including variable names and their physical meanings), please:
1. Analyze the most likely scientific domain/sub-domain for this task
2. List classical formula structures related to these variables in that domain

## Output Format
Return only a JSON object:
{"domain": "<domain name>", "analysis": "<1-2 sentence domain analysis>", "formulas": [...], "heuristics": [...]}

- domain: The scientific domain this task belongs to.
- analysis: Briefly explain why it belongs to this domain and the possible physical relationships between variables.
- formulas: List 5~15 common formula structural patterns in this domain. Use actual variable names when possible, arrange from simple to complex, include domain-specific function forms, and write each entry as "Name: formula form (applicable scenario)".
- heuristics: List 5~10 domain-specific formula-structure heuristic features describing what physical behavior corresponds to what mathematical structure.

Constant notation convention: constant parameters must appear in multiplicative form and must not be placed in the denominator. Use `c * var` rather than `var / c`.
\end{Verbatim}
\end{tcolorbox}

\begin{tcolorbox}[title=Generator domain-knowledge user prompt, breakable, colback=gray!5, colframe=gray!60]
\begin{Verbatim}[fontsize=\small, breaklines=true, breakanywhere=true]
{context}

Available variables: {variables}

Please analyze the scientific domain of this task and list common formula structures related to these variables in that domain.
\end{Verbatim}
\end{tcolorbox}

\begin{tcolorbox}[title=Generator seed system prompt, breakable, colback=gray!5, colframe=gray!60]
\begin{Verbatim}[fontsize=\small, breaklines=true, breakanywhere=true]
You are a symbolic-regression expert. Generate diverse candidate closed-form formulas for the dataset described below, to seed an evolutionary search.

## Structural Prior
The ground-truth formula may be EITHER:
- a single compact monomial / product / ratio / composition, such as `c0*x1*x2/x3` or `c0*exp(-c1*x**2)`; OR
- a sum of several terms, such as `c0*x + c1*x**2 + c2`.
Your seed set MUST cover both archetypes.

## Building Blocks
power `x**c`, trigonometric `sin`/`cos`, exponential `exp`, logarithmic `log(1+x)`, `sqrt`, and, where physically meaningful, `Abs(x)` and `Max(x, 0)`.

## Using {knowledge_basis}
{domain_knowledge}

## Constant Notation Convention
Constant parameters should always appear in multiplicative form, never alone in a denominator: write `c0*x` and `c0/(x + c1)`, not `x/c0`.

## Task Background
{context}

## Available Variables
{variables}

Please generate {n_seeds} candidate formulas with different structures at once, and return only a JSON array:
[{"expression":"...","params":["c0","c1"]}, ...]
\end{Verbatim}
\end{tcolorbox}

\begin{tcolorbox}[title=Generator seed user prompt, breakable, colback=gray!5, colframe=gray!60]
\begin{Verbatim}[fontsize=\small, breaklines=true, breakanywhere=true]
Based on the above task metadata, generate {n_seeds} candidate formulas with diverse structures.

## Requirements
Formulas must cover both of the following categories, arranged from simple to complex:

A. Basic function structures (approximately half): Include BOTH compact single-term shapes (monomial/product/ratio/composition) AND multi-term sums. Ensure at least one variant of each basic type:
- Monomial / product: c0*x1*x2, c0*x1*x2/x3
- Power-law / composition: c0*x**c1, c0*exp(-c1*x**2), c0/sqrt(1 - x**2/c1**2)
- Linear: c0*x + c1
- Quadratic/polynomial: c0*x**2 + c1*x + c2
- Non-integer power law: c0*x**c1
- Exponential: c0*exp(c1*x)
- Logarithmic: c0*log(1 + c1*x)
- Rational: c0*x/(x + c1)
- Trigonometric: c0*sin(c1*x)
- Hyperbolic: c0*tanh(c1*x + c2)
- Square root: c0*sqrt(x**2 + c1)
- Simple combinations of the above basic types

B. Knowledge-informed formulas (approximately half): Based on the task background and the domain-knowledge or dimension/variable-name structures listed above, generate formula forms most likely to match the data.

Note: All constants in formulas must be represented as c0, c1, c2..., and variables must use names from the available variables list.
\end{Verbatim}
\end{tcolorbox}

\linenumbers

\subsection{Evolutionary Tree Search}
\label{app:evolutionary_tree}

FunctionEvolve maintains the search state as an evolutionary tree rather than as an unstructured pool of candidate expressions. Each node in the tree corresponds to a symbolic expression together with its associated search state, while edges record parent-child relations induced by mutation. This representation preserves structural lineage throughout the search and allows the system to summarize the current search frontier for downstream selection.

At a high level, the tree stores two types of information: the current set of candidate expressions and the parent-child relations among them. Each node further maintains its own evaluation statistics, structural information, and parameter-related information. Table~\ref{tab:evolution_node_fields} summarizes the main information maintained for each node, together with whether it is visible to the Selector.

\begin{table*}[t]
\centering
\caption{Main information maintained for each node in the evolutionary tree. The last column indicates whether the field itself, or a derived summary of it, is visible to the Selector.}
\label{tab:evolution_node_fields}
\small
\setlength{\tabcolsep}{3pt}
\begin{tabular}{@{}p{0.20\textwidth}p{0.15\textwidth}p{0.51\textwidth}p{0.08\textwidth}@{}}
\toprule
Field & Type & Description & Selector visibility \\
\midrule
\texttt{id} & string & Identifier for this node. & Yes \\
\texttt{formula} & string & Canonical symbolic expression string. & Yes \\
\texttt{parent\_id} & string / None & Identifier of the parent node in the evolutionary tree. & Yes \\
\texttt{train\_nmse} & float / None & Training NMSE after constant fitting. & Yes \\
\texttt{test\_nmse} & float / None & Test NMSE maintained for evaluation. & No \\
\texttt{ood\_test\_nmse} & float / None & Out-of-distribution NMSE maintained for evaluation. & No \\
\texttt{is\_evaluated} & bool & Whether the node has been evaluated after constant fitting. & No \\
\texttt{tree\_depth} & integer & Depth of the parsed expression tree; exposed as \texttt{depth}. & Yes \\
\texttt{operator\_counts} & dictionary & Counts of operator types in the expression tree; exposed to the Selector only through the aggregated field \texttt{n\_operators}. & Yes \\
\texttt{ast\_features} & structured features & Additional AST-derived structural features used internally. & No \\
\texttt{param\_names} & list of strings & Names of tunable scalar parameters; exposed only through the aggregated field \texttt{n\_params}. & Yes \\
\texttt{fitted\_params} & list of floats & Optimized parameter values; exposed to the Selector as a formatted string when available. & Yes \\
\texttt{sympy\_expr} & symbolic object & Parsed symbolic expression used for internal manipulation and analysis. & No \\
\bottomrule
\end{tabular}
\end{table*}

\subsection{Selector}
\label{app:selector_summary}

\paragraph{Procedure}
At each search iteration, the Selector receives a compact summary of the current search tree, together with the task context and the record of previously chosen parents. Based on this information, it identifies a small set of parent expressions for further expansion and provides a brief rationale for each choice. These selected parents are then forwarded to the subsequent mutation stage. In this way, the Selector determines how the available search budget is allocated across competing search directions.

\begin{center}
\centering
\captionof{table}{Selector configuration used in our experiments.}
\label{tab:selector_config}
\small
\begin{tabular}{@{}p{0.38\linewidth}p{0.44\linewidth}p{0.10\linewidth}@{}}
\toprule
Parameter & Meaning & Value \\
\midrule
\texttt{candidate\_num} & Number of parents selected per iteration & 5 \\
\texttt{selector\_context\_size} & Maximum number of summarized active nodes visible to the Selector in \texttt{run.sh full} & 200 \\
\texttt{temperature} & Sampling temperature for \emph{Claude Opus 4.6} Selector decoding & 0.3 \\
\texttt{max\_tokens} & Maximum response length for Selector decoding & 128000 \\
\bottomrule
\end{tabular}
\end{center}

\paragraph{Visible node information}
Although each tree node stores a full internal state, the Selector does not access this state directly. Instead, it operates on the subset of per-node information marked as visible in Table~\ref{tab:evolution_node_fields}, together with the aggregated statistic \texttt{n\_children}. In practice, this summary provides information about predictive quality, structural complexity, local parameterization, and search history, without exposing the full internal node state. Importantly, the summary contains only training error. Test error and out-of-distribution error are maintained internally for evaluation, but are never revealed to the Selector. This design prevents evaluation-time information from leaking into the parent-selection process.

\paragraph{Degenerated Selector}
For ablations, FunctionEvolve supports two degenerated Selector variants. The first variant replaces the LLM Selector with a deterministic \texttt{MockSelector}. This fallback ranks evaluated nodes by numerical fitness and samples parents using a rank-based Boltzmann distribution, so no LLM call is made during parent selection. The second variant keeps the LLM-based parent-selection interface but removes AST-derived structural fields from the prompt, including parameter count, tree depth, and operator count. This preserves LLM decision making while hiding explicit AST structure, isolating the contribution of structural visibility in the Selector.

\paragraph{Prompting interface}
The exact prompt templates used by the Selector are shown below. The system prompt defines the selection criteria and the required JSON-only output format, while the user prompt template supplies the current tree summary and the historical selection records. The implementation also attaches the same guided JSON schema when the backend supports it. The Selector uses the same main template across datasets; AI-Feynman tasks append the dimension/variable/range task background and a short instruction to prioritize formula quality, structural diversity, variable-name cues, and dimensional metadata. For the ablation without AST-derived metadata, we use a variant with the corresponding fields removed.

\nolinenumbers
\begin{tcolorbox}[title=Selector system prompt, breakable, colback=gray!5, colframe=gray!60]
\begin{Verbatim}[fontsize=\small, breaklines=true, breakanywhere=true]
You are a strategic planning expert for symbolic regression search.

You will see a summary table of all generated formulas in the current evolution tree, each record contains:
- id          : Numeric unique identifier of the formula
- formula     : Formula in SymPy format
- train_nmse  : Training set NMSE = MSE/Var(Y) (None means not yet evaluated; lower is better, <1 is better than mean prediction)
- n_children  : Number of offspring this node has produced
- n_params    : Number of tunable parameters
- depth       : AST tree depth (nesting level; higher means deeper function nesting)
- n_operators : Total number of operator nodes in AST tree (higher means longer expression)
- fitted_params : Fitted parameter values (scientific notation)

Your task is to select **{candidate_num} parent nodes** for the next evolution iteration.
The system will automatically perform AST-based structural mutations (delete subtrees, add new terms, etc.) on each parent to generate multiple candidate offspring.

Please consider the following when selecting:
1. **Structural simplicity first**: Do not simply select the nodes with the lowest NMSE. Prefer structurally simple nodes (fewer n_params, shallower depth, fewer n_operators) that already have relatively low NMSE (e.g., ~1e-3 order) - simple good formulas are more likely to produce meaningful improvements through mutation, while low NMSE in complex formulas often comes from overfitting
2. **Algebraic structure diversity**: The selected {candidate_num} parents should cover different algebraic structures as much as possible (e.g., polynomial, power law, exponential, trigonometric, rational, etc.), avoiding highly similar formulas. Diverse starting points enable more comprehensive exploration of the search space
3. **NMSE performance**: When structures are similar, prefer nodes with lower NMSE that still have room for improvement
4. **Number of offspring (n_children)**: If a node already has many offspring (e.g., n_children >= 5), it means that direction has been sufficiently explored; prefer nodes with fewer offspring
5. **Historical selection records**: Refer to the historical parent selection records. If certain formula structures have been repeatedly selected as parents in recent rounds but NMSE has not significantly decreased, they should be temporarily shelved in favor of other directions; if certain structures have not been explored recently, encourage selecting them as parents to expand the search space

Return only a JSON array with exactly {candidate_num} objects. Use the numeric `id` shown in the node summary as `parent_id`; do not copy the formula string into `parent_id`.

Required JSON schema:
{
  "type": "array",
  "minItems": {candidate_num},
  "maxItems": {candidate_num},
  "items": {
    "type": "object",
    "required": ["parent_id", "rationale"],
    "additionalProperties": false,
    "properties": {
      "parent_id": {"type": "integer"},
      "rationale": {"type": "string"}
    }
  }
}

Example item format (shape only; your actual output must contain exactly {candidate_num} objects):
[
  {"parent_id": 12, "rationale": "<selection rationale 1>"},
  {"parent_id": 7, "rationale": "<selection rationale 2>"}
]

- rationale: short final reason for selecting this parent.
\end{Verbatim}
\end{tcolorbox}

\begin{tcolorbox}[title=Selector user prompt, breakable, colback=gray!5, colframe=gray!60]
\begin{Verbatim}[fontsize=\small, breaklines=true, breakanywhere=true]
Current evolution tree node summary ({n_nodes} nodes total, sorted by train_nmse ascending):

{tree_summary}

## Historical Parent Selection Records
{selection_history}

Please comprehensively analyze the above nodes and historical selection records, select {candidate_num} structurally diverse and concise parents, and output a JSON array.
\end{Verbatim}
\end{tcolorbox}
\linenumbers

\subsection{Mutator}
\label{app:mutator_details}

\paragraph{AST representation}
Each symbolic expression is parsed into a SymPy abstract syntax tree (AST), where internal nodes correspond to operators such as \texttt{Add}, \texttt{Mul}, \texttt{Pow}, \texttt{exp}, \texttt{log}, and \texttt{sin}, and leaf nodes correspond to variables or scalar parameters. For each expression $f$, we extract depth-aware structural features of the form
\[
\mathcal{F}(f)=\{\texttt{Op}(\text{Depth}:k)\mid \texttt{Op} \text{ is the operator type of a node and } k \text{ is its depth}\}.
\]
For example, the expression $c_0 x^{c_1}+c_2\exp(c_3 x)$ contains features such as \texttt{Add(Depth:0)}, \texttt{Mul(Depth:1)}, \texttt{Pow(Depth:2)}, and \texttt{exp(Depth:2)}.

\paragraph{Rule-based mutations}
The AST-rule Mutator applies deterministic edits directly on the expression tree. We use two classes of rule-based mutations. The first class performs deletion and simplification operations, including removing one term from an \texttt{Add} node, removing one non-numeric factor from a \texttt{Mul} node, unwrapping elementary functions such as $\exp(g)\mapsto g$, $\log(g)\mapsto g$, and $\sin(g)\mapsto g$, and reducing powers such as $x^n\mapsto x$. These operations are recursively applied to valid subtrees and remove candidates that do not contain feature variables.

The second class performs broad template-based addition. Given a parent expression $f$, the Mutator first defines an elementary content library $g$ using the families in Table~\ref{tab:ast_mutation_templates}. It then enumerates $f+g$ additive terms, $f\cdot g$ multiplicative factors, division templates, and unary wraps $\phi(c f)$ or $f^c$. In implementation, division uses the numerically stable subset $f/(1+g_s)$ with $g_s$ instantiated from linear, power, exponential, and logarithmic content, so the edit remains local to the parent and avoids bare learned constants in denominators. We intentionally keep this rule library generic and domain-agnostic to reduce dataset-specific overfitting, leaving richer structural refinements to the LLM-guided channel. For multivariate problems, the pair rational and power-law coupling families are instantiated over up to six ordered variable permutations, while single-variable families are instantiated over each feature variable.

\paragraph{Normalization and deduplication}
To reduce redundant candidates, we define a parameter-name-invariant structural fingerprint $\kappa(f)$. The fingerprint is constructed by recursively replacing all scalar parameter symbols in the AST with a shared placeholder, sorting the children of commutative operators such as \texttt{Add} and \texttt{Mul}, and serializing the resulting normalized tree. Expressions with the same fingerprint are treated as structurally equivalent up to parameter renaming.

After mutation, candidates are further normalized before evaluation. The normalization procedure applies algebraic simplifications, merges redundant scalar parameters, completes elementary-function parameterizations when necessary, and renumbers parameters in order of appearance as $c_0,c_1,c_2,\ldots$. Examples include rewriting $\exp(A+c)$ as $c'\exp(A)$, rewriting $\exp(c\log g)$ as $g^c$, simplifying $\log(\exp(g))$ to $g$, and merging products or sums of scalar parameters into a single parameter.

\paragraph{LLM-guided mutations}
In addition to rule-based mutations, the LLM Mutator proposes non-template structural refinements for each parent expression. Its input includes the task context, the parent formula, an annotated AST representation, fitted parameter values, summaries of already generated AST-rule mutations, extracted domain knowledge, and top-ranked historical expressions by training NMSE. This information allows the LLM to avoid duplicating existing programmatic mutations while proposing non-template local refinements informed by task context and domain knowledge.

The LLM Mutator uses two edit categories. \textsc{ADD} preserves the full parent and attaches out-of-library or domain-semantic material as a term, factor, or outer wrap. \textsc{SUBST} uses the annotated AST to replace a structurally mismatched subtree, so the full parent need not be preserved. In our implementation, the LLM proposes 20 candidates per parent with no fixed category quota; the prompt asks it to choose the mix according to the parent structure and fitted behavior.

\begin{table}[h]
\centering
\caption{Categories of LLM-guided mutations.}
\label{tab:llm_mutation_categories}
\small
\begin{tabular}{@{}p{0.18\linewidth}p{0.72\linewidth}@{}}
\toprule
Category & Description \\
\midrule
\textsc{ADD} & Preserve $f$ and attach non-template or domain-semantic content as $f+g$, $f\cdot g$, or $\phi(f)$. \\
\textsc{SUBST} & Replace a structurally wrong AST subtree with a different expression; the replaced subtree is not preserved. \\
\bottomrule
\end{tabular}
\end{table}

\paragraph{Degeneracy and overfitting screening}
After constant fitting, each evaluated candidate is screened before it can re-enter the active search frontier. The screening is intentionally coefficient-aware: it uses the fitted constants rather than only the symbolic skeleton. Large-coefficient overfitting is checked only for sufficiently deep expressions, because shallow expressions often require large physical scale factors. Specifically, if the SymPy AST depth of the candidate is at least \texttt{overfit\_min\_depth}, any fitted coefficient with magnitude above $10^3$ marks the node as \textsc{overfit}. Independently of depth, the Mutator applies the simplification rules described after Algorithm~1, including small-coefficient term collapse, rational snapping of exponent parameters, elementary identities such as $\exp(0)$ and $\log(1)$, sinusoidal degeneracies, and merging of like terms.

The result is not only diagnostic. A node flagged as \textsc{overfit} is excluded from all future parent-selection prompts. A node flagged as \textsc{simplified} is also excluded in its original form, while the simplified expression is treated as a new candidate and evaluated through the same optimizer and screening path. Thus the flag $\delta_g$ restricts search by preventing degenerate high-scoring expressions from consuming Selector and mutation budget, while still preserving useful simplified descendants when the fitted constants reveal a simpler active formula.

\paragraph{Configuration}
Table~\ref{tab:mutator_config} summarizes the Mutator configuration used in our symbolic-regression experiments. The deterministic channel is always enabled in the full FunctionEvolve setting, and the LLM channel is configured by the per-model YAML files in \texttt{FunctionEvolve}.

\begin{table}[h]
\centering
\caption{Mutator configuration used in our experiments.}
\label{tab:mutator_config}
\small
\begin{tabular}{@{}p{0.34\linewidth}p{0.42\linewidth}p{0.16\linewidth}@{}}
\toprule
Parameter & Meaning & Value \\
\midrule
\texttt{mutation\_channels} & Candidate-generation channels in the full setting & AST deletion, AST addition, LLM mutation \\
\texttt{llm\_candidates\_per\_parent} & Number of LLM proposals requested for each selected parent & 20 \\
\texttt{mutator\_seen\_topk} & Number of historical best expressions shown to the LLM Mutator & 100 \\
\texttt{auto\_mutation\_summary} & Maximum number of rule-generated mutation descriptions shown to the LLM & 20 \\
\texttt{max\_params} & Maximum number of scalar parameters allowed in an evaluated candidate & 10 \\
\texttt{overfit\_min\_depth} & Minimum SymPy AST depth for degeneracy and overfitting screening & 10 \\
\texttt{temperature} & Sampling temperature for \emph{Claude Opus 4.6} Mutator decoding & 0.7 \\
\texttt{reasoning\_effort} & Reasoning setting for \emph{GPT-5.2} Mutator decoding & medium \\
\texttt{max\_tokens} & Maximum response length for Mutator decoding & 128000 \\
\bottomrule
\end{tabular}
\end{table}

\paragraph{Prompting interface}
The prompt templates used by the LLM Mutator in the full FunctionEvolve setting are shown below. The system prompt defines the ADD/SUBST edit categories, domain-knowledge constraints, constant-notation convention, output schema, and allowed variables. The user prompt supplies the selected parent, its annotated AST, fitted parameter values, summaries of deterministic rule-generated candidates, and the current historical best formulas. This Mutator template is shared across LLM-SRBench and AI-Feynman; AI-Feynman changes the task background and the knowledge-section heading to dimension/variable-informed structures, but does not use a separate mutation prompt. The no-AST ablation uses the same JSON output contract but switches to a no-AST system prompt and removes the annotated AST block and AST-local substitution guidance from the user prompt.

\nolinenumbers
\begin{tcolorbox}[title=LLM Mutator system prompt, breakable, colback=gray!5, colframe=gray!60]
\begin{Verbatim}[fontsize=\small, breaklines=true, breakanywhere=true]
You are a symbolic-regression expert. Propose **{target_num}** diverse structural **edits** to the parent formula so that each resulting candidate may fit the data better. Every edit takes the parent expression and transforms it into one new candidate expression.

## Structural Prior
Do **NOT** assume the ground-truth formula is a sum of terms. It may be a compact monomial / product / ratio / composition, or a sum of several mechanistic terms. Let the parent's current fit guide you: if the parent already captures the trend, prefer small local refinements; if the parent looks structurally wrong, prefer a rewrite. Do **not** blindly keep growing the formula additively.

## Two Kinds of Edit
The deterministic engine already covers simple template grafts (linear / power / exp / log / sin terms and factors, rational fractions, safe saturating divides, and single wraps) and all prunes / deletions / simplifications. See "Auto-Generated Candidates (do not repeat)" in the task. Your job is the part it cannot template:

- **ADDITION** (keep all of `f`; attach new material): `f + g`, `f * g`, or wrap `f` inside an outer function. Propose **out-of-library / domain-semantic** content, such as Arrhenius terms, Lorentz factors, cross-variable couplings, or nested compositions. The whole parent is preserved.
- **SUBSTITUTION** (discard a subtree, replace it): use the annotated AST to locate a structurally wrong subtree and replace it with a different structure (`t -> s`, where `t` no longer appears). Use this when the parent's shape is mismatched, not when you merely want to add a correction.

Do **not** propose pure simplifications / deletions -- the program handles those. Weight your {target_num} proposals toward whichever kind the parent's fit and structure call for; there is no fixed quota.

Elementary building blocks: power `x**c`, trigonometric `sin`/`cos`, exponential `exp`, logarithmic `log(1+x)`, and `sqrt`.

## Non-Elementary Function Usage
In addition to the elementary functions above, the following non-elementary functions are very common in scientific modeling and may be used as appropriate in either edit category:
- **Abs(x)** (absolute value): suitable for modulus operations in physical systems, amplitude extraction of sign-alternating variables in oscillatory systems. E.g. Abs(sin(x)) extracts oscillation amplitude, Abs(x)**c constructs V-shaped power law, log(1 + Abs(x)) symmetrizes logarithmic growth. Prefer Abs when a variable can be positive or negative but only the magnitude matters physically.
- **Max(x, 0)** (positive part / ReLU): suitable for describing piecewise behavior such as threshold activation, one-sided response. E.g. Max(x - c, 0) means response only above threshold c, Max(f(x), 0) truncates the negative part. Suitable for systems with critical points / activation thresholds.
- Syntax: use `Abs(x)` and `Max(x, 0)` in SymPy.

## Using Domain Knowledge
Design structures informed by the knowledge below. You may introduce a structure through either kind of edit: as an **ADDITION** (a new factor / term / wrap that preserves `f`) or as a **SUBSTITUTION** (rewrite a mismatched subtree). When an edit is motivated by this knowledge, state the basis in the `mutation` field.
{domain_knowledge}
## Constant Notation Convention
Constant parameters should always appear in **multiplicative form**, never alone in a denominator:
- Correct: `c0 * x`, `c0 * sin(c1 * x)`, `c0 / (x + c1)`
- Incorrect: `x / c0`, `sin(x / c0)` -- a constant in the denominator is numerically unstable. To express a "scaled variable", write `c * var`, not `var / c`.

## Parameter Naming
Any **newly introduced** constant must use a **fresh** `cK` that does not appear in the parent formula or its fitted parameters; keep the names of parent constants you retain.

## Task Background
{context}

## Output Format (strict compliance required)
Return only a JSON array of exactly {target_num} items.

Example item format:
[{{"expression": "<SymPy-parseable formula>", "params": ["c0", "c1"], "mutation": "ADD: <what changed and why>"}}]

- expression: use c0, c1, c2... for constants to be optimized; variables must come from the available variable list.
- params: list all parameter placeholders; if no constants, use [].
- mutation: **must** start with `ADD:` or `SUBST:`, followed by a brief rationale. When the edit is knowledge-driven, name the formula / law / structure it references.
- All functions must use SymPy/Python syntax: exp(x), log(x), sqrt(x), sin(x), cos(x), Abs(x), Max(x, 0).

## Available Variables
{variables}
\end{Verbatim}
\end{tcolorbox}

\begin{tcolorbox}[title=LLM Mutator user prompt, breakable, colback=gray!5, colframe=gray!60]
\begin{Verbatim}[fontsize=\small, breaklines=true, breakanywhere=true]
## Parent Formula
{parent_formula}

## AST Structure
{labeled_ast}

## Fitted Parameters
{parent_fitted_params}

## Auto-Generated Candidates (do not repeat)
{auto_mutations_summary}

## Historical Best Formulas (top {topk})
{top_exprs}

Following the system instructions, propose {target_num} edits of two kinds: **ADDITION** (out-of-library / domain-semantic terms, factors, or wraps that preserve the parent) and **SUBSTITUTION** (rewrite a structurally wrong subtree located via the AST). Do not repeat the auto-generated template candidates above, and do not propose pure simplifications. There is no fixed quota -- weight the mix toward what the parent's AST and current fit suggest.
\end{Verbatim}
\end{tcolorbox}
\linenumbers

\paragraph{Degenerated Mutator}
For ablations, FunctionEvolve supports three degenerated Mutator variants. The first variant disables the LLM mutation channel and keeps only programmatic AST-rule mutations, namely rule deletion and template-based rule addition. The second variant disables the programmatic rule channel and keeps only LLM-generated ADD/SUBST edits, while still providing the LLM with the standard parent-level structural context. The third variant also uses only LLM-generated ADD/SUBST edits, but further removes the annotated AST representation and replaces AST-local mutation guidance with parent-formula subexpression guidance in the LLM prompt. The second and third variants do not carry a hidden blanket preprocessing stage. This set of variants separates the effects of deterministic AST-rule expansion, LLM-guided local structural refinement, and explicit AST visibility in the mutation interface.

\subsection{Optimizer}
\label{app:optimizer_details}

For each candidate expression, the structure-aware coefficient optimizer fits scalar coefficients under a fixed symbolic skeleton and returns the numerical score used by the evolutionary search. Since inaccurate coefficient fitting can make a correct skeleton appear unpromising, this appendix details how FunctionEvolve uses expression structure during coefficient fitting, following the same sequence as the main text.

\paragraph{Expression analysis before optimization}
Before numerical search, the optimizer parses the candidate into a symbolic expression and extracts parameter-specific constraints from the expression structure. Parameters that appear as exponents over feature-dependent bases are treated carefully, especially when the base can be negative, because arbitrary real exponents can leave the real domain. In this case the implementation evaluates such powers through a real-valued safe-power wrapper and snaps the exponent to the nearest multiple of $1/3$ when needed. The optimizer also identifies parameters that should be positive or searched over a smaller range, such as coefficients inside logarithms, exponentials, trigonometric arguments, or power exponents. The default full-parameter box is $[-100,100]$; rational-sensitive exponents use $[-10,10]$, positive parameters use $[10^{-6},100]$, small parameters use $[-5,5]$, and parameters that are both positive and small use $[10^{-6},5]$. In addition, data-dependent bounds are computed for parameters controlling exponential growth and feature offsets. For terms of the form $\exp(c_i g(\mathbf{x}))$, the coefficient bound is chosen so that the absolute exponential argument is at most 10 over the training data. For offset parameters in $\exp(g(x-c_i))$ or $(x-c_i)^p$, the search interval is the observed feature range enlarged by $\max(0.2\,\mathrm{range}(x),10)$.

\paragraph{Variable projection}
The main optimization path first attempts to separate linear and nonlinear parameters. Many candidates can be written as
\[
f(\mathbf{x};\mathbf{c})
=
f(\mathbf{x};\mathbf{c}_L,\mathbf{c}_N)
=
\psi(\mathbf{x};\mathbf{c}_N)+
\sum_{j=1}^{m} c_{L,j}\phi_j(\mathbf{x};\mathbf{c}_N),
\]
where $\mathbf{c}_L$ are linear coefficients and $\mathbf{c}_N$ are nonlinear parameters. For any fixed $\mathbf{c}_N$, the optimal $\mathbf{c}_L$ is obtained by ordinary least squares, so the explicit search only needs to explore the nonlinear subspace:
\[
\min_{\mathbf{c}_N}
\min_{\mathbf{c}_L}
\left\|
\mathbf{y}
-
\psi(\mathbf{X};\mathbf{c}_N)
-
\sum_{j=1}^{m} c_{L,j}\phi_j(\mathbf{X};\mathbf{c}_N)
\right\|_2^2 .
\]
In practice, the optimizer differentiates the expression with respect to candidate parameters, selects a maximal subset that is jointly affine, and constructs the OLS basis functions from the corresponding partial derivatives after setting the linear parameters to zero. The remaining fixed part $\psi$ is subtracted from the target before solving the linear coefficients analytically at each nonlinear parameter assignment.

\paragraph{Structured pre-search for difficult nonlinear parameters}
Some nonlinear parameters are particularly difficult for continuous optimization. For compound power structures such as $(x+c_1)^{c_2}$, where the power exponent is itself a single parameter and the base contains another parameter, the optimizer first enumerates
\[
\{1,2,1/2,3,1/3,-1,3/2,4,5,-1/2,-2,-3\}
\]
for the exponent and runs variable projection on the remaining parameters under each fixed value. This pre-search is executed before the main variable-projection stage and can consume up to 70\% of the variable-projection time budget. Inside variable projection, any remaining power-exponent parameters are handled by a second candidate grid,
\[
\{1/3,1/2,1,3/2,2,3,4,5,-1/3,-1/2,-1,-2,-3\},
\]
using a sequential greedy strategy when multiple exponents are present. For each fixed exponent value, the remaining nonlinear parameters are searched by trust-region least squares with five random restarts, and linear coefficients are still solved by OLS. For center or offset parameters appearing in Gaussian-like, exponential-shift, or shifted-power forms, the optimizer detects $(x-c_i)$-style subexpressions and scans 100 equally spaced candidate centers between the observed minimum and maximum of the corresponding feature. These structured pre-search steps provide strong initial basins before the general nonlinear search begins.

\paragraph{Global search over nonlinear parameters}
After variable projection, the remaining objective over $\mathbf{c}_N$ is lower-dimensional but can still be highly non-convex. FunctionEvolve therefore searches the nonlinear subspace with trust-region least squares, CMA-ES, and differential evolution, each using an independent initialization drawn from the same constraint-aware sampler. The variable-projection path uses the first 45\% of the optimizer timeout, with the remaining time in that window divided approximately equally across these three continuous search stages after structured pre-search. This design focuses global exploration on the parameters that determine the nonlinear shape of the expression, while avoiding unnecessary search over linear coefficients that can be solved exactly by OLS.

The fallback condition is explicit in the implementation. Let $\sigma_y^2=\mathrm{Var}(\mathbf{y}_{\mathrm{train}})$, with a unit fallback for nearly constant targets, and define the early-success threshold $\tau=10^{-10}\sigma_y^2$. Variable projection is treated as unavailable when no valid decomposition is produced, for example because the expression has no jointly affine coefficient, no basis functions can be constructed, or OLS/basis/fixed-part evaluation becomes non-finite. It is treated as insufficient when the best MSE found so far remains above $\tau$. In either case, the optimizer falls back to full-parameter numerical search until 90\% of the total timeout, allocating the remaining fallback budget equally to full-parameter trust-region least squares, CMA-ES, and differential evolution. Thus ``insufficient'' is not a hidden symbolic criterion; it is the same MSE threshold used for early termination of the optimizer.

\paragraph{Local refinement}
The best solutions obtained from the preceding stages are then refined with gradient-based local optimization, implemented as L-BFGS-B over the same parameter bounds. This polishing stage uses the final 10\% of the total timeout. If parent parameters are available from the evolutionary tree, the optimizer also performs an additional warm-started L-BFGS-B pass after copying the overlapping parent parameter prefix into the current skeleton.

\paragraph{Snap and refit}
Finally, the optimizer performs optional snapping for exponent-like or rational-sensitive parameters. Parameters that must remain stable over negative bases are snapped to nearby rational values whose reduced denominators are odd. Power exponents are then tested against the same nearby integer/rational grid used above, restricted to candidates within distance 3 of the current value. Crucially, snapping is followed by an L-BFGS-B refit with the snapped exponent fixed; the snapped form is accepted only if the refit preserves or improves the numerical objective. This makes the final expression more stable and interpretable without sacrificing the fitness estimate used by the search.

\subsection{Optimizer Benchmark Variants}
\label{app:optimizer_benchmark_variants}
The optimizer benchmark in Section~\ref{sec:optimizer_pipeline_analysis} is built from ground-truth LLM-SRBench skeletons so that failure can be attributed to coefficient fitting rather than structural search. Numeric constants are replaced by symbolic parameters, parameter couplings are decoupled when possible, and the resulting exact skeleton forms the original-expression group. We then create controlled variants that preserve the same target function under a known parameter assignment. The composite zero-term variants add zero-amplitude nonlinear terms, either $c_a\sin(\log(1+c_bx)+c_c)$ or $c_a\exp(c_b\cos(c_cx))$ with $c_a=0$, testing whether an optimizer can ignore unnecessary nonlinear additive structure. The power-augmentation variants introduce identity power transformations by replacing a feature with $x^{c_a}$ or raising the whole expression to $c_a$, with $c_a=1$, testing difficult exponent-like parameters. The rational zero-term variants add terms of the form $(c_ax+c_b)/(c_cx+c_d)$, or the analogous numerator using a second feature, with $c_a=c_b=0$ and $c_c=c_d=1$. These four groups are the categories reported in Figure~\ref{fig:optimizer_bench}.

\section{Additional Experimental Results}
\label{app:additional_experiments}

\subsection{Complexity-aware Final Selection Details}
\label{app:complexity_aware_final_selection}

The complexity-aware selectors in Section~\ref{sec:complexity_aware_selection} are applied only after the evolutionary search has finished. They do not choose parents, generate mutations, fit coefficients, or alter the trajectory. For each task, we collect the full candidate trajectory, preserve the training-NMSE ranking used by the search, and compute simple structural features for each expression. The selectors then choose a fixed reporting shortlist from this already generated pool. Test and OOD NMSE are never used by these selectors.

Let candidate $i$ have training error $e_i$, original training-NMSE rank $r_i$, and structural features: tree size $s_i$, operator count $o_i$, fitted-parameter count $p_i$, and special-function count $h_i$. We use a weighted complexity score
\[
q_i = s_i + w_o o_i + w_p p_i + w_h h_i ,
\]
where the weights are fixed before evaluation. Pareto selection performs non-dominated sorting over $(\log e_i, q_i)$: a candidate is dominated if another candidate has no larger value in both objectives and a strictly smaller value in at least one. Candidates from the first front are selected first, ordered by $(\log e_i, q_i, r_i)$; if the shortlist is not full, the selected front is removed and the next front is added in the same way.

Occam selection first restricts the pool to candidates with near-best training fit. Specifically, if $e_\star$ is the best finite training NMSE, the near-best set contains candidates satisfying
\[
\log_{10}(\max(e_i,10^{-300}))-\log_{10}(\max(e_\star,10^{-300})) \leq \Delta .
\]
Within this near-best set, candidates are ranked by $(q_i,\log e_i,r_i)$, so simpler expressions are preferred only after the training error is already close to the best observed fit. If the near-best set is too small to fill the shortlist, the rule falls back to the full pool with the same complexity-first ordering.

MDL selection uses a single scalar objective,
\[
m_i = \log e_i + \alpha p_i + \beta s_i + \beta_o o_i ,
\]
and returns the candidates with the smallest $m_i$, with the original rank $r_i$ used only as a deterministic tie-breaker. This rule approximates a description-length tradeoff: the log training loss rewards fit, while parameter, tree-size, and operator penalties discourage unnecessarily complex formulas. All three selectors therefore test whether post-hoc complexity-aware reporting can recover exact symbolic forms that were generated by the search but ranked below numerically strong surrogate expressions by training NMSE alone.

\subsection{Non-MatSci Failure Correlation Audit}
\label{app:non_matsci_failure_correlation}

Table~\ref{tab:non_matsci_failure_correlation} reports the pairwise linear-dependence audit for the twelve FunctionEvolve failure cases (tasks with $\mathrm{SA@50}=0$) outside the MatSci subset. Unlike MatSci, these domains do not exhibit a dataset-wide input collapse, but several individual failures still contain highly correlated variable pairs. Four of these tasks---CRK0, CRK6, CRK22, and PO17---do eventually recover a symbolically correct expression, but only among candidates ranked beyond the top-50 budget (at ranks 693, 229, 313, and 224, respectively); they therefore count as $\mathrm{SA@50}$ failures here even though they are not failures under the full-trajectory exact-match audit.

\begin{center}
\centering
\captionof{table}{Pairwise linear dependence in non-MatSci failures.}
\label{tab:non_matsci_failure_correlation}
\scriptsize
\setlength{\tabcolsep}{6pt}
\renewcommand{\arraystretch}{1.06}
\sffamily
\begin{tabular}{@{}lcc@{}}
\toprule
Case & Variable pair & $R^2$ \\
\midrule
BPG4 & $t, P$ & \tabnum{0.723803} \\
BPG20 & $t, P$ & \tabnum{0.674447} \\
CRK0 & $t, A$ & \tabnum{0.208901} \\
CRK6 & $t, A$ & \tabnum{0.951128} \\
CRK22 & $t, A$ & \tabnum{0.517110} \\
CRK28 & $t, A$ & \tabnum{0.999234} \\
CRK29 & $t, A$ & \tabnum{0.732757} \\
PO0 & $x, t$ & \tabnum{0.001431} \\
PO16 & $x, t$ & \tabnum{0.004350} \\
PO17 & $x, v$ & \tabnum{0.028476} \\
PO23 & $x, v$ & \tabnum{0.964941} \\
PO43 & $x, v$ & \tabnum{0.368555} \\
\bottomrule
\end{tabular}
\end{center}

\subsection{Baseline NMSE Summary}
\label{app:selected_nmse_baselines_summary}
Table~\ref{tab:selected_nmse_baselines} reports the selected non-FunctionEvolve baselines from the same CSV. Here symbolic accuracy is reported as $\mathrm{SA@50}$ ($\mathrm{SA@1}$), Train/Test/OOD NMSE are split-wise medians over 129 tasks, and Test $\mathrm{Acc}_{\tau}$ counts tasks with test maximum relative error at most $\tau$.

\begin{table}[t]
\centering
\caption{Baseline NMSE summary for methods run with \emph{Claude Opus 4.6}. Symbolic accuracy is reported as $\mathrm{SA@50}$ ($\mathrm{SA@1}$). NMSE entries are medians over 129 tasks, and accuracy entries are raw test-set task counts.}
\label{tab:selected_nmse_baselines}
\scriptsize
\setlength{\tabcolsep}{4pt}
\renewcommand{\arraystretch}{1.1}
\sffamily
\resizebox{\textwidth}{!}{%
\begin{tabular}{@{}lrrrrrrr@{}}
\toprule
Method & $\mathrm{SA@50}$ ($\mathrm{@1}$) $\uparrow$ & Train NMSE $\downarrow$ & Test NMSE $\downarrow$ & OOD NMSE $\downarrow$ & Test $\mathrm{Acc}_{0.1}\uparrow$ & Test $\mathrm{Acc}_{0.01}\uparrow$ & Test $\mathrm{Acc}_{0.001}\uparrow$ \\
\midrule
Direct Prompting & \tabnum{2} (\tabnum{1}) & \tabnum{1.24e-3} & \tabnum{1.36e-3} & \tabnum{7.62e-2} & \tabnum{18} & \tabnum{7} & \tabnum{4} \\
OpenEvolve \citep{openevolve} & \textbf{\tabnum{24}} (\tabnum{5}) & \tabnum{4.50e-6} & \tabnum{4.30e-6} & \tabnum{6.52e-3} & \tabnum{27} & \tabnum{8} & \tabnum{1} \\
LLM-SR \citep{shojaee2024llm} & \textbf{\tabnum{24}} (\tabnum{11}) & \textbf{\tabnum{2.62e-6}} & \textbf{\tabnum{9.80e-7}} & \textbf{\tabnum{4.45e-4}} & \textbf{\tabnum{40}} & \textbf{\tabnum{13}} & \textbf{\tabnum{5}} \\
\bottomrule
\end{tabular}%
}
\end{table}

The corresponding per-task records are available in the GitHub raw data files described above.

\subsection{Complete Ablation Results}
\label{app:extended_ablation}
\label{app:selected_nmse_ablation_summary}

Table~\ref{tab:extended_ablation} reports the domain-wise symbolic-accuracy ablation results for both \emph{GPT-5.2-medium} and \emph{Claude Opus 4.6}. Each entry gives $\mathrm{SA@50}$ ($\mathrm{SA@1}$). Tables~\ref{tab:selected_nmse_gpt52_mainpy} and~\ref{tab:selected_nmse_opus46_mainpy} complement this view with aggregate NMSE and numerical-accuracy summaries for the same ablation settings. The two model backbones show the same qualitative pattern: removing the LLM Mutator or replacing the structure-aware coefficient optimizer with L-BFGS produces the largest $\mathrm{SA@50}$ drops, while removing the Generator, Selector, or AST Mutator also consistently reduces symbolic recovery.

\begin{table*}[t]
\centering
\caption{Extended ablation results on 129 LLM-SRBench tasks. Entries report $\mathrm{SA@50}$ ($\mathrm{SA@1}$).}
\label{tab:extended_ablation}
\small
\setlength{\tabcolsep}{2pt}
\begin{tabular*}{\textwidth}{@{\extracolsep{\fill}}llccccc@{}}
\toprule
Method & Model & Chemistry & Biology & Physics & Materials Science & Total \\
& & (36) & (24) & (44) & (25) & (129) \\
\midrule
Full & \emph{GPT-5.2-medium} & \textbf{30} (\textbf{13}) & \textbf{20} (\textbf{15}) & \textbf{37} (\textbf{36}) & \textbf{16} (5) & \textbf{103} (\textbf{69}) \\
w/o All & \emph{GPT-5.2-medium} & 12 (4) & 18 (1) & 2 (1) & 3 (0) & 35 (6) \\
w/o Generator & \emph{GPT-5.2-medium} & 21 (9) & \textbf{20} (13) & 29 (24) & 9 (3) & 79 (49) \\
w/o Selector & \emph{GPT-5.2-medium} & 20 (\textbf{13}) & 17 (12) & 14 (12) & 11 (\textbf{7}) & 62 (44) \\
w/o LLM Mutator & \emph{GPT-5.2-medium} & 12 (7) & 14 (11) & 12 (8) & 7 (1) & 45 (27) \\
w/o AST Mutator & \emph{GPT-5.2-medium} & \textbf{27} (12) & 18 (\textbf{14}) & 14 (13) & 11 (3) & 70 (42) \\
w/o Structure-aware Optimizer & \emph{GPT-5.2-medium} & 18 (3) & 14 (7) & 12 (5) & 2 (1) & 46 (16) \\
\midrule
Full & \emph{Claude Opus 4.6} & \textbf{31} (\textbf{16}) & \textbf{22} (\textbf{16}) & \textbf{39} (\textbf{34}) & \textbf{15} (\textbf{6}) & \textbf{107} (\textbf{72}) \\
w/o All & \emph{Claude Opus 4.6} & 12 (4) & 16 (5) & 4 (1) & 2 (0) & 34 (10) \\
w/o Generator & \emph{Claude Opus 4.6} & 27 (12) & \textbf{22} (14) & 33 (29) & 9 (3) & 91 (58) \\
w/o Selector & \emph{Claude Opus 4.6} & 25 (13) & 20 (11) & 20 (18) & 9 (4) & 74 (46) \\
w/o LLM Mutator & \emph{Claude Opus 4.6} & 11 (8) & 15 (11) & 13 (10) & 7 (2) & 46 (31) \\
w/o AST Mutator & \emph{Claude Opus 4.6} & 30 (\textbf{16}) & \textbf{22} (11) & 23 (20) & 9 (3) & 84 (50) \\
w/o AST Structure & \emph{Claude Opus 4.6} & 24 (14) & 20 (15) & 8 (7) & 8 (4) & 60 (40) \\
w/o Structure-aware Optimizer & \emph{Claude Opus 4.6} & 17 (6) & 17 (4) & 16 (12) & 3 (0) & 53 (22) \\
\bottomrule
\end{tabular*}
\end{table*}

\paragraph{NMSE and numerical accuracy summaries}
The following tables are computed from the NMSE result records and report all ablation settings used in our experiments. The records contain five aggregate rows per run (\texttt{BPG\_all}, \texttt{CRK\_all}, \texttt{MatSci\_all}, \texttt{PO\_all}, and \texttt{Total\_all}); these tables use only the 129 individual benchmark tasks. In each table, \textit{Method} names the same setting used in the ablation study, \textit{SA} is the number of exact symbolic matches, Train/Test/OOD NMSE report the median normalized mean squared error on the corresponding split, and Test $\mathrm{Acc}_{\tau}$ reports the number of tasks with test maximum relative error at most $\tau$. To keep the appendix compact, we omit the full per-task detail tables; detailed raw data are available at \url{https://anonymous.4open.science/r/FunctionEvolve}.

\paragraph{\emph{GPT-5.2-medium}}
Table~\ref{tab:selected_nmse_gpt52_mainpy} reports the \emph{GPT-5.2-medium} backbone. The \textit{w/o Generator}, \textit{w/o Selector}, \textit{w/o LLM Mutator}, \textit{w/o AST Mutator}, \textit{w/o Structure-aware Optimizer}, and \textit{w/o All} rows follow the ablation definitions in Section~\ref{sec:ablation}: they respectively remove LLM-based initialization, LLM-based parent selection, LLM mutation proposals, AST-rule mutations, the structure-aware coefficient optimizer, and the combined Generator/Selector/AST-Mutator/Structure-aware-Optimizer stack.

\begin{center}
\centering
\captionof{table}{Complete ablation NMSE summary for \emph{GPT-5.2-medium}. NMSE entries are medians over 129 tasks, and accuracy entries are raw test-set task counts.}
\label{tab:selected_nmse_gpt52_mainpy}
\scriptsize
\setlength{\tabcolsep}{4pt}
\renewcommand{\arraystretch}{1.1}
\sffamily
\resizebox{\textwidth}{!}{%
\begin{tabular}{@{}lrrrrrrr@{}}
\toprule
Method & SA $\uparrow$ & Train NMSE $\downarrow$ & Test NMSE $\downarrow$ & OOD NMSE $\downarrow$ & Test $\mathrm{Acc}_{0.1}\uparrow$ & Test $\mathrm{Acc}_{0.01}\uparrow$ & Test $\mathrm{Acc}_{0.001}\uparrow$ \\
\midrule
Full & \textbf{\tabnum{103}} & \textbf{\tabnum{1.82e-13}} & \textbf{\tabnum{2.08e-13}} & \textbf{\tabnum{2.21e-10}} & \textbf{\tabnum{111}} & \textbf{\tabnum{88}} & \textbf{\tabnum{73}} \\
w/o All & \tabnum{35} & \tabnum{4.20e-7} & \tabnum{4.27e-7} & \tabnum{3.98e-4} & \tabnum{49} & \tabnum{19} & \tabnum{11} \\
w/o Generator & \tabnum{79} & \tabnum{2.88e-13} & \tabnum{2.42e-13} & \tabnum{2.38e-9} & \tabnum{107} & \tabnum{81} & \textbf{\tabnum{71}} \\
w/o Selector & \tabnum{62} & \tabnum{3.77e-13} & \tabnum{4.63e-13} & \tabnum{3.72e-9} & \tabnum{99} & \tabnum{72} & \tabnum{63} \\
w/o LLM Mutator & \tabnum{45} & \tabnum{1.24e-12} & \tabnum{1.16e-12} & \tabnum{1.13e-8} & \tabnum{91} & \tabnum{67} & \tabnum{55} \\
w/o AST Mutator & \tabnum{70} & \tabnum{3.77e-13} & \tabnum{3.47e-13} & \tabnum{4.14e-9} & \tabnum{99} & \tabnum{74} & \tabnum{66} \\
w/o Structure-aware Optimizer & \tabnum{46} & \tabnum{4.01e-8} & \tabnum{4.33e-8} & \tabnum{1.80e-5} & \tabnum{69} & \tabnum{42} & \tabnum{22} \\
\bottomrule
\end{tabular}%
}
\end{center}

The corresponding per-task records are available in the GitHub raw data files described above.

\paragraph{\emph{Claude Opus 4.6}}
Table~\ref{tab:selected_nmse_opus46_mainpy} uses the same column definitions for the \emph{Claude Opus 4.6} backbone. In addition to the main ablation rows, \textit{w/o AST in LLM Interfaces} keeps the LLM channels but removes AST-heavy information from their prompts, and \textit{Rule-only + Structure-aware Optimizer} disables the Generator, the LLM Selector, and the LLM Mutator while retaining the default structure-aware coefficient optimizer.

\begin{center}
\centering
\captionof{table}{Complete ablation NMSE summary for \emph{Claude Opus 4.6}. NMSE entries are medians over 129 tasks, and accuracy entries are raw test-set task counts.}
\label{tab:selected_nmse_opus46_mainpy}
\scriptsize
\setlength{\tabcolsep}{4pt}
\renewcommand{\arraystretch}{1.1}
\sffamily
\resizebox{\textwidth}{!}{%
\begin{tabular}{@{}lrrrrrrr@{}}
\toprule
Method & SA $\uparrow$ & Train NMSE $\downarrow$ & Test NMSE $\downarrow$ & OOD NMSE $\downarrow$ & Test $\mathrm{Acc}_{0.1}\uparrow$ & Test $\mathrm{Acc}_{0.01}\uparrow$ & Test $\mathrm{Acc}_{0.001}\uparrow$ \\
\midrule
Full & \textbf{\tabnum{107}} & \textbf{\tabnum{1.34e-13}} & \textbf{\tabnum{1.38e-13}} & \textbf{\tabnum{1.78e-10}} & \textbf{\tabnum{113}} & \textbf{\tabnum{94}} & \textbf{\tabnum{78}} \\
w/o All & \tabnum{34} & \tabnum{2.31e-7} & \tabnum{1.93e-7} & \tabnum{2.15e-4} & \tabnum{53} & \tabnum{27} & \tabnum{15} \\
w/o Generator & \tabnum{91} & \tabnum{2.43e-13} & \tabnum{2.06e-13} & \tabnum{4.23e-10} & \tabnum{110} & \tabnum{87} & \tabnum{75} \\
w/o Selector & \tabnum{74} & \tabnum{4.14e-13} & \tabnum{4.65e-13} & \tabnum{4.71e-9} & \tabnum{100} & \tabnum{78} & \tabnum{63} \\
w/o LLM Mutator & \tabnum{46} & \tabnum{1.04e-12} & \tabnum{1.01e-12} & \tabnum{1.09e-8} & \tabnum{89} & \tabnum{67} & \tabnum{55} \\
w/o AST Mutator & \tabnum{84} & \tabnum{2.43e-13} & \tabnum{2.06e-13} & \tabnum{2.52e-9} & \tabnum{108} & \tabnum{83} & \tabnum{71} \\
w/o AST Structure & \tabnum{60} & \tabnum{4.25e-13} & \tabnum{5.09e-13} & \tabnum{6.73e-9} & \tabnum{96} & \tabnum{71} & \tabnum{59} \\
w/o Structure-aware Optimizer & \tabnum{53} & \tabnum{5.37e-9} & \tabnum{8.19e-9} & \tabnum{5.68e-6} & \tabnum{66} & \tabnum{43} & \tabnum{27} \\
Rule-only + Structure-aware Optimizer & \tabnum{29} & \tabnum{1.42e-10} & \tabnum{1.47e-10} & \tabnum{3.65e-7} & \tabnum{78} & \tabnum{53} & \tabnum{39} \\
\bottomrule
\end{tabular}%
}
\end{center}

The corresponding per-task records are available in the GitHub raw data files described above.

\subsection{Numerical Accuracy at Stricter Tolerances}
\label{app:strict_numerical_accuracy}

We distinguish two tolerance-based task-count metrics. Our $\mathrm{Acc}_{\tau}$ follows the convention used in LLM-SRBench reporting and in the main text: a task is counted only when the top-1 prediction has relative error at most $\tau$ at every evaluation point, equivalently when its maximum relative error is at most $\tau$. For a single test task, this indicator can be written as
\[
\mathrm{Acc}_{\tau}
= \mathbf{1}\left(
\max_{1 \le i \le N_{\mathrm{test}}}
\left|\frac{\hat{y}_i-y_i}{y_i}\right|
\le \tau
\right).
\]
A relaxed variant, denoted $95\%\mathrm{Acc}_{\tau}$, counts a task when at least 95\% of its evaluation points have relative error at most $\tau$.

SR-Scientist~\citep{xia2025srscientist} reports this relaxed $95\%\mathrm{Acc}_{\tau}$ metric for $\tau=0.01$ and $\tau=0.001$, which is not directly aligned with the reporting convention in LLM-SRBench-style papers or with our main-text $\mathrm{Acc}_{\tau}$ numbers. Table~\ref{tab:strict_test_accuracy} therefore reports test-set entries as $\mathrm{Acc}_{\tau}$ ($95\%\mathrm{Acc}_{\tau}$): values cited from SR-Scientist appear only in parentheses, while results from our executable runs report strict all-point counts with the corresponding relaxed 95\% counts in parentheses. Table~\ref{tab:strict_ood_accuracy} uses the same convention for OOD results.

\begin{table*}[t]
\centering
\caption{Test-set numerical accuracy at stricter tolerances on the 129-task synthetic subset of LLM-SRBench. Entries report raw task counts as $\mathrm{Acc}_{\tau}$ ($95\%\mathrm{Acc}_{\tau}$), with $\tau=0.01$ or $0.001$; parenthesized-only entries are the relaxed $95\%\mathrm{Acc}_{\tau}$ results cited from \citet{xia2025srscientist}. Best all-point $\mathrm{Acc}_{\tau}$ values are shown in bold, while the ground-truth row is a reference row. Entries marked with \textsuperscript{*} are converted from percentages reported in \citet{xia2025srscientist} using the corresponding domain sizes, with counts rounded independently.}
\label{tab:strict_test_accuracy}
\small
\setlength{\tabcolsep}{3pt}
\renewcommand{\arraystretch}{1.1}
\sffamily
\resizebox{\textwidth}{!}{%
\begin{tabular}{@{}l*{5}{cc}@{}}
\toprule
& \multicolumn{2}{c}{Chemistry (36)} & \multicolumn{2}{c}{Biology (24)} & \multicolumn{2}{c}{Physics (44)} & \multicolumn{2}{c}{Materials Science (25)} & \multicolumn{2}{c}{Total (129)} \\
\cmidrule(lr){2-3}\cmidrule(lr){4-5}\cmidrule(lr){6-7}\cmidrule(lr){8-9}\cmidrule(lr){10-11}
\textbf{Model} & $\mathsf{Acc}_{0.01}\uparrow$ & $\mathsf{Acc}_{0.001}\uparrow$ & $\mathsf{Acc}_{0.01}\uparrow$ & $\mathsf{Acc}_{0.001}\uparrow$ & $\mathsf{Acc}_{0.01}\uparrow$ & $\mathsf{Acc}_{0.001}\uparrow$ & $\mathsf{Acc}_{0.01}\uparrow$ & $\mathsf{Acc}_{0.001}\uparrow$ & $\mathsf{Acc}_{0.01}\uparrow$ & $\mathsf{Acc}_{0.001}\uparrow$ \\
\midrule
\rowcolor{groupgray}\multicolumn{11}{c}{\textit{Direct Prompting}} \\
\midrule
\emph{Claude Opus 4.6} & \tabnum{0} (\tabnum{1}) & \tabnum{0} (\tabnum{0}) & \tabnum{2} (\tabnum{4}) & \tabnum{1} (\tabnum{2}) & \tabnum{2} (\tabnum{3}) & \tabnum{2} (\tabnum{2}) & \tabnum{3} (\tabnum{10}) & \tabnum{1} (\tabnum{5}) & \tabnum{7} (\tabnum{18}) & \tabnum{4} (\tabnum{9}) \\
\midrule
\rowcolor{groupgray}\multicolumn{11}{c}{\textit{OpenEvolve \citep{openevolve}}} \\
\midrule
\emph{Claude Opus 4.6} & \tabnum{5} (\tabnum{8}) & \tabnum{0} (\tabnum{2}) & \tabnum{1} (\tabnum{2}) & \tabnum{0} (\tabnum{0}) & \tabnum{0} (\tabnum{9}) & \tabnum{0} (\tabnum{2}) & \tabnum{2} (\tabnum{11}) & \tabnum{1} (\tabnum{1}) & \tabnum{8} (\tabnum{30}) & \tabnum{1} (\tabnum{5}) \\
\midrule
\rowcolor{groupgray}\multicolumn{11}{c}{\textit{LLM-SR \citep{shojaee2024llm}}} \\
\midrule
\emph{Claude Opus 4.6} & \tabnum{5} (\tabnum{9}) & \tabnum{1} (\tabnum{1}) & \tabnum{1} (\tabnum{6}) & \tabnum{1} (\tabnum{1}) & \tabnum{0} (\tabnum{7}) & \tabnum{0} (\tabnum{2}) & \tabnum{7} (\tabnum{19}) & \tabnum{3} (\tabnum{10}) & \tabnum{13} (\tabnum{41}) & \tabnum{5} (\tabnum{14}) \\
\midrule
\rowcolor{groupgray}\multicolumn{11}{c}{\textit{SR-Scientist \citep{xia2025srscientist}}} \\
\midrule
\emph{Qwen3-Coder-480B-A35B-Instruct}\textsuperscript{*} & (\tabnum{15}) & (\tabnum{2}) & (\tabnum{12}) & (\tabnum{6}) & (\tabnum{15}) & (\tabnum{6}) & (\tabnum{22}) & (\tabnum{17}) & (\tabnum{63}) & (\tabnum{32}) \\
GLM-4.5-Air\textsuperscript{*} & (\tabnum{16}) & (\tabnum{4}) & (\tabnum{10}) & (\tabnum{4}) & (\tabnum{16}) & (\tabnum{7}) & (\tabnum{20}) & (\tabnum{18}) & (\tabnum{62}) & (\tabnum{32}) \\
\emph{GPT-OSS-120B}\textsuperscript{*} & (\tabnum{29}) & (\tabnum{23}) & (\tabnum{16}) & (\tabnum{10}) & (\tabnum{18}) & (\tabnum{15}) & (\tabnum{19}) & (\tabnum{15}) & (\tabnum{82}) & (\tabnum{64}) \\
\emph{GPT-OSS-20B}\textsuperscript{*} & (\tabnum{18}) & (\tabnum{8}) & (\tabnum{8}) & (\tabnum{5}) & (\tabnum{13}) & (\tabnum{7}) & (\tabnum{16}) & (\tabnum{10}) & (\tabnum{55}) & (\tabnum{30}) \\
\emph{Qwen3-Coder-30B-A3B-Instruct}\textsuperscript{*} & (\tabnum{8}) & (\tabnum{2}) & (\tabnum{5}) & (\tabnum{2}) & (\tabnum{8}) & (\tabnum{4}) & (\tabnum{20}) & (\tabnum{13}) & (\tabnum{42}) & (\tabnum{21}) \\
\emph{Qwen3-Coder-30B-A3B-Instruct} + RL\textsuperscript{*} & (\tabnum{13}) & (\tabnum{3}) & (\tabnum{7}) & (\tabnum{3}) & (\tabnum{11}) & (\tabnum{5}) & (\tabnum{21}) & (\tabnum{16}) & (\tabnum{53}) & (\tabnum{27}) \\
\midrule
\rowcolor{groupgray}\multicolumn{11}{c}{\textit{FunctionEvolve (Ours)}} \\
\midrule
\rowcolor{oursblue}\emph{GPT-5.2-medium} & \tabnum{26} (\tabnum{33}) & \tabnum{22} (\tabnum{32}) & \tabnum{14} (\tabnum{23}) & \tabnum{9} (\tabnum{23}) & \tabnum{28} (\tabnum{36}) & \tabnum{24} (\tabnum{32}) & \tabnum{20} (\tabnum{25}) & \tabnum{18} (\tabnum{24}) & \tabnum{88} (\tabnum{117}) & \tabnum{73} (\tabnum{111}) \\
\rowcolor{oursblue}\emph{Claude Opus 4.6} & \textbf{\tabnum{27}} (\tabnum{35}) & \textbf{\tabnum{23}} (\tabnum{34}) & \textbf{\tabnum{15}} (\tabnum{23}) & \textbf{\tabnum{9}} (\tabnum{23}) & \textbf{\tabnum{31}} (\tabnum{34}) & \textbf{\tabnum{27}} (\tabnum{34}) & \textbf{\tabnum{21}} (\tabnum{25}) & \textbf{\tabnum{19}} (\tabnum{22}) & \textbf{\tabnum{94}} (\tabnum{117}) & \textbf{\tabnum{78}} (\tabnum{113}) \\
\midrule
\rowcolor{groupgray}\textit{Reference: GT equations} & \tabnum{27} (\tabnum{35}) & \tabnum{24} (\tabnum{35}) & \tabnum{14} (\tabnum{23}) & \tabnum{10} (\tabnum{23}) & \tabnum{36} (\tabnum{39}) & \tabnum{33} (\tabnum{39}) & \tabnum{25} (\tabnum{25}) & \tabnum{25} (\tabnum{25}) & \tabnum{102} (\tabnum{122}) & \tabnum{92} (\tabnum{122}) \\
\bottomrule
\end{tabular}%
}
\end{table*}

\begin{table*}[t]
\centering
\caption{OOD numerical accuracy at stricter tolerances on the 129-task synthetic subset of LLM-SRBench. Entries report raw task counts as $\mathrm{Acc}_{\tau}$ ($95\%\mathrm{Acc}_{\tau}$), with $\tau=0.01$ or $0.001$; best all-point $\mathrm{Acc}_{\tau}$ values are shown in bold.}
\label{tab:strict_ood_accuracy}
\small
\setlength{\tabcolsep}{3pt}
\renewcommand{\arraystretch}{1.1}
\sffamily
\resizebox{\textwidth}{!}{%
\begin{tabular}{@{}l*{5}{cc}@{}}
\toprule
& \multicolumn{2}{c}{Chemistry (36)} & \multicolumn{2}{c}{Biology (24)} & \multicolumn{2}{c}{Physics (44)} & \multicolumn{2}{c}{Materials Science (25)} & \multicolumn{2}{c}{Total (129)} \\
\cmidrule(lr){2-3}\cmidrule(lr){4-5}\cmidrule(lr){6-7}\cmidrule(lr){8-9}\cmidrule(lr){10-11}
\textbf{Model} & $\mathsf{Acc}_{0.01}\uparrow$ & $\mathsf{Acc}_{0.001}\uparrow$ & $\mathsf{Acc}_{0.01}\uparrow$ & $\mathsf{Acc}_{0.001}\uparrow$ & $\mathsf{Acc}_{0.01}\uparrow$ & $\mathsf{Acc}_{0.001}\uparrow$ & $\mathsf{Acc}_{0.01}\uparrow$ & $\mathsf{Acc}_{0.001}\uparrow$ & $\mathsf{Acc}_{0.01}\uparrow$ & $\mathsf{Acc}_{0.001}\uparrow$ \\
\midrule
\rowcolor{groupgray}\multicolumn{11}{c}{\textit{Direct Prompting}} \\
\midrule
\emph{Claude Opus 4.6} & \tabnum{2} (\tabnum{2}) & \tabnum{0} (\tabnum{1}) & \tabnum{3} (\tabnum{4}) & \tabnum{1} (\tabnum{2}) & \tabnum{3} (\tabnum{3}) & \tabnum{2} (\tabnum{2}) & \tabnum{13} (\tabnum{13}) & \tabnum{11} (\tabnum{11}) & \tabnum{21} (\tabnum{22}) & \tabnum{14} (\tabnum{16}) \\
\midrule
\rowcolor{groupgray}\multicolumn{11}{c}{\textit{OpenEvolve \citep{openevolve}}} \\
\midrule
\emph{Claude Opus 4.6} & \tabnum{6} (\tabnum{7}) & \tabnum{1} (\tabnum{1}) & \tabnum{3} (\tabnum{3}) & \tabnum{1} (\tabnum{2}) & \tabnum{2} (\tabnum{9}) & \tabnum{0} (\tabnum{2}) & \tabnum{19} (\tabnum{21}) & \tabnum{14} (\tabnum{14}) & \tabnum{30} (\tabnum{40}) & \tabnum{16} (\tabnum{19}) \\
\midrule
\rowcolor{groupgray}\multicolumn{11}{c}{\textit{LLM-SR \citep{shojaee2024llm}}} \\
\midrule
\emph{Claude Opus 4.6} & \tabnum{7} (\tabnum{8}) & \tabnum{2} (\tabnum{2}) & \tabnum{4} (\tabnum{6}) & \tabnum{1} (\tabnum{2}) & \tabnum{2} (\tabnum{7}) & \tabnum{0} (\tabnum{2}) & \tabnum{20} (\tabnum{20}) & \tabnum{17} (\tabnum{17}) & \tabnum{33} (\tabnum{41}) & \tabnum{20} (\tabnum{23}) \\
\midrule
\rowcolor{groupgray}\multicolumn{11}{c}{\textit{FunctionEvolve (Ours)}} \\
\midrule
\rowcolor{oursblue}\emph{GPT-5.2-medium} & \tabnum{31} (\tabnum{34}) & \tabnum{26} (\tabnum{31}) & \textbf{\tabnum{16}} (\tabnum{23}) & \textbf{\tabnum{10}} (\tabnum{20}) & \tabnum{29} (\tabnum{35}) & \tabnum{21} (\tabnum{30}) & \textbf{\tabnum{25}} (\tabnum{25}) & \textbf{\tabnum{25}} (\tabnum{25}) & \tabnum{101} (\tabnum{117}) & \tabnum{82} (\tabnum{106}) \\
\rowcolor{oursblue}\emph{Claude Opus 4.6} & \textbf{\tabnum{32}} (\tabnum{35}) & \textbf{\tabnum{29}} (\tabnum{34}) & \textbf{\tabnum{16}} (\tabnum{23}) & \textbf{\tabnum{10}} (\tabnum{21}) & \textbf{\tabnum{32}} (\tabnum{34}) & \textbf{\tabnum{24}} (\tabnum{33}) & \textbf{\tabnum{25}} (\tabnum{25}) & \textbf{\tabnum{25}} (\tabnum{25}) & \textbf{\tabnum{105}} (\tabnum{117}) & \textbf{\tabnum{88}} (\tabnum{113}) \\
\bottomrule
\end{tabular}%
}
\end{table*}

\subsection{Reliability Audit for LLM-as-a-Judge Symbolic Accuracy}
\label{app:llm_judge_reliability}

Symbolic accuracy at $k$ counts a task as successful if any of the top-$k$ candidates is mathematically equivalent to the ground-truth expression. In this equivalence check, ground-truth constants are fixed, candidate constants are free, and physics-oscillator (PO) variables are allowed to take negative values, so expressions such as \texttt{v**c} are not treated as equivalent to \texttt{Abs(v)**c}. To reduce dependence on a single model judgment, each verification case is judged independently by \emph{GPT-5.2} and \emph{Claude Opus 4.6}. If their valid verdicts disagree, or if either run returns \texttt{verify\_error}, the case is manually adjudicated.

We audit 500 verification cases sampled randomly. For \emph{Opus 4.6}, repeated rows caused by retries are deduplicated by \texttt{sample\_id}; when available, the final non-\texttt{verify\_error} result is used. Table~\ref{tab:llm_judge_reliability} summarizes agreement. The judges agree on 476 valid \texttt{match}/\texttt{no\_match} decisions. There are 21 valid verdict disagreements and three cases involving \texttt{verify\_error}, so only 24/500 cases require manual intervention. Table~\ref{tab:judge_accuracy_after_audit} reports the resulting judge accuracies after manual adjudication, with and without counting the single unscorable case in the denominator.

\begin{center}
\centering
\captionof{table}{Reliability audit of the LLM-as-a-judge protocol for symbolic accuracy.}
\label{tab:llm_judge_reliability}
\small
\begin{tabular}{lrr}
\toprule
Category & Count & Rate \\
\midrule
Total audited samples & 500 & 100.0\% \\
Same valid verdict & 476 & 95.2\% \\
Valid verdict disagreement & 21 & 4.2\% \\
Cases involving \texttt{verify\_error} & 3 & 0.6\% \\
Manual intervention cases & 24 & 4.8\% \\
\bottomrule
\end{tabular}
\end{center}

\begin{center}
\centering
\captionof{table}{Judge accuracy after manual adjudication. The single unscorable case had no extractable final candidate formulas.}
\label{tab:judge_accuracy_after_audit}
\small
\begin{tabular}{lrrrr}
\toprule
Judge & Correct / 499 & Accuracy / 499 & Correct / 500 & Strict accuracy / 500 \\
\midrule
\emph{GPT-5.2} & 482 & 96.59\% & 482 & 96.40\% \\
\emph{Claude Opus 4.6} & 493 & 98.80\% & 493 & 98.60\% \\
\bottomrule
\end{tabular}
\end{center}

Table~\ref{tab:manual_judge_adjudication} lists the 24 cases requiring manual intervention. Among the 23 scorable cases in this set, \emph{GPT-5.2} is correct on 6 and \emph{Opus 4.6} is correct on 17. The only unscorable case is \texttt{sample\_id=294}, where both judges fail because the log contains no extractable final candidate formulas.

\begingroup
\scriptsize
\setlength{\tabcolsep}{3pt}
\begin{longtable}{rllllp{0.38\textwidth}}
\caption{Manual adjudication for cases where the two judges disagreed or at least one returned \texttt{verify\_error}.}
\label{tab:manual_judge_adjudication}\\
\toprule
ID & Case & \emph{GPT-5.2} & \emph{Opus 4.6} & Final & Main reason \\
\midrule
\endfirsthead
\toprule
ID & Case & \emph{GPT-5.2} & \emph{Opus 4.6} & Final & Main reason \\
\midrule
\endhead
\bottomrule
\endfoot
002 & BPG5 & \texttt{no\_match} & \texttt{match} & \texttt{match} & Rational common-denominator rewriting makes a candidate equivalent. \\
038 & BPG1 & \texttt{no\_match} & \texttt{match} & \texttt{match} & Candidate has both $P\exp(ct)$ and a rational term that can match after rewriting. \\
044 & BPG23 & \texttt{match} & \texttt{no\_match} & \texttt{match} & Candidate can represent the polynomial plus $P\exp(cP)$ structure. \\
080 & PO20 & \texttt{no\_match} & \texttt{match} & \texttt{no\_match} & PO domain requires \texttt{Abs(v)}; \texttt{v**c*x} is not equivalent. \\
082 & PO34 & \texttt{no\_match} & \texttt{match} & \texttt{no\_match} & Same PO-domain absolute-value issue. \\
112 & CRK1 & \texttt{match} & \texttt{no\_match} & \texttt{match} & Candidate can reduce to polynomial terms plus a phase-shifted log-trigonometric term. \\
145 & MatSci1 & \texttt{match} & \texttt{no\_match} & \texttt{no\_match} & Required Arrhenius term $\exp(-c/T)$ is absent. \\
169 & CRK29 & \texttt{no\_match} & \texttt{match} & \texttt{match} & Denominator form can represent the rational term and the linear/exponential terms. \\
173 & MatSci14 & \texttt{no\_match} & \texttt{match} & \texttt{match} & Candidate can represent the required linear-in-$T$ and $\epsilon^p$ cross terms. \\
231 & CRK19 & \texttt{match} & \texttt{no\_match} & \texttt{no\_match} & Requires $\sin(\log(A+1))$; candidates have no-phase cosine or wrong log arguments. \\
257 & CRK22 & \texttt{no\_match} & \texttt{match} & \texttt{match} & Common-denominator rewriting matches the ground truth. \\
294 & MatSci12 & \texttt{verify\_error} & \texttt{verify\_error} & unscorable & No final formulas or code snippets were extractable from the log. \\
304 & BPG3 & \texttt{no\_match} & \texttt{match} & \texttt{match} & Candidate has the needed $P^3$, $P^2$, and $P\exp(cP)$ basis functions. \\
315 & MatSci8 & \texttt{match} & \texttt{no\_match} & \texttt{match} & Candidate can represent $T\epsilon^p$, $T\epsilon$, $\epsilon^p$, and $\epsilon$. \\
362 & CRK3 & \texttt{no\_match} & \texttt{match} & \texttt{match} & Candidate has $A\exp(ct)$, $A^2$, and a phase-shifted $\sin(\log(A+1))$ term. \\
363 & MatSci2 & \texttt{verify\_error} & \texttt{no\_match} & \texttt{no\_match} & Candidates lack the required combination of $\epsilon^3$, $T\exp(-\epsilon)$, and $\exp(-\epsilon)$. \\
368 & MatSci18 & \texttt{verify\_error} & \texttt{match} & \texttt{match} & Candidate has $\epsilon^2+\epsilon(T+c)^2$. \\
371 & PO12 & \texttt{no\_match} & \texttt{match} & \texttt{no\_match} & \texttt{sqrt(v**c)} is not a valid substitute for \texttt{Abs(v)**(1/3)} on the PO domain. \\
374 & CRK22 & \texttt{no\_match} & \texttt{match} & \texttt{match} & Common-denominator rewriting matches the ground truth. \\
383 & MatSci28 & \texttt{no\_match} & \texttt{match} & \texttt{match} & Candidate contains the required $T\epsilon$, $T\log(\epsilon+1)$, and $\epsilon^2$ terms. \\
386 & PO22 & \texttt{no\_match} & \texttt{match} & \texttt{match} & Candidate can express $\exp(-|x|)$ via $\exp(c/\sqrt{x^{-2}})$. \\
391 & BPG11 & \texttt{no\_match} & \texttt{match} & \texttt{match} & Candidate can represent $P$, $P^2$, $P^{1/3}$, and $P\exp(ct)$. \\
415 & BPG16 & \texttt{no\_match} & \texttt{match} & \texttt{match} & Candidate can reduce to $P^3+P^2+P^{1/3}$. \\
439 & BPG1 & \texttt{no\_match} & \texttt{match} & \texttt{match} & Candidate can match with $P\exp(ct)$ plus a reciprocal rational term. \\
\end{longtable}
\endgroup

This audit does not assume that LLM judges are infallible. Instead, it shows that single-judge errors are measurable and rare under our prompt, and that the reported $\mathrm{SA@}k$ values use a two-judge-plus-manual-adjudication protocol rather than unchecked single-model judgments. The main \emph{GPT-5.2} error mode is rejecting valid rational-function equivalences that become clear after common-denominator rewriting. The main \emph{Opus 4.6} error mode is over-relaxing domain constraints, especially treating \texttt{v**c} as if it could stand for \texttt{Abs(v)**c} in PO tasks where negative velocities are allowed.






\end{document}